\documentclass{article}
\usepackage{lmodern}
\usepackage{etoolbox}
\usepackage{tcolorbox}
\usepackage{amsmath, amssymb}
\usepackage{graphicx}
\usepackage{url}
\usepackage{array}
\usepackage{adjustbox}
\usepackage{threeparttable}
\usepackage{arydshln}
\usepackage{etoolbox}
\usepackage{pifont}
\usepackage{placeins}
\usepackage{subcaption}
\usepackage{float}
\usepackage{comment}
\usepackage{caption}
\usepackage{natbib}
\usepackage[hidelinks]{hyperref}
\usepackage{xr-hyper} 
\usepackage{booktabs}
\usepackage{multirow}
\newcommand{\pkg}[1]{\texttt{#1}}
\usepackage[font=small]{caption}
\usepackage[font=footnotesize]{caption}
\usepackage[a-1b]{pdfx}  
\usepackage{subfiles}

\usepackage{titlesec} 
\usepackage{arxiv}

\titleformat{\section}
  {\normalfont\large\bfseries} 
  {\thesection}{1em}{}

\titleformat{\subsection}
  {\normalfont\normalsize\bfseries} 
  {\thesubsection}{1em}{}

\titleformat{\subsubsection}
  {\normalfont\small\bfseries} 
  {\thesubsubsection}{1em}{}

\newcommand{\review}[1]{\textcolor{black}{#1}}
\newcommand{\addition}[1]{\textcolor{black}{#1}}

\title{The C-index Multiverse}

\author{
Bego\~na B. Sierra \thanks{To whom correspondence should be addressed}\\
Edinburgh Cancer Centre \\
University of Edinburgh \\
Edinburgh, UK \\
\texttt{B.Bolos@ed.ac.uk} \\
\And
Colin McLean \\
Edinburgh Cancer Centre \\
University of Edinburgh \\
Edinburgh, UK \\
\texttt{Colin.D.McLean@ed.ac.uk} \\
\And
Peter S. Hall \\
Edinburgh Cancer Centre \\
University of Edinburgh \\
Edinburgh, UK \\
\texttt{p.s.hall@ed.ac.uk} \\
\And
Catalina A. Vallejos $^*$\\
MRC Human Genetics Unit \\
University of Edinburgh \\
Edinburgh, UK \\
\texttt{catalina.vallejos@ed.ac.uk} \\
}

\date{}

\begin{document}

\maketitle

\begin{abstract}{Quantifying out-of-sample discrimination performance for time-to-event outcomes is a fundamental step for model evaluation and selection in the context of predictive modelling. The concordance index, or C-index, is a widely used metric for this purpose, particularly with the growing development of machine learning methods. Beyond differences between proposed C-index estimators (e.g. Harrell’s, Uno’s and Antolini’s), we demonstrate the existence of a C-index multiverse among available R and \review{Python} software, where seemingly equal implementations can yield different results. This can undermine reproducibility and complicate fair comparisons across  models and studies. Key variation sources include tie handling and adjustment to censoring. Additionally, the absence of a standardised approach to summarise risk from survival distributions, result in another source of variation dependent on input types. We demonstrate the consequences of the C-index multiverse when quantifying predictive performance for several survival models (from Cox proportional hazards to recent deep learning approaches) on publicly available breast cancer data, and semi-synthetic examples. Our work emphasises the need for better reporting to improve transparency and reproducibility. This article aims to be a useful guideline, helping analysts when navigating the multiverse, providing unified documentation and highlighting potential pitfalls of existing software. All code is publicly available at:
\url{www.github.com/BBolosSierra/CindexMultiverse}.

}
\bigskip

\end{abstract}
\section{Introduction}
\label{sec1}


In time-to-event analysis, also known as survival analysis, the overarching goal is to model the time until an event of interest (e.g.~death, disease relapse) occurs. 
Whilst traditional statistical approaches such as the Cox proportional hazards (CPH) model \citep{Cox1972} primarily aim to quantify how a set of covariates (or features) affects event risk, there is an increasing interest in using time-to-event models to perform risk prediction, e.g.~to predict whether an individual will experience an event within a pre-specified prediction horizon. A variety of statistical and machine learning approaches have been proposed in this context, including random survival forests \citep[RSF;][]{Ishwaran2008} and deep learning (DL) methods \citep[e.g.][]{deephit2008, deepsurv2018}. See \cite{wang2019machine} and \cite{wiegrebe2024deep} for a review.

A critical step in the development and validation of risk prediction models is to assess predictive performance. This may inform model selection (e.g.~CPH vs RSF) or quantify how well a model's predictions extrapolate out-of-sample. Several metrics have been proposed for this purpose \citep{royston2013external}. 
In particular, discrimination metrics measure a model's ability
to correctly rank individuals according to their predicted risk. 
Here, we focus on the widely used concordance index (C-index, often referred to as the C-statistic). 
The C-index was proposed by \cite{Harrell1982} as natural extension for the area under the receiver operating characteristic curve (AUROC), which is a popular measure for predictive performance 
when modelling binary outcomes. Unlike in binary classification, time-to-event outcomes introduce ambiguity to the ranking of individuals due to censoring, where exact event times are not observed. Ties, where two individuals have the same recorded event time or for which a model predicts the same risk, can also be problematic when raking individuals. Several variations of the C-index have been proposed to address these challenges. These include those proposed by \cite{antolini2005} and \cite{Uno2011} which are often used as an alternative to Harrell's estimator. Antolini's C-index directly ranks individuals using the predicted survival distribution. Instead, Uno's and Harrell's typically require a transformation step, where the predicted survival distribution is summarised into a single measure of risk, which is subsequently used to rank individuals (note that such transformation is not required in the context of a CPH model).

\cite{sonabend2022avoiding} highlighted that the choice between different C-index estimators (and, where needed, transformations for the predicted survival function) can have important consequences for model selection, as estimators can differ in how they rank models, introducing the potential for \emph{C-hacking}, where users may choose the implementation which may give an unfair advantage to their proposed method. In this article, we demonstrate further issues related to the available software (in R and \review{Python}) that can be used to calculate such estimators. Indeed, two implementations of the same estimator (e.g.~Uno's) can produce different C-index estimates due to subtle choices in; for example, how ties are taken into account. These differences in C-index implementations 
leave users to navigate a \emph{C-index multiverse}. Despite this, 
the documentation provided by software packages often lacks sufficient detail to fully understand how the C-index has been computed (e.g.~whether tied observations were included). This lack of transparency can undermine reproducibility and complicate fair comparisons across models and studies.

This article is organised as follows. First, we review some of the most popular C-index estimators, highlighting conceptual discrepancies in how the concordance probability is defined. We also summarise different strategies to handle ties and to adjust for the censoring distribution, as well as possible transformations for the predicted survival distribution. We also propose a new transformation that bypasses numerical issues associated to existing options.
Subsequently, we investigate how existing software implementations (R and \review{Python}) handle these choices, providing a consistent documentation that can aid interpretation and support users when planning their analyses. Finally, we demonstrate practical implications of the C-index multiverse using real and synthetic data examples. All our code is publicly available at \url{www.github.com/BBolosSierra/CindexMultiverse}. This includes a Docker image to enhance reproducibility.

\section{Notation}
\label{sec:notation}




Let $T_1, \ldots, T_n$ be independent random variables representing time-to-event outcomes for $n$ individuals. For each subject $i \in \{1, \ldots, n\}$, let $\mathbf{x}_i$ be a $p$-dimensional vector of covariates. A common characteristic of time-to-event data is \emph{censoring}. In the presence of right censoring, the observed outcome is $T_i = \min( \tilde{T}_i, C_i)$, where $ \tilde{T}_i$ is the \emph{true} event time and $C_i$ is a censoring time. Let $\Delta_i = I\{ \tilde{T}_i \leq C_i\}$ be an event indicator, i.e.~$\Delta_{i} = 1$ when the event for subject $i$ is observed before or at censoring time $C_i$, and $\Delta_i = 0$ indicating censoring. 
Generally, survival models rely on \emph{independence} and \emph{non-informative} assumptions for the censoring mechanism \citep{andersen2014censored}. 

We assume that a time-to-event model has been trained based on some observed data, and that its primary goal is to predict event risk based on observed covariate values. 
\review{We denote the risk score used to assess concordance as $M(\cdot)$. This may be an output risk score directly provided by the model; for example, the linear predictor from a CPH model. More generally, this may be defined as a one-dimensional summary that is derived from the predicted survival distribution. For example, this may be based on the predicted risk at a specific time point $t$ or an aggregated metric across a given time range (more details are provided in Section 4.3.).}




\section{The C-index}
\label{sec:Cindex}

The C-index is a measure of discrimination that quantifies a model's ability to correctly rank individuals based on their predicted risk and observed time-to-event outcomes. Specifically, it evaluates \addition{whether} the model assigns a higher risk to individuals that experience the event earlier. \addition{Throughout this section, $M(\cdot)$ denotes the one-dimensional risk score used to assess concordance, with higher values indicating higher predicted risk.}  Given a random pair of subjects $(i, j)$, their observed survival times $(T_i < T_j)$ and covariates $(\mathbf{x}_i, \mathbf{x}_j)$, the C-index can be defined in terms of probability as follows:
\begin{equation} \label{eq:harrells}
C = P(M(\mathbf{x_i}) > M(\mathbf{x}_j) | T_{i} < T_{j}).
\end{equation}


\cite{Harrell1982} proposed to estimate the C-index as the ratio between the number of \emph{concordant} and \emph{comparable} pairs. Comparable pairs are those for which it is possible to identify who experienced the event earlier, i.e.~a pair is comparable if the subject that experienced the event at earlier time (${T_i} <  {T_j}$) is uncensored ($\Delta_i = 1$). 
A comparable pair is concordant if higher risk prediction is assigned to the individual that experienced the event earlier ($M(\mathbf{x}_i) > M(\mathbf{x}_j)$). Formally, the estimator proposed by \cite{Harrell1982} can be written as:
\begin{equation}\label{eq:harrells_estim}
\widehat{C} = \frac{\sum_{i=1}^n \sum_{j=1}^n \Delta_i I( {T_i} <  {T_j})  I({M}(\mathbf{x}_i) >  {M}(\mathbf{x}_j))}{\sum_{i=1}^n  \sum_{j=1}^n \Delta_i I( {T_i} <  {T_j}) },
\end{equation} where $I(\cdot)$ is an indicator function, which is equal to 1 if the argument value is true, 0 otherwise. 
A C-index of 1.0 indicates perfect discrimination, 
while a value of 0.5 suggests that the model performs no better than random (below 0.5 indicates poor performance and worse than random). 

Another popular definition 
was proposed by \cite{heagerty2005}, 
using a time truncation $\tau$ that limits the evaluation at a specific and clinically relevant pre-specified time window:
\begin{equation}\label{eq:uno}
C_\tau = P( {M}(\mathbf{x}_i) >  {M}(\mathbf{x}_j) | T_{i} < T_{j}, T_{i} < \tau)
\end{equation}


The definitions above 
were originally specified for a CPH model, where $M(\mathbf{x}_i)$ is the associated linear predictor $\mathbf{x}_i^T \beta$.
Under the PH assumption, the relative risk between individuals 
is constant, 
and the ranking is preserved by the survival distribution,
i.e.~$\mathbf{x}_i^T \beta > \mathbf{x}_j^T \beta$ if and only if $S(t|\mathbf{x}_i) < S(t|\mathbf{x}_j)$ for all $t$.
Hence, 
\eqref{eq:harrells} and \eqref{eq:uno} 
are time-independent measures of concordance. 
However, this is not true for all models. 
\review{In such cases, $M(\mathbf{x}_i)$ can be defined as a one-dimensional summary of the survival distribution, such as expected mortality, or restricted mean survival time (RMST),} \citep[][see also Section \ref{subsec:transformations}]{sonabend2022avoiding}.  Alternatively, \eqref{eq:harrells} and \eqref{eq:uno} can be adapted to use time-dependent predictions $M_t(\mathbf{x}_i)$ when assessing concordance. For example, \cite{Gerds2013} considers the estimated probability of observing an event by time $t$, i.e.~$M_t(\mathbf{x}_i) = 1 - S(t|\mathbf{x}_i)$. Whilst this may be appropriate when the goal is to make predictions at a specific time-point, it can also be problematic \citep{blanche2019}.

\review{\cite{antolini2005} proposed an alternative measure that bypasses the need to reduce 
each predicted survival distribution to a single measure of risk}:
\begin{equation}
 \label{eq:antolini}
C_{td} = P(S(T_i | \mathbf{x}_i) < S(T_i |\mathbf{x}_j) | T_i < T_j).  
\end{equation} 
\review{For each comparable pair $(i,j)$, $C_{td}$ uses the predicted survival probabilities of both individuals at subject $i$'s event time to determine  their ranking. Thus, as subject $i$ changes across comparable pairs, the time at which predictions are evaluated also changes.} \review{Note that, in this sense, $C_{td}$ can be considered time-dependent, as the ranking depends on subject $i$'s event time. This is conceptually different to the use of a time-dependent risk score $M_t(\mathbf{x}_i)$ in the context of $C$ or $C_{\tau}$. In the latter, concordance is evaluated based on a common pre-specified evaluation time $t$ across all comparable pairs.} 
\review{$C_{td}$ is a popular choice when assessing predictive performance for models that allow non-proportional hazards \citep[e.g.~][]{randomForestSRC, deephit2008}. For such models, predicted survival curves can cross,
and the different transformations of the predicted survival distribution can result in different risk rankings.} The reasoning underlying the definition in \eqref{eq:antolini} is that, \addition{if subject $i$ experiences the event before subject $j$, their predicted survival probability at $T_i$ should be lower than that predicted for subject $j$}. \review{ As with other definitions of concordance, however, $C_{td}$ has limitations. In particular, it does not adjust for censoring, which can result in biased estimates \citep{faruk2025}; we discuss censoring adjustment in Section \ref{subsec:censoring}.\\
 Moreover, concordance measures have been shown to lack propriety as scoring rules in different settings, meaning that the true prediction is not guaranteed to achieve the optimal expected score. In particular, the C-index based on predictions at a fixed time $t$ is not a proper scoring rule for evaluating $t$-year predictions \citep{blanche2019}, while $C_{td}$ is not a proper scoring rule for evaluating predicted survival distributions \citep{rindt22aPMLR}.} \addition{More recently, \cite{gandy2025SJS} proposed a hazard-based concordance measure for models with crossing hazards, and established its propriety under stated conditions.}






\section{The C-index multiverse}
\label{sec:multiverse}

The C-index definitions in Section \ref{sec:Cindex} conceptually differ in how concordant and comparable pairs are specified. Whilst $C$ and $C_{\tau}$ use a one-dimensional summary $M(\mathbf{x}_i)$ to assess 
concordance, 
\review{$C_{td}$ uses the predicted survival probabilities of both individuals at subject $i$'s event time to evaluate their ranking} (Figure \ref{fig:diagCs}). Furthermore, only $C_{\tau}$ considers a time truncation (controlled by $\tau$) in the definition of comparable pairs. Here, we focus on further discrepancies across C-index estimators, even when they estimate the same probability. These are primarily driven by two choices. Firstly, how the formation of concordant and comparable pairs is affected by ambiguous pairs arising in the presence of ties (times 
or predictions). 
Secondly, whether the estimator takes into account the censoring distribution. 
Furthermore, we summarise possible transformations for the predicted survival distribution that can be applied when estimating $C$ or $C_{\tau}$. The combination of different concordance probability definitions and varying strategies to handle ambiguous pairs, censoring and input transformations, result in a multiverse of C-index estimators.


\subsection{Accounting for ties}
\label{subsec:ties}

The C-index relies on the ordering of times and the predicted risk to rank individuals. Tied times (${T_i} = {T_j}$) and tied predictions $(M(\mathbf{x}_i) = M(\mathbf{x}_j))$ introduce ambiguity in the ranking. Tied times can occur when two individuals experience the event simultaneously. Although such situations have a zero probability if $T_i$ is a continuous random variable, they can be observed in practice when survival times are rounded (e.g.~to days) or due to administrative censoring (i.e.~fixed end of follow-up). Therefore, the number of comparable pairs and, indirectly, the number of concordant pairs (only comparable pairs can be concordant) are affected by tied times. Tied predictions are likely to occur with categorical or discrete covariates, as individuals with the same covariate values receive same predicted values, increasing ties. In contrast, continuous predictors tend to produce a wider range of predictions, reducing the likelihood of a pair of subjects having a tie. Tied predictions affect the evaluation of concordant pairs during C-index computation.

The estimator in \eqref{eq:harrells_estim} ignores all ties, and it is sometimes interpreted as a simplification to what was proposed in \cite{Harrell1982}. A more general form can be defined as: \begin{equation} \label{eq:generalCties}
\hat{C} = \frac{\sum_{i = 1}^n \sum_{j=1}^n \left[A_{ij} + w_o B_{ij}\right]\left[C_{ij} + w_p D_{ij}\right]}{\sum_{i = 1}^n \sum_{j=1}^n \left[A_{ij} + \omega_o B_{ij}\right]}, 
\end{equation} where the treatment of ties is specified via $\omega_o$ (for tied outcomes) and $\omega_p$ (for tied predictions). The remaining elements in \eqref{eq:generalCties} are defined as $A_{ij} = \Delta_i I(T_i < T_j)$, $B_{ij} = \Delta_i (1-\Delta_j) I(T_i = T_j)$, $C_{ij} = I(M(\mathbf{x}_i) > M(\mathbf{x}_j))$ and $D_{ij} = I(M(\mathbf{x}_i) = M(\mathbf{x}_j))$. If $\omega_o = 0$, the definition of comparable pairs is as in \eqref{eq:harrells_estim}. Instead, if $\omega_o = 1$, a pair with tied times $({T_i} = {T_j})$ is counted as comparable if subject $i$ is uncensored ($\Delta_i = 1$) and $j$ is censored ($\Delta_j = 0$) (whilst this is generally the case for estimators that consider tied times in the definition of comparable pairs, there are exceptions where a slightly different behavior is applied; see Section \ref{sec:software}). Similarly, if $\omega_p = 0$, tied predictions $(M(\mathbf{x}_i) = M(\mathbf{x}_j))$ are ignored when counting concordant pairs. When $\omega_p > 0$, tied predictions are often interpreted as \emph{partially concordant} and given a half credit ($\omega_p = 0.5$).

Similar to 
\addition{\cite{alabdallah2024concordance}}, \eqref{eq:generalCties} can be decomposed as a weighted average between the simplified estimator in \eqref{eq:harrells_estim} and three additional components, i.e. \eqref{eq:generalCties} is equal to: \begin{equation} \label{eq:generalCties_decomposition}
\alpha \underbrace{\left[ \hat{C} + \omega_p \overbrace{\frac{\sum_{i = 1}^n \sum_{j=1}^n A_{ij} D_{ij}}{\sum_{i = 1}^n \sum_{j=1}^n A_{ij}}}^{\shortstack{Tied predictions}} \right]}_{\shortstack{Comparable pairs such that \\ $T_i < T_j$, $\Delta_i = 1$}} + (1-\alpha)  \underbrace{\left[ \overbrace{\frac{\sum_{i = 1}^n \sum_{j=1}^n B_{ij} C_{ij}}{\sum_{i = 1}^n \sum_{j=1}^n B_{ij}}}^{\shortstack{Concordant predictions}} + \omega_p \overbrace{\frac{\sum_{i = 1}^n \sum_{j=1}^n B_{ij} D_{ij}}{\sum_{i = 1}^n \sum_{j=1}^n B_{ij}}}^{\shortstack{Tied predictions}}  \right]}_{\shortstack{Comparable pairs such that \\ $T_i = T_j$, $\Delta_i = 1$, $\Delta_j = 0$}},
\end{equation} where $\alpha = \left[ \sum_{i = 1}^n \sum_{j=1}^n A_{ij} \right] / \left[ \sum_{i = 1}^n \sum_{j=1}^n \left[ A_{ij} + \omega_o B_{ij}  \right] \right] $. In \eqref{eq:generalCties_decomposition}, the term linked to concordant predictions among tied times, quantifies discrimination between censored and non censored observations. The two terms associated to tied predictions corresponds to cases in which the model cannot effectively distinguish individuals $i$ and $j$. Their inclusion can have an strong influence when the proportion of tied predictions is high (e.g.~when only categorical covariates are included), and may deflate the final estimate \citep{yan2008investigating, hartman2023}.

\subsection{Accounting for the censoring distribution}
\label{subsec:censoring}

Time-to-event data often includes right-censored observations 
for which the true event time ($\tilde{T}_i$) is unknown and only a lower bound $C_i$ is recorded. 
Censoring affects the estimator in \eqref{eq:harrells_estim} by: (1) defining the observed event time of subject $i$ as ${T_i} = \min(\tilde{T_i}, C_i)$, and (2) forming comparable pairs depending on the event indicators $(\Delta_i, \Delta_j)$. 
Fundamentally, the estimator excludes pairs with both censored individuals ($\Delta_i = \Delta_j = 0$), or when the censoring time occurs earlier than the event time within a pair ($\Delta_i = 0$). 
As a result, censoring directly affects which pairs are comparable, leading to a potentially biased estimator that depends on the study-specific censoring distribution \citep{Uno2011, Gerds2013}.


For a CPH model, \cite{gonen2005concordance} proposed an alternative estimator for $C$ that is more robust to censoring as it is solely calculated using $\mathbf{x}^T_i \beta$ without considering $T_i$ or $\Delta_i$. Instead, 
\cite{Uno2011} extended \eqref{eq:harrells_estim}, adjusting the contribution of each pair $(i,j)$ via Inverse Probability Censoring Weights (IPCW), which we denote as $W_{ij}$. This 
up-weights pairs where $\tilde{T}_i$ is observed 
despite a high probability of being censored.
 Unlike Harrell's, \cite{Uno2011} estimates $C_{\tau}$. This is to avoid unstable estimates arising when censoring occurs predominantly at later times, as the IPCWs applied to individuals having an event at later time points (often low risk subjects) can become extremely large. 
Uno's estimator is defined as:
\begin{equation}\label{eq:uno_estim}
\widehat{C}_\tau = 
\frac{
\sum_{i=1}^n \sum_{j=1}^n W_{ij} \Delta_i I( {T}_i <  {T}_j,  {T}_i < \tau) I(M(\mathbf{x}_i) > M(\mathbf{x}_j))
}{
\sum_{i=1}^n \sum_{j=1}^n W_{ij} \Delta_i  I( {T}_i <  {T}_j,  {T}_i < \tau) 
}.
\end{equation}
Assuming that the censoring distribution is independent of the covariates, \cite{Uno2011} proposed to use $W_{ij} = \hat{G}( {T}_i)^{-2}$, where $G(t) = P(C_i > t)$ is calculated 
using a Kaplan-Meier estimate \citep{kaplan1958nonparametric} of the censoring distribution. 
However, this can be extended to account for the effect of covariates (e.g.~estimating $G(\cdot)$ using a CPH model). 
\review{The IPCW estimator proposed by \cite{Uno2011} assumes that censoring is independent of both the event time and covariates. When censoring depends on observed covariates, this can be accounted for by modelling the censoring distribution as a function of those covariates (e.g.~using a CPH model), assuming independent censoring conditional on the covariates \citep{Gerds2013}. Therefore the validity of IPCW-based estimators depends on the corresponding censoring assumptions being satisfied.} 

\review{A less commonly used approach to handling censoring is the marginal C-index \citep{kumar2022SR, haider2020JMLR}. This approach imputes event times for individuals censored at $C_i$ with a ``best-guess'', defined via the expected event time $T_i$, conditional on being larger than $C_i$. This is based on a Kaplan-Meier estimate of the event time distribution, which ignores the effect of the available covariates. 
        Thereby, otherwise incomparable pairs due to censoring become comparable with a partial weight (i.e., with earlier censoring receiving less weight).  
         Unlike IPCW, these weights reflect confidence in the imputation (i.e., with earlier censoring receiving less weight) rather than the probability of remaining uncensored. However, the ordering of these pairs still relies on imputed event times, and the estimation of event time distribution used for the ``best guess'' relies on the validity of censoring assumptions being satisfied.}

\subsection{Survival distribution transformations}
\label{subsec:transformations}

The above estimators for $C$ and $C_{\tau}$ require the survival distribution $S(\cdot|\mathbf{x}_i)$ to be reduced into a single measure of risk $M(\mathbf{x}_i)$. 
For instance, when assessing how well a model ranks individuals at a given time point $t$ ($\in \mathbb{R}^+$), time-dependent predictions $M_{t}(\mathbf{x}_i) = 1 - S(t |\mathbf{x}_i)$ 
can be used. However, \cite{blanche2019} noted that this method is not ``proper" as an optimal model may not maximize its discrimination performance when evaluated at a specific time-point.
Alternatively, the mean or median survival time could be used to rank subjects, \review{as in \cite{kumar2022SR}. In this case, the quantity used for ranking, ($M(\mathbf{x}_i)$), is a predicted median \emph{time} (i.e., the predicted time at which their survival probability reaches 0.5)}. However, \review{the mean or median may not be} well defined 
due to the presence of censoring and limited follow-up as survival probabilities not always reaching zero within the study time horizon \citep{sonabend2022avoiding}. 
Similarly, when the survival probability does not drop below 0.5, the median survival time is undefined.

Instead, \cite{Ishwaran2008} considered a measure of expected mortality based on the cumulative hazard function, i.e.~${M(\mathbf{x}_i)} = \sum_{t \in \mathcal{T}} H(t | \mathbf{x}_i)$, where $H(t | \mathbf{x}_i) = -\log(S(t | \mathbf{x}_i))$ 
and $\mathcal{T}$ represents the set of observed (censored and non-censored) 
survival times. 
This approach was also advocated by \cite{sonabend2022avoiding}. However, as demonstrated in Section \ref{sec:metabric}, the sum can become numerically unstable for high risk individuals, 
for which $S(t|\mathbf{x}_i) \approx 0$ 
at later time points (in the limiting case, where $S(t|\mathbf{x}_i) = 0$, the summation is not finite). 
To address the numerical instability, one could truncate the sum to a maximum time $T^*$ but the choice of such limit is not trivial a priori as it depends of the predicted values for $S(\cdot | \mathbf{x}_i)$. 

Here, we suggest to define $M(\mathbf{x}_i)$ based on the Restricted Mean Survival Time (RMST), i.e.: \begin{equation} \label{eq:RMST}
M(\mathbf{x}_i) = - \text{RMST} = -
\mathbb{E} \left[\min (\tilde{T}_i, T^*) \right] = - \int_0^{T^*} S(t \mid \mathbf{x}_i) \, dt \approx - \sum_{t \in \mathcal{T},\, t \leq T^*} S(t \mid \mathbf{x}_i) \cdot \Delta t.
\end{equation} where $\Delta t$ is the difference between consecutive values in the grid $\mathcal{T}$ (this notation assumes equally spaced time-points). The negative sign was added to ensure $M(\mathbf{x}_i)$ follows the same direction as expected mortality (i.e.~higher values for higher risk individuals). RMST has been proposed as an outcome measure for clinical trials that does not rely on a PH assumption \citep{royston2013restricted, Averbuch2025}. Similar to the expected mortality measure proposed by \cite{Ishwaran2008}, this can be seen as a Riemman sum approximation for the area under the survival function $S(\cdot | \mathbf{x}_i)$. 
Unlike the mean or median survival, RMST is always well defined as $S(\cdot | \mathbf{x}_i)$ is bounded to $[0,1]$ and the integral is truncated to a finite interval ($[0, T^*]$). 
Whilst the upper limit $T^*$ needs to be specified, this can be done  a priori based on domain knowledge (e.g.~a clinically relevant period), rather than as a data-dependent value which is chosen to avoid numerical stability when  
$S(t | \mathbf{x}_i) \approx 0$. 
\review{For individual-level predictions, RMST should be interpreted with respect to this pre-specified horizon, rather than as the unrestricted expected event time. Nevertheless, its interpretation may not be straightforward when communicating individual-level predictions to patients or clinicians.}

There can be practical challenges when using expected mortality or RMST to compare discrimination performance for different models whose software implementation generates estimates of $S(\cdot | \mathbf{x}_i)$ with a different time resolution (i.e.~$\mathcal{T}$ can vary across models). This can be addressed by interpolating model-specific predictions across a common grid (Supplementary Note S1). 
 



\begin{figure}[H]
    \centering
    \vspace{-3mm}
    \makebox[\textwidth][c]{\includegraphics[width=18.5cm]{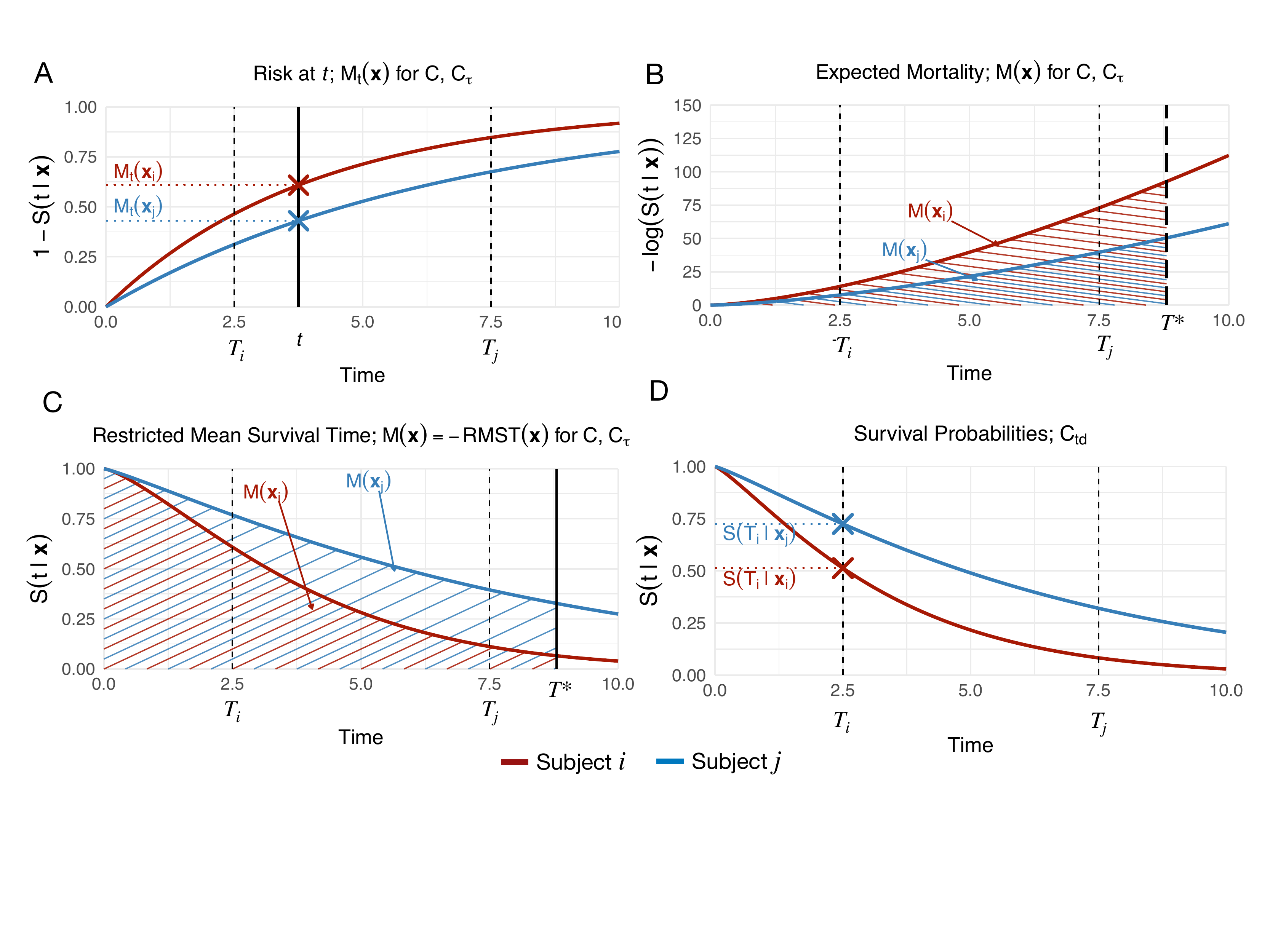}}
    \vspace{-28mm}
    \caption{\review{Schematic representation of the predicted survival distributions and derived risk scores used to assess concordance for two individuals, $i$ and $j$, denoted by red and blue, respectively. \textbf{(A--C)} illustrate three transformations of predicted survival distributions into one-dimensional risk scores, as required by $C$ and $C_\tau$ (Section 4.3). \textbf{(A)} Risk at a fixed time $t$, $M_t(\mathbf{x}) = 1-S(t\mid\mathbf{x})$, where the predicted survival distribution for both individuals are evaluated at the same pre-specified time $t$. \textbf{(B)} Expected mortality, $M(\mathbf{x})$, calculated from $-\log S(t\mid\mathbf{x})$ over the prediction time grid $\mathcal{T}$; the hatched areas illustrate the subject-specific quantities contributing to this transformation, up to the maximum prediction time $T^*$ used in the grid. \textbf{(C)} Restricted mean survival time (RMST), where the hatched areas represent the subject-specific areas under the predicted survival curves up to the pre-specified horizon $T^*$, where the risk score is defined as $M(\mathbf{x})=-\mathrm{RMST}(\mathbf{x})$. \textbf{(D)} $C_{td}$, for which the ranking is determined separately for each comparable pair by evaluating the predicted survival distributions at the subject $i$'s event time $T_i$ for both individuals.}}
    \label{fig:diagCs}
    \end{figure}

\section{Software implementations}
\label{sec:software}

Several C-index implementations are available as 
in R \citep{Rlanguage} and \review{Python} \citep{python}. Beyond the differences in the estimators described above, seemingly equal implementations often have subtle variations, leading to different C-index estimates (Table \ref{tab:implementations1}; Supplementary Tables S1-S5). Whilst users often cite the original publications in their analyses \citep[e.g.~][]{Harrell1982}, some implementations can be seen as a \emph{mix-and-match} of options that is not always consistent with published estimators. For instance, {\tt pysurvival} incorporates IPCW 
but does not match the estimator by \cite{Uno2011} due to the absence of time truncation (as defined by $\tau$). In contrast, the implementation in {\tt survival} does include a time truncation but, as a default, does not consider IPCW. Note that {\tt pysurvival} reports the C-index as $\max(\hat{C}
, 1 - \hat{C})$, potentially masking poor model performance. 

The inclusion of tied times or predictions as comparable or concordant pairs also varies across implementations. For example, {\tt pec} provides flexible options for the inclusion of different types of ties (Supplementary Table S5). Most implementations count tied times as comparable when $\Delta_i = 1$ and $\Delta_j = 0$, but this is not the case for {\tt survC1}. Moreover, only {\tt SurvMetrics} considers tied times such that $\Delta_i = 1$ and $\Delta_j = 1$ as comparable. Such cases are also counted as concordant when there is also a tie in the predictions (if not, a half-credit is added when computing the number of concordant pairs). Note that {\tt SurvMetrics} counts tied times as partially concordant when $\Delta_i = 1$, $\Delta_j = 0$ and $M(\mathbf{x}_i) < M(\mathbf{x}_j)$ (all other implementations count such cases as comparable but non concordant pairs). For most implementations, the inclusion of tied predictions as concordant is a user's choice (in which case, they are counted as a half-credit). Whilst most implementations require an exact match to identify tied predictions, {\tt sksurv} allows users to specify a tolerance threshold in terms of the absolute difference between the individuals' predictions. Note that, by default, {\tt pycox} calculates a modified version of Antolini's estimator ({\tt method = "adj\_antolini"}). The latter considers most of the modifications proposed by \cite[Section 5.1, step 3]{Ishwaran2008} to account for  ties when calculating the number of comparable and concordant pairs. To compute the original Antolini's C-index, users need to specify {\tt method = "antolini"}. 


Among implementations that include IPCW, there are differences in how these are calculated. For example, {\tt pec} supports multiple methods to estimate the censoring distribution ($G(t)$). These include non-parametric (e.g.~Kaplan-Meier), and (semi-)parametric methods (e.g.~CPH). To assess out-of-sample performance, {\tt sksurv} requires a training set to estimate $G(t)$, while others compute this from the test set directly (this can be overridden by the user by providing the same data as a test and training set). \addition{This choice can affect the accuracy of IPCW-based performance estimates under some data settings \citep{prince2025ipcw}}. There are also minor differences in how to calculate 
IPCW. 
For instance, {\tt pec} uses the product of the inverse of $G(T_i -)$ (i.e.~censoring probability just before $T_i$) and the inverse of $G(T_i)$. 
In contrast, {\tt sksurv} defines IPCW as the inverse of $G(T_i)^2$. As in \citeauthor{antolini2005}, \pkg{pycox} does not use IPCW. However, the {\tt EvalSurv} function (which is required by \pkg{pycox} prior to applying {\tt concordance\_td}) allows users to specify a censoring distribution. The latter is only used when calculating alternative metrics \citep[e.g. IPCW adjusted Brier Score;][]{graf1999assessment}, but may unintentionally lead users to assume that IPCW are incorporated. 

Another important distinction between C-index estimators is how time truncation ($\tau$) is handled in the presence of IPCW. As highlighted by \cite{Uno2011}, failing to truncate time can result in very large weights, and a handful of pairs can dominate the final estimate. This issue is particularly evident in the case of {\tt pysurvival}, which includes IPCW but does not allow for time truncation. However, this can be easily overlooked in other implementations which do allow users to specify $\tau$ but where no time truncation is done under default settings. This is the case for {\tt sksurv} ({\tt concordance\_index\_ipcw} function) and {\tt survival}. Note that, {\tt pec} mitigates this issue by setting a default value for $\tau$ as the maximum of the non-censored survival times. However, as illustrated in the next section, this is not sufficient to avoid exploding IPCW. {\tt survC1} does not specify a default value for $\tau$ and it returns an error message if this is not provided by the user. 

Finally, there are also differences in how the input needs to be specified. Most implementations that estimate $C$ or $C_\tau$ allow a numeric vector of individual-specific predictions $M(\mathbf{x}_i)$. This allows users to specify an arbitrary transformation for the survival function (see Section \ref{subsec:transformations}). Others, can internally extract predictions from a trained model (e.g.~a {\tt coxph} object generated with the {\tt survival} R package) using a specific type of transformation. For example, \pkg{SurvMetrics} can extract time-dependent predictions 
from the survival distribution i.e.~$M_t(\mathbf{x}_i) = 1- S(t | \mathbf{x}_i)$ (if $t$ is not specified, it 
uses the median of the non-censored event times).  
Following \cite{blanche2019},  \pkg{pec} outputs a warning when such transformation is selected. 
\review{More recently, other packages such as \pkg{SurvivalEVAL} use predicted median survival times as the default for ranking.}

\begin{table}[htbp]
\centering
\caption{\textbf{Software implementations of C-index estimators}. [\ding{51}] indicates that the implementation considers ties  (times or predictions) or IPCW. [\ding{55}] indicates that the implementation does not consider ties or IPCW to account for the censoring distribution. $^*$ indicates inclusion of ties or IPCW is a user-defined option (relevant parameters are described in Supplementary Tables S1-S5). In such cases, [\ding{51}] or [\ding{55}] denotes the default behaviour of the function. Whilst most methods only count tied times ($T_i = T_j$) as comparable when $\Delta_i = 1$ and $\Delta_j = 0$, $^*$$^*$ indicates that \pkg{SurvMetrics} also considers cases where $\Delta_i = \Delta_j = 1$ and that these are counted differently when counting the number of concordant pairs (Supplementary Table S7).
All implementations that estimate $C$ or $C_{\tau}$ are consistent with the use of a scalar $M(\textbf{x}_i)$ input (e.g.~as defined by the transformations in Section \ref{subsec:transformations}). However, the format required for each function's input varies (e.g.~a vector of predictions or a fitted model). By default, \pkg{sksurv} defines tied predictions using a tolerance threshold of $10^{-8}$. Specifically, two predictions are considered tied if the absolute difference is less than or equal than the threshold. By default, \pkg{survival} and \pkg{sksurv} do not use a time truncation parameter $\tau$ (effectively reverting to a $C$ estimator). This needs to be explicitly specified by the user (via the {\tt ymax} and {\tt tau} arguments, respectively). As a default value for $\tau$, \pkg{pec} use the maximum of the non-censored survival times. \pkg{survC1} does not specify a default value for $\tau$ and it returns an error message if this is not provided by the user. \pkg{survival} allows for different weighting schemes, where the default is uniform ($W_{ij} = 1$, i.e.~no IPCW). Implementations that estimate $C_{td}$ use the survival distribution $S(t|\mathbf{x}_i)$ as input. By default, \pkg{pycox} calculates a modified version of the estimator proposed by \cite{antolini2005} which takes into account ties. Users need to specify {\tt method = "antolini"} to calculate Antolini's original estimator. $^*$$^*$$^*$ The original estimator considers tied times when $\Delta_i = 1$ and $\Delta_j = 0$. The modified version includes further cases, following most of the adjustments proposed by \cite{Ishwaran2008}.}
\label{tab:implementations1}%
\centering
\begin{minipage}{0.8\textwidth}
\centering
\begin{threeparttable}
   {
   \begin{tabular}{llcllc}
    \toprule
    \multicolumn{1}{l}{\multirow{1}[1]{*}{Package/Function }}  &
    \multicolumn{1}{l}{\multirow{1}[1]{*}{Language}} &
        \multicolumn{1}{c}{\multirow{1}[1]{*}{Tied Times}}    &
    \multicolumn{1}{c}{\multirow{1}[1]{*}{Tied Predictions}}&
   \multicolumn{1}{c}{\multirow{1}[1]{*}{IPCW}}\\
   
    \midrule
      \multicolumn{5}{l}{\bf Implementations that estimate ${C}$} \\
    \midrule

          \multicolumn{1}{l}{\multirow{1}[1]{*}{\pkg{Hmisc} / \pkg{rcorr.cens} }\tnote{1}} &
          \multicolumn{1}{l}{\multirow{1}[1]{*}{R}} &
          \multicolumn{1}{c}{\multirow{1}[1]{*}{\ding{51}}} & 
          \multicolumn{1}{c}{\multirow{1}[1]{*}{\ding{51}$^*$}} &
          \multicolumn{1}{c}{\multirow{1}[1]{*}{\ding{55}}} \\

          \multicolumn{1}{l}{\multirow{1}[1]{*}{\pkg{SurvMetrics} / \pkg{Cindex} \tnote{2}}} &
          \multicolumn{1}{l}{\multirow{1}[1]{*}{R}} &
          \multicolumn{1}{c}{\multirow{1}[1]{*}{\ding{51$^*$$^*$}}} & 
          \multicolumn{1}{c}{\multirow{1}[1]{*}{\ding{51$^*$$^*$}}} &
          \multicolumn{1}{c}{\multirow{1}[1]{*}{\ding{55}}} \\

          \multicolumn{1}{l}{\multirow{1}[1]{*}{\pkg{lifelines} / \pkg{concordance\_index}\tnote{3}}}&
          \multicolumn{1}{l}{\multirow{1}[1]{*}{\review{Python}}} &
          \multicolumn{1}{c}{\multirow{1}[1]{*}{\ding{51}}} & 
          \multicolumn{1}{c}{\multirow{1}[1]{*}{\ding{51}}} &
          \multicolumn{1}{c}{\multirow{1}[1]{*}{\ding{55}}}  \\

          \multicolumn{1}{l}{\multirow{1}[1]{*}{\pkg{pysurvival} / \pkg{concordance\_index}\tnote{4}}} &
          \multicolumn{1}{l}{\multirow{1}[1]{*}{\review{Python}}} &
          \multicolumn{1}{c}{\multirow{1}[1]{*}{\ding{51}}} &  
          \multicolumn{1}{c}{\multirow{1}[1]{*}{\ding{51}$^*$}} &
          \multicolumn{1}{c}{\multirow{1}[1]{*}{\ding{51}}} \\

          \multicolumn{1}{l}{\multirow{1}[1]{*}{\pkg{scikit-survival (sksurv) /} }} & 
          \multicolumn{1}{l}{\multirow{2}[1]{*}{\review{Python}}} &
          \multicolumn{1}{c}{\multirow{2}[1]{*}{\ding{51}}} & 
          \multicolumn{1}{c}{\multirow{2}[1]{*}{\ding{51$^*$}}} & 
          \multicolumn{1}{c}{\multirow{2}[1]{*}{\ding{55}}} \\
          \multicolumn{1}{l}{\multirow{1}[1]{*}{\pkg{ concordance\_index\_censored} \tnote{5}}} & \\ 
          
     \midrule
       \multicolumn{5}{l}{\bf Implementations that estimate $ C_\tau$} \\
    \midrule  

          \multicolumn{1}{l}{\multirow{1}[1]{*}{\pkg{pec} / \pkg{cindex} \tnote{6}}} &
          \multicolumn{1}{l}{\multirow{1}[1]{*}{R}} &\multicolumn{1}{c}{\multirow{1}[1]{*}{\ding{51}$^*$}} & 
          \multicolumn{1}{c}{\multirow{1}[1]{*}{\ding{51}$^*$}} & 
          \multicolumn{1}{c}{\multirow{1}[1]{*}{\ding{51}}}    \\

          \multicolumn{1}{l}{\multirow{1}[1]{*}{\pkg{survival} / \pkg{concordance}} \tnote{7}} &
          \multicolumn{1}{l}{\multirow{1}[1]{*}{R}} &
          \multicolumn{1}{c}{\multirow{1}[1]{*}{\ding{51}}} & 
          \multicolumn{1}{c}{\multirow{1}[1]{*}{\ding{51}}} &
          \multicolumn{1}{c}{\multirow{1}[1]{*}{\ding{55}$^*$}} \\

          \multicolumn{1}{l}{\multirow{1}[1]{*}{\pkg{survC1} / \pkg{Est.Cval} \tnote{8}}}&
          \multicolumn{1}{l}{\multirow{1}[1]{*}{R}} &
          \multicolumn{1}{c}{\multirow{1}[1]{*}{\ding{55}}} & 
          \multicolumn{1}{c}{\multirow{1}[1]{*}{\ding{51}}} & 
          \multicolumn{1}{c}{\multirow{1}[1]{*}{\ding{51}}} \\

          \multicolumn{1}{l}{\multirow{1}[1]{*}{\pkg{scikit-survival (sksurv)/}}}&
          \multicolumn{1}{l}{\multirow{2}[1]{*}{\review{Python}}} &\multicolumn{1}{c}{\multirow{2}[1]{*}{\ding{51}}} & 
          \multicolumn{1}{c}{\multirow{2}[1]{*}{\ding{51$^*$}}} & 
          \multicolumn{1}{c}{\multirow{2}[1]{*}{\ding{51}}}\\
          \multicolumn{1}{l}{\multirow{1}[1]{*}{ \pkg{concordance\_index\_ipcw} \tnote{5}}} &  \\
    \midrule
       \multicolumn{5}{l}{\bf Implementations that estimate $ C_{td}$} \\
    \midrule

          \multicolumn{1}{l}{\multirow{1}[1]{*}{\pkg{pycox} / \pkg{concordance\_td}\tnote{9}}} &
          \multicolumn{1}{l}{\multirow{1}[1]{*}{\review{Python}}} &
          \multicolumn{1}{c}{\multirow{1}[1]{*}{\ding{51$^{***}$}}} &
          \multicolumn{1}{c}{\multirow{1}[1]{*}{\ding{51$^{***}$}}} &
          \multicolumn{1}{c}{\multirow{1}[1]{*}{\ding{55}}} \\

          
     \bottomrule
    \end{tabular}

    \begin{tablenotes}
    \item[]
      \textsuperscript{1} \cite{hmisc},
      \textsuperscript{2} \cite{survMetrics},
      \textsuperscript{3} \cite{lifelines},
      \textsuperscript{4} \cite{pysurvival_cite},
      \textsuperscript{5} \cite{sksurv},
      \textsuperscript{6} \cite{pec},
      \textsuperscript{7} \cite{survival-package},
      \textsuperscript{8} \cite{survC1},
      \textsuperscript{9} \cite{pycox}
      
    \end{tablenotes}}

\end{threeparttable}
\end{minipage}
\end{table}%
\AtBeginEnvironment{tablenotes}{\normalsize}

\section{Metabric Case Study} \label{sec:metabric}

\subsection{Dataset overview}

The Molecular Taxonomy of Breast Cancer International Consortium (METABRIC) dataset  \citep{curtis2012} is a popular benchmark for new survival methods \citep[e.g.~][]{deephit2008, deepsurv2018}. The data is publicly available through the cBioPortal \citep{cerami2012cbio, gao2013integrative}. Among others, the dataset contains  clinical observations ($n = 2509$ patients) and the gene expression profiles ($n = 1980$ out of $2509$) measured at the time of diagnosis. 
We used cBioPortal 
to download and link 
clinical and molecular data. In our analysis, we define time to death (all cause, in months) as the outcome. As in \cite{deepsurv2018}, covariates are defined by five clinical features and four gene indicators. Only individuals with complete information in all covariates were included, leading to a $n=1937$ cohort. 
Summary statistics for the derived cohort are available in Table \ref{tab:tab:summary}. 

\begin{table}[htbp]
\caption{\label{tab:tab:summary}Summary statistics for the derived METABRIC cohort with patients stratified by survival status (censored versus all cause death). This includes information about four gene indicators (MKI67, EGFR, ERBB2, PGR) and three clinical variables (hormone therapy, radiotherapy, chemotherapy and ER IHC). ER IHC stands for ER (Estrogen Receptor) status determined by immunohistochemistry (IHC).}
\centering
\begin{tabular}{lcc}
\toprule
Covariate &  \shortstack{Censored\\(N = 812)} & \shortstack{Event\\(N = 1125)} \\
\midrule
MKI67 & 5.83 (0.35) & 5.90 (0.33)\\
EGFR & 6.26 (0.82) & 6.17 (0.88)\\
ERBB2 & 10.64 (1.35) & 10.85 (1.36)\\
PGR & 6.31 (1.06) & 6.19 (0.99)\\
Hormone Therapy &  & \\
\hspace{.5em}No & 319 (39\%) & 413 (37\%)\\
\hspace{.5em}Yes & 493 (61\%) & 712 (63\%)\\
Radiotherapy &  & \\
\hspace{.5em}No & 271 (33\%) & 501 (45\%)\\
\hspace{.5em}Yes & 541 (67\%) & 624 (55\%)\\
Chemotherapy &  & \\
\hspace{.5em}No & 623 (77\%) & 903 (80\%)\\
\hspace{.5em}Yes & 189 (23\%) & 222 (20\%)\\
ER IHC &  & \\
\hspace{.5em}Negative & 193 (24\%) & 246 (22\%)\\
\hspace{.5em}Positive & 619 (76\%) & 879 (78\%)\\
Age at diagnosis & 56 (11) & 64 (13)\\
Survival times (in months) & 159 (71) & 100 (70)\\
\bottomrule
\footnotesize 
Mean (standard deviation); n (\%)
\end{tabular}
\end{table}

\subsection{Methods}

We consider the following time-to-event models: CPH \citep{Cox1972}, Random Survival Forests \citep[RSF;][]{Ishwaran2008}, 
Cox-Time \citep{kvamme2019time}, DeepSurv \citep{deepsurv2018} and DeepHit \citep{deephit2008}. Note that CPH and DeepSurv are based on a PH specification; whilst CPH considers a linear predictor, DeepSurv uses a neural network to account for more complex covariate effects. Software packages (and the corresponding versions) used to fit each model are shown in Section \ref{sec:softwareimple}. Where relevant, model-specific hyperparameter values are shown in Supplementary Table S6. 
For each model, all the implementations in Table \ref{tab:implementations1} were used to assess out-of-sample predictive performance \review{(i.e.,  performance on individuals not included in the data used to train the model)} via 
5-fold cross-validation (CV; Supplementary Figure S1). Stratified sampling was used to ensure similar rates of events and censoring across folds (Supplementary Table S7). Numerical covariates were standarised to have zero mean an unit variance using statistics computed on the training set. To quantify uncertainty, bootstrap was used to calculate 95\% confidence intervals for C-index estimates. 


For each CV fold, all models were used to estimate $S(\cdot|\mathbf{x_i})$ for each subject in the corresponding test set. For CPH, the estimator by \cite{breslow1972discussion} (using the \pkg{survfit} function of the \pkg{survival} R package) was used for this purpose. When estimating $C_{\tau}$, we considered two possible values for the truncation parameter: $\tau = 120$ months (i.e.~10 years) to resemble a potentially relevant clinical time window, and the maximum of the non-censored survival times (i.e.~$\tau = \max\{T_i : \Delta_i = 1\}$). When evaluating out-of-sample predictive performance, the latter was calculated on the test set, so it slightly varied across folds (and across bootstrap samples). 
Note that the second choice is often used as the default value for $\tau$ in $C_{\tau}$ implementations (Section \ref{sec:software}). 

As in \cite{sonabend2022avoiding}, we calculated $C$ and $C_{\tau}$ estimates using different transformations of $S(\cdot|\mathbf{x_i})$ (Section \ref{subsec:transformations}). For this purpose, we considered: (i) time-dependent predictions, i.e.~$M_t(\mathbf{x}_i) = 1 - S(t | \mathbf{x}_i)$ for varying values of $t$, (ii) expected mortality, i.e.~$M(\mathbf{x}_i) = \sum_{t \in \mathcal{T}} -\log S(t|\mathbf{x}_i)$, and (iii) the RMST-based summary introduced in Equation \eqref{eq:RMST}. 
When calculating (ii) and (iii), model-specific estimated survival probabilities were linearly interpolated over a common grid defined as $\mathcal{T} = \{0, 1, \ldots, 355\}$ (Supplementary Note S1). Expected mortality is not well defined in the limiting case where $S(t | \mathbf{x}_i) = 0$. In those cases, a small constant ($\epsilon = \min(S(t | x_i) : S(t|x_i) > 0,  t \in \mathcal{T})$) was added. When calculating RMST, we used $T^* = 355$. 

Furthermore, we performed a sensitivity analysis  in which we artificially increased the number of ties present in the data. For this analysis, we only considered C-index implementations for which the inclusion of ties is a user's choice (Table \ref{tab:implementations1}). To increase the number of tied times ($T_i = T_j$), we rounded the observed times (censored and uncensored) to their nearest integer. To increase the number of tied predictions ($M(\mathbf{x}_i) = M(\mathbf{x}_i)$), we considered a simplified model with age as the only covariate. Tied predictions were obtained by rounding age to: (i) its nearest integer or (ii) into 5-year bins. For each CV fold, the rounding was applied in the test set only (i.e.~models were fitted using the original, unrounded, data). 

Finally, to understand the effects of IPCW under varying censoring rates, we simulated semi-syntethic datasets (100 datasets for each setting) based on METABRIC (Supplementary Note S2). For each dataset, covariate values were set as the observed values for $n = 1000$ randomly selected individuals. Uncensored times $\tilde{T}_i$ were generated using a Weibull PH model with parameters equal to those estimated for corresponding METABRIC subsample. Censoring times $C_i$ were generated from a Weibull distribution, varying its parameters to generate different levels of censoring (within the same simulation setting, there rate of censoring varies stochastically; the range is reported). Observed times and event indicators were set as $T_i = \min(\tilde{T}_i, C_i)$ and $\Delta_i = I(\tilde{T}_i < C_i)$, respectively. For each dataset, out-of-sample discrimination for a CPH model was quantified via 5-fold CV. RMST was used to calculate $C$ and $C_{\tau}$ estimates. We considered $\tau = \max(T_i: \Delta_i=1)$ and $\tau = 100$ months (chosen to avoid $\tau$ being higher than the maximum uncensored time, thereby losing the effect of time truncation). 
As reference values, we also calculated oracle estimates based on the uncensored event times ($\tilde{T}_i$), with survival probabilities calculated using the true model. See details in Supplementary Note S3.

\subsection{Results}
\begin{figure}[htbp]
    \centering
    \includegraphics[width=0.98\textwidth]{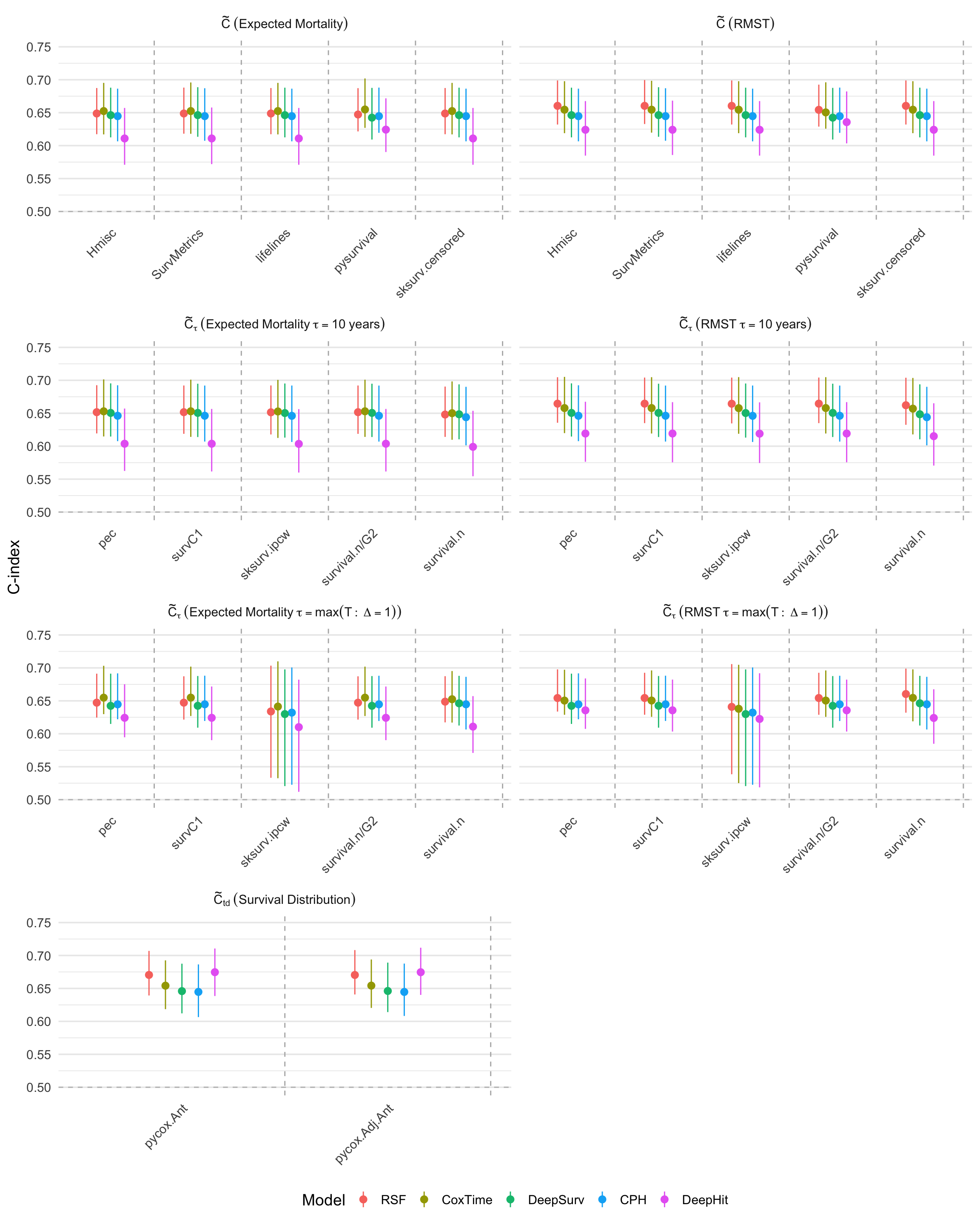}
    \caption{Comparison of C-index estimates calculated by different implementations when applied to the METABRIC first CV fold as a test (Supplementary Figure S1). Dots represent the estimates and vertical lines represent the 95\% confidence interval which was calculated using bootstrap. Estimates are grouped by the corresponding definition of the concordance probability ($C$, $C_\tau$ or $C_{td}$). For $C_\tau$, we consider $\tau$ equal to 10 years or to the maximum of the uncensored survival times. For $C$ and $C_{\tau}$ two transformations of the survival function were used: expected mortality and negative RMST (Equation \eqref{eq:RMST}). sksurv.censored and sksurv.ipcw indicate whether the {\tt concordance\_index\_censored} or {\tt concordance\_index\_ipcw} functions were used. survival.n/G2 and survival.n indicate different weighting options:  IPCW and uniform, respectively. pycox.Ant is the original Antolini's estimator, pycox.Adj.Ant which adjusts the treatment of ties by applying most of the modifications suggested by \cite{Ishwaran2008}.}
    \label{fig:fold1_RMSF}
\end{figure}

\begin{figure}[htbp]
    \centering
    \includegraphics[width=\textwidth]{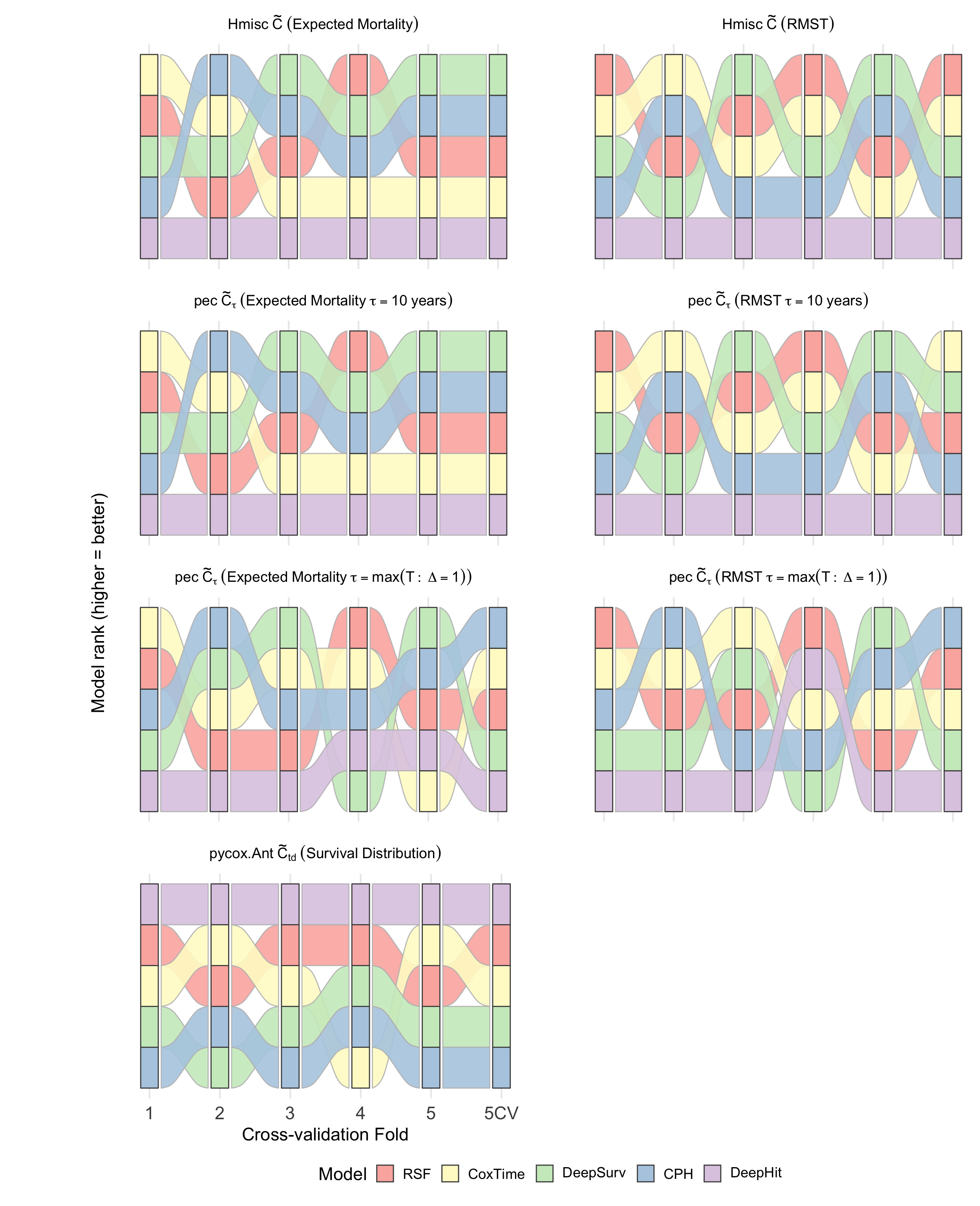}
    \caption{Model ranking (top = highest; bottom = lowest) based on different C-estimates. These were calculated for each CV fold, and combined across all CV folds (5CV) within the METABRIC analysis (Supplementary Figure S1). For brevity, we only consider estimates obtained using \pkg{Hmisc} ($C$), \pkg{pec} ($C_{\tau}$) and \pkg{pycox} ($C_{td}$). For  \pkg{Hmisc}, and \pkg{pec}, expected mortality and RMST were considered as possible transformations for the survival function. For \pkg{pycox}, we calculated Antolini's original estimator (obtained by using {\tt method = "antolini"}; denoted as pycox.Ant).  }
    \label{fig:alluvial}
\end{figure}

\vspace{-2mm}
Overall, different C-index implementations led to a varying model ranking 
(Figure \ref{fig:fold1_RMSF}, Supplementary Figures S2-S4, \review{Supplementary Figure S14}). C-index estimates, and hence model rankings, also varied across CV folds (Figure \ref{fig:alluvial}). 
\review{The most prominent difference was for DeepHit when comparing estimates of $C_{td}$ (which, for a pair $(i,j)$, uses the predicted survival probabilities evaluated at subject $i$'s event time to determine their ranking);}
Figure \ref{fig:diagCs}D) versus $C$ and $C_{\tau}$ (which use a one-dimensional summary; Figure \ref{fig:diagCs}A-C). Whilst DeepHit was consistently ranked as the top model by $C_{td}$ estimates, the opposite occurred for $C$ and $C_{\tau}$. This is likely because DeepHit training is based on a loss function that optimises the ranking of individuals as in $C_{td}$. Furthermore, DeepHit predicted survival curves that were more tightly concentrated than those predicted by other methods (Supplementary Figure S5), resulting in a more narrow distribution of RMST (and expected mortality, Supplementary Figure S6). 
When considering all folds combined, DeepHit had particularly low $C$ and $C_{\tau}$ estimates (Supplementary Figure S4). This is due to fold-to-fold differences in the predicted survival curves (Supplementary Figure S7).

As it may be expected, C-index estimates for models that rely on a PH specification (CPH and DeepSurv) were less sensitive to the definition of concordance probability ($C$, $C_{\tau}$ or $C_{td}$) or the transformation applied to the survival distribution. When comparing $C$ and $C_{\tau}$ estimates calculated using the same transformation (e.g.~RMST), the differences were driven by the interplay between the use of time truncation in the definition of comparable pairs (controlled by $\tau$) and the use of IPCW to adjust for the censoring distribution. In the absence of time truncation (or when $\tau$ is the maximum of the observed times), IPCW tend to disproportionally up-weight the contribution of pairs for which $T_i$ is large and an event is observed ($\Delta_i = 1$; Supplementary Figure S8). As highlighted by \cite{Uno2011}, this leads to more unstable estimates. This instability is particularly prominent for the $C_{\tau}$ estimator implemented in \pkg{sksurv}. The latter quantifies out-of-sample predictive performance using IPCW calculated on the training set, resulting into wider confidence intervals in comparison with those that calculate IPCW using the test set (\pkg{sksurv.ipcw} in Figure \ref{fig:fold1_RMSF}). As such, the increased uncertainty is influenced by discrepancies in the upper tail of the distribution of the censored times (Supplementary Figure S8), and it is attenuated when $\tau = 10$ years. For these data, $C$ and $C_{\tau}$ estimates (and, consequently, model rankings) were largely similar when $\tau = 10$ years, regardless of the inclusion of IPCW (except for \pkg{pysurvival} which incorporates IPCW but estimates $C$). \review{\pkg{SurvivalEVAL} also allows $\tau$ truncation without IPCW for Harrell's $C$ and Antolini's $C_{td}$, affecting the resulting estimates (Supplementary Figure S14).}

For $C$ and $C_{\tau}$, the choice of transformation used to summarise the survival distribution was critical. With the exception of CPH and DeepSurv, the use of time-dependent risk $M_t(\mathbf{x}_i)$ as an input led to estimates that varied with respect to $t$, and the pattern differed across models (Supplementary Figures S2-S3). Expected mortality and RMST produced time-independent measures of concordance, but were not consistent with each other (Figures \ref{fig:fold1_RMSF}-\ref{fig:alluvial}; Supplementary Figure S4). This is largely explained by numerical issues associated to the calculation of expected mortality. As seen in Supplementary Figure S9, the latter is sensitive to the right tail of the survival distribution where $S(t | \mathbf{x_i}) \approx 0$ and $\log(S(t | \mathbf{x_i}))$ is unbounded. This is particularly problematic for high risk individuals, who are likely to experience the event early, where some models predicted $S(t | \mathbf{x_i}) = 0$. In principle, this could be resolved by truncating the sum (similar to the use of $T^*$ in RMST) but this would need to be done in a data-specific manner rather than on a clinically relevant time-period. As a result, we suggest to use RMST as a more robust transformation. \review{The importance of the transformation choice was further supported by a supplementary analysis using \pkg{SurvivalEVAL}, that uses the predicted median time when assessing concordance, producing different estimates that differ from those obtained based RMST (Supplementary Figure S14).}  

\begin{figure}[htbp]
    \centering
    \includegraphics[width=15cm]{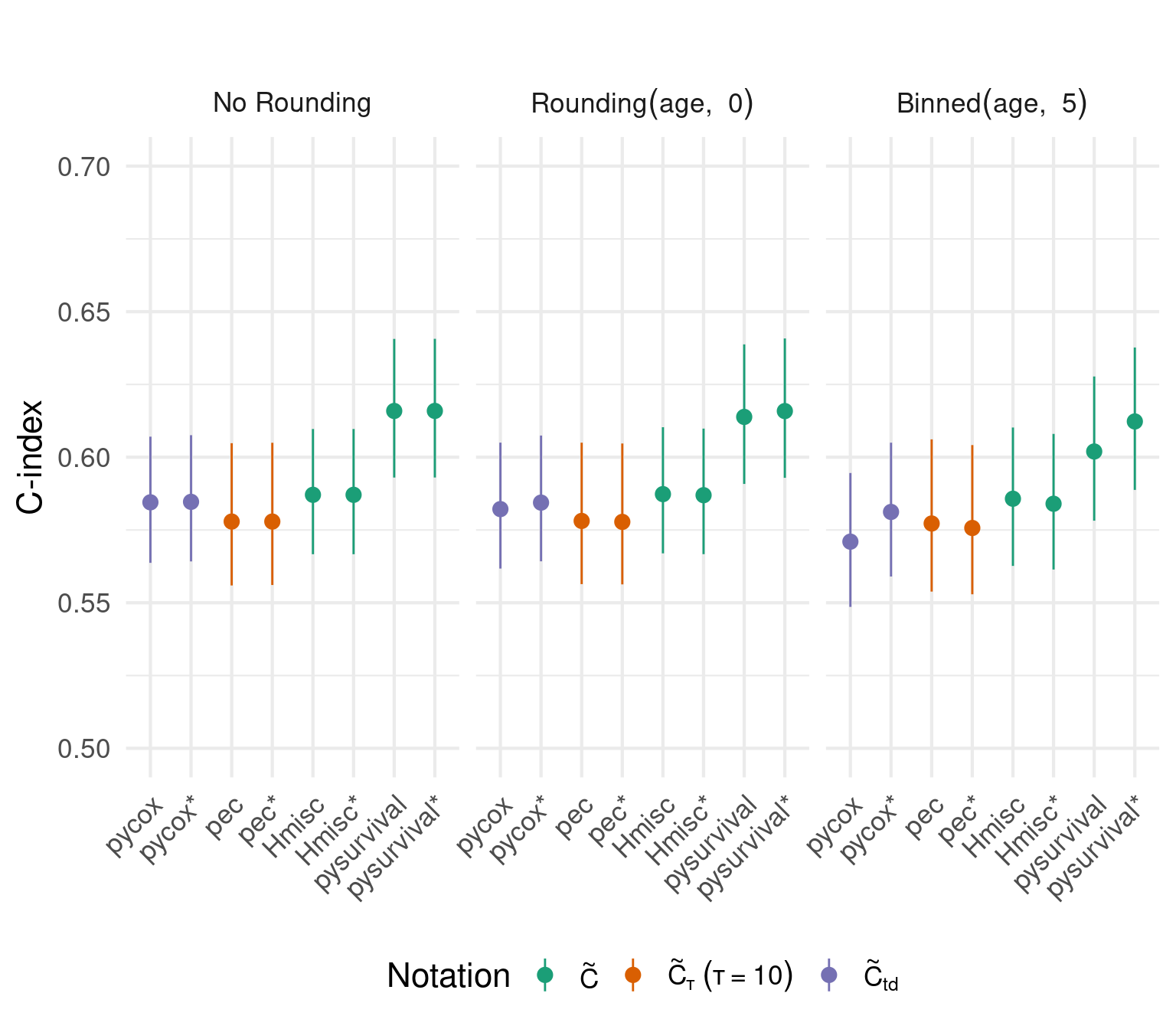}
    \caption{C-index estimates calculated for a simplified CPH model which uses age as the only covariate. We only consider implementations that have different handling options for tied predictions: \pkg{Hmisc} ($C$), \pkg{pysurvival} ($C$), \pkg{pec} ($C_{\tau}$) and \pkg{pycox} ($C_{td}$). These were calculated based on the original METABRIC dataset (no rounding) and after rounding age, to the nearest integer or into 5-year binds. The rounded leads to an increasing number of tied predictions (Supplementary Table S8). Dots represent the C-index estimates and vertical lines the 95\% confidence intervals. Generally, C-index estimates marked with $^*$ include tied predictions, and its absence indicate the exclusion of tied predictions. For \pkg{pec} the parameters are {\tt tiedPredictionsIn = FALSE}, {\tt tiedOutcomeIn = TRUE} and {\tt tiedMatchIn = FALSE}, while for \pkg{pec}$^*$ these parameters are set to {\tt TRUE} as the default. For \pkg{pycox}, the C-index presents estimates based on Antolini's original estimator ({\tt method = "antolini"}; denoted as pycox.Ant in other figures). pycox$^*$, includes more tie predictions ({\tt method = "adj\_antolini"} detailed in Supplementary Table S6; denoted as pycox.Adj.Ant in other figures). The latter is the default setting in \pkg{pycox}. For \pkg{Hmisc}$^*$, default includes tied predictions ({\tt outx = FALSE}), and \pkg{Hmisc} excludes tied predictions ({\tt outx = TRUE}). For \pkg{pysurvival}$^*$, include ties in predictions ({\tt include\_ties = TRUE}) as default, and \pkg{pysurvival} excludes ties ({\tt include\_ties = FALSE}).}
    \label{fig:tiespreds}
\end{figure}

As seen in Supplementary Tables 
S8 and S9, the METABRIC dataset contains a very small number of tied time pairs and predictions. As a result, the inclusion or exclusion of ties led to negligible differences in C-index estimates (Table \ref{tab:tied_times_results}, Figure \ref{fig:tiespreds}). After rounding the survival times, we observed C-index estimates to decrease when tied times were considered as comparable (Table \ref{tab:tied_times_results}). This effect was subtle because, despite the rounding, the proportion of comparable pairs with tied predictions remained small ($<1\%$). In contrast, the inclusion of tied predictions increases the C-index in some implementations (\pkg{pycox}, \pkg{pysurvival}), while slightly decrease in others (\pkg{pec}, \pkg{Hmisc}). These differences are due to discrepancies in how tied predictions affect the calculation of comparable and concordant pairs for each implementation (Supplementary Table S1-S5). 

\begin{figure}[htbp]
    \centering
    \includegraphics[width=15cm]{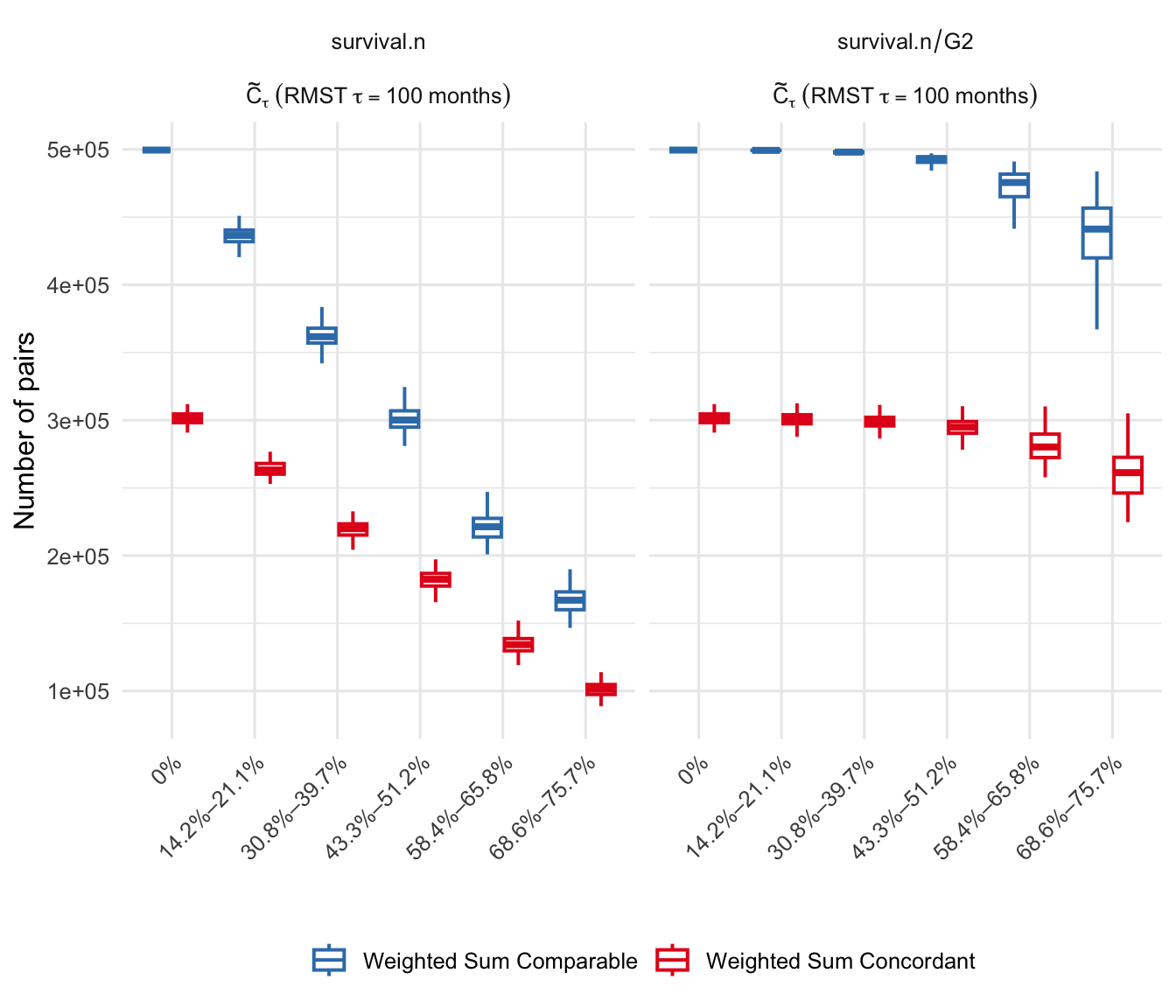}
    \caption{Weighted sum of concordant and comparable pairs across increasing censoring levels (shown as a range in the x-axis), as calculated using the \pkg{survival} R package ($C_{\tau}$). These correspond to the numerator and denominator used to estimate the concordance probability. survival.n indicates uniform weighting ($W_{ij} = 1$). survival.n/G2 indicates that pairs were weighted using IPCW. For each censoring level, boxplots show the distribution across 100 semi-synthetic datasets (with 1000 observations each; see details on Supplementary Note S2). C-index estimates are computed based on RMST as the transformation of choice with $\tau$ = 100 months for $C_{\tau}$.}
    \label{fig:synthetic_ComCon}
\end{figure}

\begin{figure}[htbp]
    \centering
    \includegraphics[width=\textwidth]{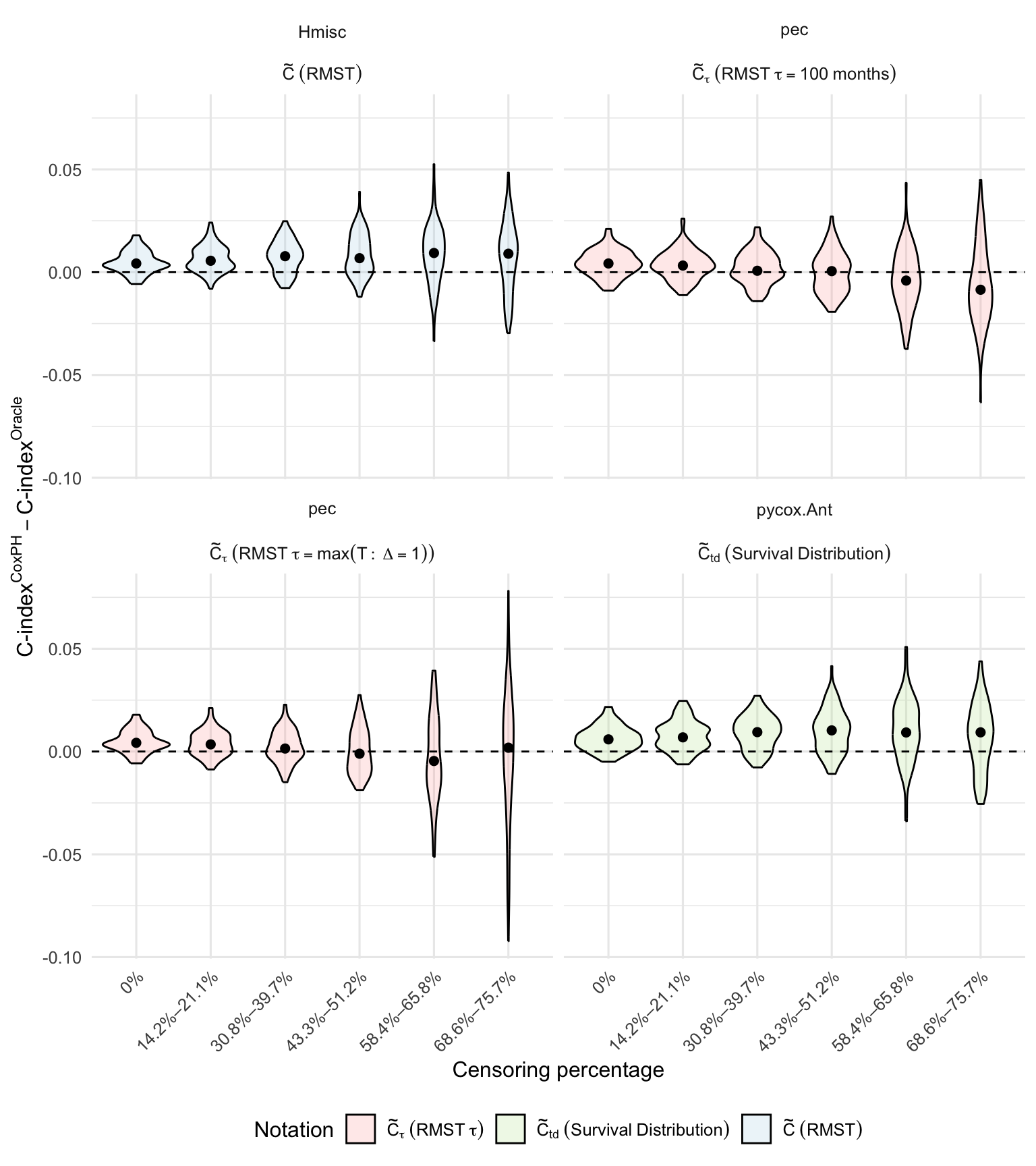}
    \caption{Difference between C-index estimates and their corresponding oracle values (calculated in the absence of censoring, using ground truth model parameters). Violin plots show the distribution across 100 semi-synthetic datasets with increasing levels of censoring (shown as a range in the x-axis). For each dataset, a CPH model was fitted and its performance was evaluated using 5-fold CV. For \pkg{Hmisc} ($C$) and \pkg{pec} ($C_{\tau}$), C-index estimates were computed based on RMST as the transformation of choice. Further details on semi-synthetic data generation and oracle calculation are included in Supplementary Notes S2 and S3.}
    \label{fig:Weibull4panelsdiff}
\end{figure}

As illustrated in Figure \ref{fig:synthetic_ComCon} using semi-synthetic data, 
the number of comparable pairs rapidly declines with increasing censoring levels.   
By re-weighting uncensored observations, 
IPCW stabilise the numerator and denominator used to estimate the concordance probability. 
This led to a bias-variance trade-off. For instance, {\tt Hmisc} and {\tt pycox} ($C$ and $C_{td}$, without IPCW) had a higher bias with respect to oracle values, particularly for high censoring (Figure \ref{fig:Weibull4panelsdiff}). As per \cite{Uno2011}, {\tt pec} ($C_{\tau}$, with IPCW) produced estimates that were, on average, closer to their corresponding oracle across all censoring levels. However, \pkg{pec} led to a wider range of C-index estimates, suggesting poorer stability (Supplementary Figures S10). Similar results were obtained for alternative censoring mechanisms (Supplementary Note S4). As censoring increases, IPCW estimation can be more stable as it is based on the distribution of the censored survival times; but, with fewer events, 
a small number of subjects with high IPCW can more strongly influence in the resultant estimate. Indeed, the spread of C-index estimates was reduced when decreasing $\tau$. Note that, although more subtle, the wider spread of C-index estimates with increasing censoring also occurred for {\tt Hmisc} and {\tt pycox}. This can relate to a greater uncertainty in the predictions (the data contains less information to infer model parameters) or to a reduced number of comparable pairs (which increases uncertainty when estimating the concordance probability).

\newpage



\begin{table}[ht]
\centering
\caption{C-index estimations and 95\% confidence intervals for implementations that have different handling options for tied times ($T_i = T_j$): \pkg{pec} ($C_{\tau}$) and \pkg{pycox} ($C_{td}$). These were calculated based on the METABRIC observed survival times (no rounding) and after rounding the observed times (censored and uncensored) to the nearest integer. For \pkg{pec}, we used $\tau = $ 10 years and RMST as a transformation. For each implementation, two C-index estimates are reported. For \pkg{pec}, the C-index estimates that ignore tied times were obtained by using the following parameters: {\tt tiedPredictionsIn = TRUE}, {\tt tiedOutcomeIn = FALSE} and {\tt tiedMatchIn = FALSE}. Instead, the C-index estimates that include tied times as comparable were obtained using {\tt tiedPredictionsIn = TRUE}, {\tt tiedOutcomeIn = TRUE} and {\tt tiedMatchIn = TRUE}; this is the default in \pkg{pec}. Similarly for \pkg{pycox}, the C-index estimates that exclude ties are based on Antolini's original estimator ({\tt method = "antolini"}; denoted as pycox.Ant in other figures); C-index that includes ties are applying most of the modifications suggested by \cite{Ishwaran2008} ({\tt method = "adj\_antolini"}; denoted as pycox.Adj.Ant in other figures). The latter is the default setting in \pkg{pycox}.}
\centering
\begin{tabular}{llcccc}
\toprule
  Model & Package  & Rounding  & C-index (ties excluded) & C-index (ties included)\\
\midrule
\multirow{4}{*}{CPH} & {\tt pec}  & No & 0.6346 [0.6112, 0.6588] & 0.6346 [0.6113, 0.6589]\\
& {\tt pec} & Yes & 0.6351 [0.6114, 0.6595] & 0.6348 [0.6113, 0.6593]\\
& {\tt pycox} & No & 0.6317 [0.6047, 0.6544] & 0.6318 [0.6053, 0.6548]\\
& {\tt pycox}  & Yes & 0.6298 [0.6018, 0.6519] & 0.6294 [0.6020, 0.6516]\\
\midrule
\multirow{4}{*}{RSF} & {\tt pec}  & No & 0.6395 [0.6170, 0.6643] & 0.6395 [0.6171, 0.6644]\\
& {\tt pec} & Yes & 0.6400 [0.6173, 0.6646] & 0.6397 [0.6172, 0.6644] \\
& {\tt pycox} & No & 0.6587 [0.6341, 0.6861] &  0.6588 [0.6345, 0.6863] \\
& {\tt pycox}  & Yes & 0.6579 [0.6328, 0.6860] & 0.6572 [0.6326, 0.6853] \\
\midrule
\multirow{4}{*}{Cox-Time} & {\tt pec}  & No & 0.6411 [0.6169, 0.6705] & 0.6411 [0.6170, 0.6706]\\
& {\tt pec} & Yes & 0.6413 [0.6169, 0.6708] & 0.6411 [0.6169, 0.6705]\\
& {\tt pycox} & No & 0.6492 [0.6252, 0.6737] & 0.6506 [0.6269, 0.6753] \\
& {\tt pycox}  & Yes & 0.6484 [0.6245, 0.6730] & 0.6488 [0.6256, 0.6735] \\
\midrule
\multirow{4}{*}{DeepSurv} & {\tt pec}  & No & 0.6401 [0.6170, 0.6665] & 0.6401 [0.6172, 0.6667] \\
& {\tt pec} & Yes & 0.6404 [0.6172, 0.6667] & 0.6401 [0.6171, 0.6665]\\
& {\tt pycox} & No & 0.6340 [0.6111, 0.6562] &  0.6344 [0.6119, 0.6566]\\
& {\tt pycox}  & Yes & 0.6327 [0.6095, 0.6552] & 	0.6325 [0.6098, 0.6549]\\
\midrule
\multirow{4}{*}{DeepHit} & {\tt pec}  & No & 0.5912 [0.5682, 0.6161] & 0.5912 [0.5684, 0.6162]\\
& {\tt pec} & Yes & 0.5912 [0.5682, 0.6157] & 0.5910 [0.5681, 0.6157]\\
& {\tt pycox} & No & 0.6601 [0.6386, 0.6891] & 0.6603 [0.6390, 0.6895] \\
& {\tt pycox}  & Yes & 0.6589 [0.6374, 0.6874] & 0.6584 [0.6373, 0.6870]\\
\bottomrule
\end{tabular}
\label{tab:tied_times_results}
\end{table}

\FloatBarrier
\newpage

\section{Discussion}
\label{sec5}

In this work, we examined multiple factors that influence C-index estimation in the context of survival analysis. Our motivation originates from the observation that conceptually equal C-index implementations can lead to substantially different  estimates when applied to the same data.  

In terms of model comparison, our results are consistent with those of \cite{sonabend2022avoiding}: model ranking varies depending on the definition of 
\addition{concordance} 
(i.e.~$C$, $C_{\tau}$ or $C_{td}$), and the choice of input used to rank individuals (e.g.~time-dependent risk or RMST) --- leading to potential \emph{C-hacking} (e.g.~authors selectively reporting C-index results to favor a desired conclusion). We extend these findings by highlighting  discrepancies between existing software and published estimators \citep[e.g.][]{Harrell1982, Uno2011, antolini2005}. \review{Such differences are often not apparent from the software's documentation, leaving users with a potentially overwhelming \emph{C-index multiverse}, e.g.~
due to differences in censoring adjustment, tie handling or truncation. Supplementary analysis incorporating more recent software, such as \pkg{SurvivalEVAL}, further illustrates this complexity, as it permits combinations of these choices that do not correspond to estimators previously defined in the literature (i.e., Harrell's and Antolini's with $\tau$ truncation).} 
Our work will help users to navigate the multiverse,  providing unified documentation  and highlighting potential pitfalls of existing software.

For $C$ and $C_{\tau}$ estimators, the reduction of the survival distribution into a single measure of risk is not trivial. Different transformations capture distinct aspects of the survival distribution.  For instance, time-dependent predictions capture the probability of a subject to experience the event up to a specific time $t$, but the ranking of individuals is not the same across different values of $t$ (unless a PH assumption holds).
Instead, expected mortality integrates information across time but weights later time-points more heavily due to the logarithmic transformation. This leads to an unstable ranking of individuals, 
as survival probabilities have higher uncertainty at later time points 
and 
because survival probabilities approaching zero result in numerical instabilities that can accentuate minor differences in the tail of the distribution. 
As an alternative, we propose the restricted mean survival time (RMST), \review{which was previously suggested as} an outcome measure for clinical trials when the PH assumption does not hold \citep{royston2013restricted, Averbuch2025}. 
\review{Its interpretation is explicitly linked to a pre-specified time horizon, corresponding to the area under the survival curve up to that horizon (e.g.~10 years). While this provides a well defined statistical summary, its interpretation for individual-level prediction may be less straightforward for patients or clinicians than, for example, a predicted survival probability at a specific time point (i.e.~$t$-year risk).}

$C_{td}$ estimators bypass the need for a transformation, as the ranking of individuals is 
directly based on the survival distribution: 
for a pair ($i,j$)  
this is based on the survival distribution evaluated at subject $i$'s event time ($T_i$). As such, $C_{td}$ may 
emphasise differences in survival probabilities for periods with a higher event frequency \review{(Supplementary Figure \ref{fig:supp_ct_distrib})}. 
 Our analysis of the METABRIC data suggests this can have important consequences, leading to conflicting results with respect to $C$ and $C_{\tau}$. 
 This was particularly the case when evaluating the performance of DeepHit which was selected as the top model by $C_{td}$ but the worst for $C$ and $C_{\tau}$. This result is not entirely unexpected, as DeepHit's loss function encourages the correct ranking of individuals 
as in $C_{td}$. Moreover, available $C_{td}$ implementations do not account for censoring, resulting in potentially biased estimates   \citep[recent work by][addresses this limitation]{faruk2025}. 

Beyond the choice between $C$, $C_{\tau}$ and $C_{td}$ (and associated transformations), the use of IPCW to adjust for  censoring also affects C-index estimates. As demonstrated by our simulations, this leads to a bias-variance trade-off: IPCW can reduce bias in C-index estimates but potentially increase the variance, particularly when the rate of censoring is high. 
Our results also emphasise earlier observations by \cite{Uno2011}, highlighting the importance of a time truncation (as controlled by $\tau$) to avoid exploding IPCW. Despite this, existing software does 
\addition{not always} alert users, who may inadvertently calculate unstable estimates. This is particularly problematic for {\tt pysurvival}, which includes IPCW but were time truncation is not an option. 

C-index estimates can also be sensitive to the inclusion of tied times or predictions as comparable and concordant pairs. However, the effect can be negligible if the number of ties is small (as in METABRIC). For datasets with a large number of ties, users may perform a sensitivity analysis and potentially report an overall concordance metric \citep{yan2008investigating}. As demonstrated by \cite{alabdallah2024concordance}, decompositions such as the one in Equation \eqref{eq:generalCties_decomposition} can also be used to inform model training, by targeting different types of observations (note that \citeauthor{alabdallah2024concordance} separated pairs based on censoring status, not with respect to ties).

In cases where models provide different outputs (e.g.~predicted survival distribution with a varying time resolution), 
the harmonization of C-index inputs is essential for fair model comparison, but it is often unreported. Furthermore, as demonstrated by the METABRIC analysis, model rankings can be sensitive to specific data splits for moderate sample sizes. Despite this, a single data split is often used to showcase the performance of new survival models --- which may also lead to C-hacking. In such cases, resampling strategies (e.g.~cross-validation and bootstrapping) can be used to more robustly evaluate predictive performance \citep{collins2024evaluation}. Furthermore, the C-index multiverse extends to competing risks cases, where multiple mutually exclusive event types may occur \citep{van2022validation}.

Our results also emphasise the importance of using the same C-index implementation when comparing different survival models. This is the case, even for implementations estimate the same concordance probability ($C$, $C_{\tau}$ and $C_{td}$), as estimates may vary due to subtle differences in e.g.~tie handling and censoring adjustment. 
When fitting survival models using different software (e.g.~R versus \review{Python}), tools such as {\tt reticulate} \citep{reticulate} may facilitate this process. 
Differences in C-index implementations also impose an important challenge for meta-analyses, whose aim is to  summarize discrimination performance across multiple published analyses \citep{debray2019framework}.
Whilst meta-analyses methods can account for inter-study heterogeneity (e.g.~underlying different populations), aggregating C-index across studies without accounting for implementation differences could lead to invalid conclusions.


%


\cite{blanche2019} noted that the C-index estimated using time-dependent predictions as input is not a proper scoring rule to evaluate $t$-year predictions, as it ranks individuals based on predicted risk at time $t$ regardless of whether $T_i$ or $T_j$ occurred before or after $t$. 
\addition{This issue is particularly relevant for non PH models, such as RSF.} The $t$-year area under the receiver operating characteristic curve $\text{AUC}_{t}$ was recommended as an alternative metric that bypasses this issue \citep{blanche2019}. However, as with the C-index, several definitions and implementations also exist \citep{kamarudin2017time}. 

\review{Lack of propriety is not restricted to $C$ and $C_{\tau}$ estimates based on $t$-year predictions. \cite{rindt22aPMLR} showed that $C_{td}$ is also not a proper scoring rule for evaluating predicted survival distributions: an incorrect survival distribution can achieve a higher expected $C_{td}$ than the true survival distribution.} \addition{Recent variants have been proposed specifically to address propriety when evaluating predicted survival distributions \citep{gandy2025SJS}.}
Importantly, regardless of the implementation or the transformation applied to the survival function, 
the C-index (and AUC$_t$) 
fails to capture the magnitude of the predicted difference between subjects. 
This 
highlights the need for alternative metrics that can asses calibration and clinical utility, which is essential in healthcare contexts \citep{steyerberg2010assessing}. 
A clear example 
is DeepHit, which was selected as the top performing model based on $C_{td}$, but tends to predict narrowly distributed survival probabilities, suggesting lack of calibration.  

It is important to note that, for the METABRIC analysis, we selected the same covariates as in 
\cite{deepsurv2018}. Such set has been also 
adopted 
in subsequent studies 
\citep[e.g.~][]{kvamme2019time,deephit2008}, partly facilitated by the METABRIC extract 
available in the \pkg{pycox} package. Our choice ensured comparability between existing benchmarks, and the re-use of tuned hyperparameters for deep learning based approaches. 
However, from a clinical perspective, the inclusion of post-diagnoses treatment variables (chemotherapy, radiotherapy and hormone therapy) can introduce 
immortal time bias \citep[i.e.~patients have to survive long enough to receive treatment,][]{yadav2021Jama}. 
Therefore, while our set of covariates is appropriate for technical benchmarks, 
the trained models themselves should be interpreted with caution in clinical context. Indeed, the inclusion of treatment information outwidth a randomised clinical trial setting can lead to causal blindspots \citep{vanGeloven2025Risks}.

Regardless of the challenges discussed above, the C-index is one of the most popular prediction performance for time-to-event outcomes. Indeed, most readers and reviewers would expect to have it reported. This article highlights the existence of a C-index multiverse, where seemingly equal implementations can led to different estimates and, in the context of model selection, to different model rankings. Our work aims to help prospective users navigate available software options and understand limitations in their use. To conclude, if we could summarise our paper into one sentence, it would be that \emph{it's not enough to say: ``we calculated the C-index".}

    \vspace{0.4em}
    \begin{tcolorbox}[
        title=\textbf{Box 1: Key learnings and practical recommendations},
        colback=gray!5,
        colframe=black,
        boxrule=0.6pt,
        arc=2mm,
    ]
    
    \textbf{There is no such thing as ``the C-index".}
    Different C-index definitions and implementations 
    differ in multiple ways; for example, how observation pairs are 
    labelled as
    comparable or concordant, 
    the handling of censoring and ties, and how model predictions are used to rank observations. Reporting a ``C-index'' without specifying how it was calculated is insufficient to reproduce or interpret a result. 

    \vspace{0.4em}
    
    \textbf{C-index choices must be consistent across models.}
    When comparing discrimination performance across different models, a consistent definition of concordance and 
    estimation procedure should be used. 
    For models fitted using different software or programming languages (e.g.,  R and Python), predictions can be evaluated using a common C-index implementation, either in a common environment or by calling the same implementation across languages.
    
    \vspace{0.4em}
    
    \textbf{Consider what model outputs are used to assess concordance.}
    Whilst $C$ and $C_{\tau}$ transform the predicted survival distribution into a one-dimensional risk score, $C_{td}$ determines the ranking from the predicted survival probabilities. 
    The choice 
    is particularly critical for non-proportional hazards models, as different transformations can produce different rankings and, therefore, C-index estimates. 

    \vspace{0.4em}
    
    \textbf{Consider how the evaluation setting affects C-index choices.}
    Increasing censoring makes censoring adjustment more influential, IPCW adjustment makes assumptions about the censoring mechanism and requires time truncation, and frequent ties make their handling more consequential. 
    These considerations highlight which choices require particular attention in a given setting, rather than favouring one C-index estimator universally.
    \vspace{0.4em}

    \textbf{Reconsider whether the C-index is consistent with the evaluation objective.} For discrimination at a specific prediction horizon, $\text{AUC}_{t}$ may be more appropriate. Furthermore measures of calibration (i.e., a model may correctly rank patients, while providing inaccurate absolute risk predictions), or clinical utility may be preferable in clinical settings. 
    
    \end{tcolorbox}

\section{Software and reproducibility}\label{sec:softwareimple}

The following R packages were used to fit time-to-event models: \pkg{survival} version 3.8-3 (CPH), \pkg{randomForestSRC} version 3.3.3 (RSF), \pkg{survivalmodels} version 0.1.19 (DeepHit, DeepSurv and Cox-Time). To reproduce the results of this study, the code is publicly available at \url{www.github.com/BBolosSierra/CindexMultiverse}. The corresponding Docker image, also provided in the repository, contains the environment (i.e R and \review{Python} packages and dependencies) required to reproduce the results.

\section*{Acknowledgments}

The authors thank Dr. Nathan Constantine-Cooke for constructive comments during the project development, and \addition{John Zobolas for his thorough review and valuable input on the manuscript.}
Funding for the project was provided by Cancer Research UK Scotland Centre (CTRQQR\-2021\text{/}100006).
BBS was supported by The Alan Turing Institute’s Enrichment scheme. 
{\it Conflict of Interest}: None declared.

\newpage

\bibliographystyle{biorefs}
\bibliography{refs}
\newpage
\section*{Supplementary Material}
\addcontentsline{toc}{section}{Supplementary Material}

\setcounter{section}{0}
\setcounter{figure}{0}
\setcounter{table}{0}
\setcounter{equation}{0}
\renewcommand{\thefigure}{S\arabic{figure}}
\renewcommand{\thesection}{S\arabic{section}}
\renewcommand{\thesubsection}{S\arabic{section}.\arabic{subsection}}
\renewcommand{\thetable}{S\arabic{table}}


\setcounter{section}{0}
\setcounter{figure}{0}
\setcounter{table}{0}
\renewcommand{\thesection}{S\arabic{section}}
\renewcommand{\thefigure}{S\arabic{figure}}
\renewcommand{\thetable}{S\arabic{table}}


\maketitle

\section{Interpolation of survival probabilities into a common time grid $\mathcal{T}$} \label{supsec:interpolation}

The definition of expected mortality and RMST as potential transformations to summarise the survival distribution are based on a sum over a time grid $\mathcal{T}$ (Section 4.3). Lack of a common grid can preclude meaningful results when comparing discrimination performance for different survival models whose implementations associated do not return survival probabilities at the same grid of time-points $\mathcal{T}$. For instance, when fitting a CPH \citep{Cox1972} model using the {\tt survival} R package \citep{survival-package}, survival probabilities are evaluated at the uniquely observed survival times in the training set ($\approx 1400$ per fold in the case of the METABRIC analysis; see Section 6). Instead, the time grid resolution of Cox-Time \citep{kvamme2019time} depends on a random subset of the training data (this is to reduce computational costs). Others allow users to specify the number of time-points in the grid. This is the case for {\tt randomForestSRC} \citep[which can be used to fit a RSF;][]{randomForestSRC} and DeepHit \citep{deephit2008}. 

For the METABRIC analysis (Section 6), to ensure a homogeneous time horizon across models, the time grid for RSF and DeepHit was fixed to 1400 points, approximately matching the number of unique event times in each training set per fold. Additionally, survival probabilities were linearly interpolated over a common evaluation grid defined as $\mathcal{T} = \{0, 1, \ldots, 355\}$. 

\section{Semi-synthetic data generation}
\label{supsec:datageneration}

Here, we describe how each semi-synthetic dataset (Section 6) was generated based on METABRIC. A total of 100 datasets with $n = 1000$ individuals were generated for each simulation setting. All simulation settings shared the same data generation process for the uncensored survival times $\tilde{T}_i$, but differed in terms of how censoring times $C_i$ were simulated. 

\subsection{Generation of uncensored survival times $\tilde{T}_i$}

Firstly, a random subset of $n = 1000$ individuals was extracted from METABRIC. The corresponding covariate values (five clinical features and four gene indicators) were then used to generated uncensored survival times from a Weibull PH model. Model parameters (regression coefficients and shape parameter) were set to match those inferred for the METABRIC subsample. These were obtained using the {\tt survreg} function from the {\tt survival} R package \citep{survival-package}.  

Note that {\tt survreg} estimates a Weibull Accelerated Failure Time (AFT) model: \begin{equation}
  \log (T_i) = \mu + \mathbf{x}_i^T \alpha  + \sigma W_i,  
\end{equation}
where $\mu$ is an intercept, $\alpha$ is a vector of the regression coefficients, $\mathbf{x}_i$ is a vector of observed covariate values, $\sigma$ is a scale parameter, and $W_i$ is a random variable that follows a extreme value distribution. The later can be re-parametrised in terms of a PH specification such that \begin{equation}
h(t | \mathbf{x}_i) = \gamma t^{\gamma -1} \lambda e^{\mathbf{x}_i^T \beta }, \hspace{0.1cm} \gamma = \frac{1}{\sigma}, \hspace{0.1cm}
\lambda = e^{-\frac{\mu}{\gamma}}, \hspace{0.1cm} \text{\hspace{0.1cm}} \hspace{0.1cm}
\beta = -\frac{\alpha}{\sigma}.  
\end{equation}

Finally, uncensored survival times were generated as follows: i.e.\begin{equation} \label{suppeq::UncensoredTimesGeneration}
    U_i \sim \text{Uniform(0,1)}, \hspace{1cm} \tilde{T}_i =  \left[\frac{-\log(U_i)}{\lambda e^{\mathbf{x}_i^T \beta }}\right]^{1/\gamma}.
\end{equation}

\subsection{Generation of censoring times $C_i$}






Censoring times $C_i$ follow a Weibull distribution parameterised such that:
\begin{equation} \label{supp:eq:weibullcens}
h_C(t) = \gamma_C t^{\gamma_C -1} \epsilon \lambda_C.
\end{equation}

The value of the scaling factor $\epsilon \geq0$ was varied to control the level of censoring (increasing values of $\epsilon$ increases the amount of censoring in the generated dataset). We used $\epsilon = \{0, 0.5, 1, 3, 7, 13\}$.

\section{Oracle C-index calculation}

For each semi-synthetic dataset, an oracle C-index value was calculated based on the uncensored survival times $\tilde{T}_i$ and the true parameter values used to generate them (Supplementary Note \ref{supsec:datageneration}). When calculating C-index estimates, 
survival probabilities of each patient in test set were calculated based on the Weibull PH model, given by: \begin{equation}
S(t|\mathbf{x}_i) = \exp\{-\lambda t^{\gamma} e^{\mathbf{x}_i^T \beta} \}. \end{equation} Survival probabilities are computed over a discrete grid $t \in \{0,1,2,...,T^*\}$, where $T^*$ corresponds to the maximum observed time in the test set. Note that the true parameter values ($\lambda$, $\gamma$ and $\beta$) are dataset-dependent, as these where calculated based on different random sub-samples from METABRIC. However, oracle values 
are invariant to the censoring mechanism or censoring levels. This is because 
all C-index estimates were calculated based on the uncensored event times $\tilde{T}_i$. 

\section{Semi-synthetic data with alternative censoring mechanisms}

As a sensitivity analysis, we considered two alternative censoring mechanisms to generate $C_i$: (i) from a Weibull PH model using age as a covariate and (ii) from a uniform distribution. Results for both censoring mechanisms are displayed in Supplementary Figures S11 and S12. 

\subsubsection*{Case (i): Age-informed Weibull PH.} Censoring times $C_i$ were generated from a Weibull PH model using age as a covariate.  
As in Supplementary Note S2, parameters were calculated based ob censored times observed for the corresponding METABRIC subsample. These were obtained from a re-parametrized AFT model fitted with the {\tt survreg} function from the {\tt survival} R package \citep{survival-package}. $\gamma_c$, $\lambda_c$ and $\epsilon$ were set as in Supplementary Note S2. 

\subsubsection*{Case (ii): Uniform.}
$C_i$ follows a Uniform distribution such that $C_i \sim U(C_{\min
},C_{\max})$, where $C_{\min} = \min(\tilde{T}_i)$.
To control the level of censoring, the maximum censoring time $C_{\max}$ was defined using the quantiles of the event times (uncensored) distribution, i.e. $C_{\max} = Q_{1 - \epsilon}(\tilde{T_i})$, where $Q_{1 - \epsilon}(\cdot)$ represents the $(1-\epsilon)*100$ quantile. We considered $\epsilon \in \{ 0, 0.1, 0.2, 0.3, 0.4, 0.5 \}$. Increasing values of $\epsilon$ led to increasing amount of censoring in the generated dataset. 

\newpage

\section{Supplementary Tables}

\begin{table}[htbp]
\centering
\renewcommand{\arraystretch}{1.6}
\begin{threeparttable}
\small
\setlength{\tabcolsep}{3pt}
\begin{tabular}{cccc|*{5}{cc}@{\hspace{7pt}}|>{\hspace{3pt}}ll}
\multicolumn{4}{r|}{} &
\multicolumn{2}{c}{\rotatebox{90}{Hmisc}} &
\multicolumn{2}{c}{\rotatebox{90}{Survmetrics}} &
\multicolumn{2}{c}{\rotatebox{90}{lifelines}} &
\multicolumn{2}{c}{\rotatebox{90}{pysurvival}} &
\multicolumn{2}{c|}{\rotatebox{90}{sksurv (cens.)}} &
\multicolumn{2}{c}{\review{\shortstack{SurvivalEVAL\\(Harrell / Naive)}}} \\
\cmidrule(lr){5-6}
\cmidrule(lr){7-8}
\cmidrule(lr){9-10}
\cmidrule(lr){11-12}
\cmidrule(lr){13-14}
\cmidrule(lr){15-16}
\textbf{Case} & \textbf{Time} & \textbf{($\Delta_i, \Delta_j$)} & \textbf{Predictions} & CP & CN & CP & CN & CP & CN & CP & CN & CP & CN & \hspace{8pt} CP & CN \\
\midrule
1A & $T_i < T_j$ & (1,1) & $M(\mathbf{x}_i)>M(\mathbf{x}_j)$ & \ding{51} & 1 & \ding{51} & 1 & \ding{51} & 1 & \ding{51} & 1 & \ding{51} & 1 & N/T/R/A: \ding{51} & N/T/R/A: 1 \\
1B & $T_i < T_j$ & (1,1) & $M(\mathbf{x}_i)<M(\mathbf{x}_j)$ &  \ding{51} & 0 & \ding{51} & 0 & \ding{51} & 0 & \ding{51} & 0 & \ding{51} & 0 & N/T/R/A: \ding{51} & N/T/R/A: 0 \\
1C & $T_i < T_j$ & (1,1) & $M(\mathbf{x}_i)=M(\mathbf{x}_j)$ &  \ding{51}\tnote{(1)}  & 0.5& \ding{51} & 0.5 & \ding{51} & 0.5 & \ding{51} & 0.5\tnote{(1)} & \ding{51} & 0.5\tnote{(*)} & N/T: \ding{55}; R/A: \ding{51} & N/T: --; R/A: 0.5 \\
\midrule
2A & $T_i < T_j$ & (1,0) & $M(\mathbf{x}_i)>M(\mathbf{x}_j)$ & \ding{51} & 1 & \ding{51} & 1 & \ding{51} & 1 & \ding{51} & 1 & \ding{51} & 1 & N/T/R/A: \ding{51} & N/T/R/A: 1 \\
2B & $T_i < T_j$ & (1,0) & $M(\mathbf{x}_i)<M(\mathbf{x}_j)$  & \ding{51} & 0 & \ding{51} & 0 & \ding{51} & 0 & \ding{51} & 0 & \ding{51} & 0 & N/T/R/A: \ding{51} & N/T/R/A: 0 \\
2C & $T_i < T_j$ & (1,0) & $M(\mathbf{x}_i)=M(\mathbf{x}_j)$ & \ding{51}\tnote{(1)}  & 0.5 & \ding{51} & 0.5 & \ding{51} & 0.5 & \ding{51} & 0.5\tnote{(1)} & \ding{51} & 0.5\tnote{(*)} & N/T: \ding{55}; R/A: \ding{51} & N/T: --; R/A: 0.5 \\
\midrule
3 & $T_i < T_j$ & (0,1) & Any & \ding{55} & - & \ding{55} & - & \ding{55} & - & \ding{55} & - & \ding{55} & - & N/T/R/A: \ding{55} & N/T/R/A: -- \\
\midrule
4 & $T_i < T_j$ & (0,0) & Any & \ding{55} & - & \ding{55} & - & \ding{55} & - & \ding{55} & - & \ding{55} & - & N/T/R/A: \ding{55} & N/T/R/A: -- \\
\midrule
5A & $T_i = T_j$ & (1,1) & $M(\mathbf{x}_i)>M(\mathbf{x}_j)$ & \ding{55} & - & \ding{51} & 0.5 & \ding{55} & - & \ding{55} & - & \ding{55} & - & N/R: \ding{55}; T/A: \ding{51} & N/R: --; T/A: 0.5 \\
5B & $T_i = T_j$ & (1,1) & $M(\mathbf{x}_i)<M(\mathbf{x}_j)$ & \ding{55} & - & \ding{51} & 0.5 & \ding{55} & - & \ding{55} & - & \ding{55} & - & N/R: \ding{55}; T/A: \ding{51} & N/R: --; T/A: 0.5 \\
5C & $T_i = T_j$ & (1,1) & $M(\mathbf{x}_i)=M(\mathbf{x}_j)$ & \ding{55} & - & \ding{51} & 1 & \ding{55} & - & \ding{55} & - & \ding{55} & - & N/R: \ding{55}; T/A: \ding{51} & N/R: --; T/A: 0.5 \\
\midrule
6A & $T_i = T_j$ & (1,0) & $M(\mathbf{x}_i)>M(\mathbf{x}_j)$ & \ding{51} & 1 & \ding{51} & 1 & \ding{51} & 1 & \ding{51} & 1 & \ding{51} & 1 & N/T/R/A: \ding{51} & N/T/R/A: 1 \\
6B & $T_i = T_j$ & (1,0) & $M(\mathbf{x}_i)<M(\mathbf{x}_j)$  & \ding{51} & 0 & \ding{51} & 0.5 & \ding{51} & 0 & \ding{51} & 0 & \ding{51} & 0 & N/T/R/A: \ding{51} & N/T/R/A: 0 \\
6C & $T_i = T_j$ & (1,0) & $M(\mathbf{x}_i)=M(\mathbf{x}_j)$ & \ding{51}\tnote{(1)} & 0.5 & \ding{51} & 0.5 & \ding{51} & 0.5 & \ding{51} & 0.5\tnote{(1)} & \ding{51} & 0.5\tnote{(*)} & N/T: \ding{55}; R/A: \ding{51} & N/T: --; R/A: 0.5 \\
\midrule
7 & $T_i = T_j$ & (0,0) & Any & \ding{55} & - & \ding{55} & - & \ding{55} & - & \ding{55} & - & \ding{55} & - & N/T/R/A: \ding{55} & N/T/R/A: -- \\
\bottomrule
\end{tabular}
\end{threeparttable}
\vspace{6pt}
\caption{{\bf Definition of comparable and concordant pairs by different software implementations that estimate $C$.} For each pair $(i,j)$, different cases are defined based on the observed survival times ($T_i$, $T_j$), the corresponding censoring indicators ($\Delta_i$, $\Delta_j$) and model outputs ($M(\mathbf{x}_i)$, $M(\mathbf{x}_j)$). $M(\cdot)$ refers to a scalar risk score (e.g. RMST or expected mortality). The table summarises the inclusion (\ding{51}) or exclusion (\ding{55}) of a pair $(i,j)$ as comparable (CP). When tied event times ($T_i$, $T_j$) are included, a weight $\omega_o = 0.5$ is applied (see (4.6). For comparable pairs, concordant (CN) status is summarised using a weight: 0 if not concordant, 1 if concordant. Implementations where tied predictions ($M(\mathbf{x}_i) = M(\mathbf{x}_j)$) can be concordant use a 0.5 weight ($\omega_p$ in (4.6))). Implementation specific behaviors determined by a user-defined parameter include: (1) Default inclusion of tied predictions (e.g \pkg{Hmisc} when {\tt outx = FALSE}, \pkg{pysurvival} if { \tt include\_ties = TRUE}). (*) The definition of tied predictions is determined based on a user-defined tolerance parameter (e.g \pkg{sksurv} where $\epsilon = 10^{-8}$ (referred to as {\tt tied\_tol}, such that predictions are considered tied if $|M(\mathbf{x}_i) - M(\mathbf{x}_j)| \leq  \epsilon$). 
Finally, note that \pkg{pysurvival} always yields higher or equal than 0.5, since there is a condition where the estimated C-index is calculated as max($\hat C$ , $ 1-\hat C$). \pkg{pysurvival}'s adjustment for censoring is included in Supplementary Table S4. \review{ \pkg{SurvivalEVAL} version 0.8.0 ({\tt method = "Harrell"} and { \tt method = "Naive"} ) have user-defined tie-handling options; N = \texttt{None}, T = \texttt{Time}, R = \texttt{Risk}, and A = \texttt{All}, where the default is  { \tt ties = "Risk"} that accounts for ties with a established tie tolerance of 1e-8. Note that \pkg{SurvivalEVAL}  also supports optional truncation at $\tau$ (i.e. as $C_{\tau}$ without censoring adjustment), retaining pairs where $T_i<\tau$ for subject $i$; this table presents the untruncated case.}
}
\label{supp:tableC}
\end{table}

\begin{table}[htbp]
\centering
\renewcommand{\arraystretch}{1.6}
\begin{threeparttable}
\small
\setlength{\tabcolsep}{1.55pt}
\begin{tabular}{llll|*{5}{cc}|cc}
\multicolumn{4}{r|}{} &
\multicolumn{2}{c}{\rotatebox{90}{pec}} &
\multicolumn{2}{c}{\rotatebox{90}{survC1}} &
\multicolumn{2}{c}{\rotatebox{90}{survival ("n")}} &
\multicolumn{2}{c}{\rotatebox{90}{survival ("n/G2")}} &
\multicolumn{2}{c|}{\rotatebox{90}{sksurv (ipcw)}} &
\multicolumn{2}{c}{\review{\shortstack{SurvivalEVAL (Uno/IPCW)}}} \\
\cmidrule(lr){5-6}
\cmidrule(lr){7-8}
\cmidrule(lr){9-10}
\cmidrule(lr){11-12}
\cmidrule(lr){13-14}
\cmidrule(lr){15-16}
\textbf{Case} & \textbf{Time} & \textbf{($\Delta_i, \Delta_j$)} & \textbf{Rank} &
CP & CN & CP & CN & CP & CN & CP & CN & CP & CN & CP & CN \\
\midrule
1A & $T_i < T_j$ & (1,1) & $M(\mathbf{x}_i)>M(\mathbf{x}_j)$ & \ding{51} & 1 & \ding{51} & 1 & \ding{51} & 1 & \ding{51} & 1 & \ding{51} & 1 & N/T/R/A: \ding{51} & N/T/R/A: 1 \\
1B & $T_i < T_j$ & (1,1) & $M(\mathbf{x}_i)<M(\mathbf{x}_j)$ & \ding{51} & 0 & \ding{51} & 0 & \ding{51} & 0 & \ding{51} & 0 & \ding{51} & 0 & N/T/R/A: \ding{51} & N/T/R/A: 0 \\
1C & $T_i < T_j$ & (1,1) & $M(\mathbf{x}_i)=M(\mathbf{x}_j)$ & \ding{51}\tnote{(1)} & 0.5 & \ding{51} & 0.5 & \ding{51} & 0.5 & \ding{51} & 0.5 & \ding{51} & 0.5 & N/T: \ding{55}; R/A: \ding{51} & N/T: --; R/A: 0.5 \\
\midrule
2A & $T_i < T_j$ & (1,0) & $M(\mathbf{x}_i)>M(\mathbf{x}_j)$ & \ding{51} & 1 & \ding{51} & 1 & \ding{51} & 1 & \ding{51} & 1 & \ding{51} & 1 & N/T/R/A: \ding{51} & N/T/R/A: 1 \\
2B & $T_i < T_j$ & (1,0) & $M(\mathbf{x}_i)<M(\mathbf{x}_j)$ & \ding{51} & 0 & \ding{51} & 0 & \ding{51} & 0 & \ding{51} & 0 & \ding{51} & 0 & N/T/R/A: \ding{51} & N/T/R/A: 0 \\
2C & $T_i < T_j$ & (1,0) & $M(\mathbf{x}_i)=M(\mathbf{x}_j)$ & \ding{51}\tnote{(1)} & 0.5 & \ding{51} & 1 & \ding{51} & 0.5 & \ding{51} & 0.5 & \ding{51} & 0.5 & N/T: \ding{55}; R/A: \ding{51} & N/T: --; R/A: 0.5 \\
\midrule
3 & $T_i < T_j$ & (0,1) & Any & \ding{55} & - & \ding{55} & - & \ding{55} & - & \ding{55} & - & \ding{55} & - & N/T/R/A: \ding{55} & N/T/R/A: -- \\
\midrule
4 & $T_i < T_j$ & (0,0) & Any & \ding{55} & - & \ding{55} & - & \ding{55} & - & \ding{55} & - & \ding{55} & - & N/T/R/A: \ding{55} & N/T/R/A: -- \\
\midrule
5A & $T_i = T_j$ & (1,1) & $M(\mathbf{x}_i)>M(\mathbf{x}_j)$ & \ding{51}\tnote{(2)} & 1 & \ding{55} & - & \ding{55} & - & \ding{55} & - & \ding{55} & - & N/R: \ding{55}; T/A: \ding{51} & N/R: --; T/A: 0.5 \\
5B & $T_i = T_j$ & (1,1) & $M(\mathbf{x}_i)<M(\mathbf{x}_j)$ & \ding{51}\tnote{(2)} & 0 & \ding{55} & - & \ding{55} & - & \ding{55} & - & \ding{55} & - & N/R: \ding{55}; T/A: \ding{51} & N/R: --; T/A: 0.5 \\
5C & $T_i = T_j$ & (1,1) & $M(\mathbf{x}_i)=M(\mathbf{x}_j)$ & \ding{51}\tnote{(2)} & 0.5/1\tnote{(3)} & \ding{55} & - & \ding{55} & - & \ding{55} & - & \ding{55} & - & N/R: \ding{55}; T/A: \ding{51} & N/R: --; T/A: 0.5 \\
\midrule
6A & $T_i = T_j$ & (1,0) & $M(\mathbf{x}_i)>M(\mathbf{x}_j)$ & \ding{51} & 1 & \ding{55} & - & \ding{51} & 1 & \ding{51} & 1 & \ding{51} & 1 & N/T/R/A: \ding{51} & N/T/R/A: 1 \\
6B & $T_i = T_j$ & (1,0) & $M(\mathbf{x}_i)<M(\mathbf{x}_j)$ & \ding{51} & 0 & \ding{55} & - & \ding{51} & 0 & \ding{51} & 0 & \ding{51} & 0 & N/T/R/A: \ding{51} & N/T/R/A: 0 \\
6C & $T_i = T_j$ & (1,0) & $M(\mathbf{x}_i)=M(\mathbf{x}_j)$ & \ding{51}\tnote{(1)} & 0.5\tnote{(3)} & \ding{55} & - & \ding{51} & 0.5 & \ding{51} & 0.5 & \ding{51} & 0.5 & N/T: \ding{55}; R/A: \ding{51} & N/T: --; R/A: 0.5 \\
\midrule
7 & $T_i = T_j$ & (0,0) & Any & \ding{55} & - & \ding{55} & - & \ding{55} & - & \ding{55} & - & \ding{55} & - & N/T/R/A: \ding{55} & N/T/R/A: -- \\
\bottomrule
\end{tabular}
\end{threeparttable}
\vspace{6pt}
\caption{{\bf Definition of comparable and concordant pairs across different software implementations used to estimate the concordance index $C_{\tau}$}. For each pair of individuals $(i,j)$, different conditions are defined based on observed survival times ($T_i$, $T_j$), censoring indicators ($\Delta_i$, $\Delta_j$), and model outputs ($M(\mathbf{x}_i)$, $M(\mathbf{x}_j)$), where $M(\cdot)$ denotes a scalar-valued risk prediction (e.g., RMST or expected mortality). The table summarises the inclusion (\ding{51}) or exclusion (\ding{55}) of a pair $(i,j)$ as comparable (CP). When tied event times ($T_i$, $T_j$) are included, a weight $\omega_o = 0.5$ is applied (see (4.6). For comparable pairs, concordant (CN) status is summarised using a weight: 0 if not concordant, 1 if concordant. Implementations where tied predictions ($M(\mathbf{x}_i) = M(\mathbf{x}_j)$) can be concordant use a 0.5 weight ($\omega_p$ in (4.6)). For \pkg{pec},  specific behaviors for tie inclusion are determined by user-defined parameters include: (1) included if  ${\tt tiedPredIn}={\tt TRUE}$, (2) included if ${\tt tiedOutcomeIn}={ \tt TRUE}$, and (3) included depending on ${\tt tiedOutcomeIn}$, ${ \tt tiedMatchIn}$, further details in Supplementary Table S5).
Those implementations with adjustment to censoring via IPCW, weight the number of concordant and comparable pairs according to a estimated censoring distribution $G(t)$. Depending on the implementation, IPCW estimation vary, further details in Supplementary Table S4. 
\review{\pkg{SurvivalEVAL} version 0.8.0 have user-defined tie-handling options; N = \texttt{None}, T = \texttt{Time}, R = \texttt{Risk}, and A = \texttt{All}, where the default is  { \tt ties = "Risk"} with a established tolerance of 1e-8.}
}
\label{supp:tableCipcw}
\end{table}

\begin{table}[htbp]
\centering
\renewcommand{\arraystretch}{1.6}
\begin{threeparttable}
\small
\setlength{\tabcolsep}{3pt}

\begin{tabular}{llll|cc|cc|ll}
\multicolumn{4}{r|}{} &
\multicolumn{2}{c|}{\rotatebox{90}{pycox.Ant}} &
\multicolumn{2}{c|}{\rotatebox{90}{pycox.Adj.Ant}} &
\multicolumn{2}{c}{\rotatebox{90}{\review{SurvivalEVAL}}} \\
\cmidrule(lr){5-6}
\cmidrule(lr){7-8}
\cmidrule(lr){9-10}

\textbf{Case} &
\textbf{Time} &
\textbf{($\Delta_i,\Delta_j$)} &
\textbf{Rank} &
CP & CN &
CP & CN &
CP & CN \\

\midrule

1A & $T_i<T_j$ & (1,1) &
$S(T_i|\mathbf{x}_i)<S(T_i|\mathbf{x}_j)$ &
\ding{51} & 1 &
\ding{51} & 1 &
N/T/R/A: \ding{51} & N/T/R/A: 1 \\

1B & $T_i<T_j$ & (1,1) &
$S(T_i|\mathbf{x}_i)>S(T_i|\mathbf{x}_j)$ &
\ding{51} & 0 &
\ding{51} & 0 &
N/T/R/A: \ding{51} & N/T/R/A: 0 \\

1C & $T_i<T_j$ & (1,1) &
$S(T_i|\mathbf{x}_i)=S(T_i|\mathbf{x}_j)$ &
\ding{51} & 0 &
\ding{51} & 0.5 &
N/T: \ding{55}; R/A: \ding{51} &
N/T: --; R/A: 0.5 \\

\midrule

2A & $T_i<T_j$ & (1,0) &
$S(T_i|\mathbf{x}_i)<S(T_i|\mathbf{x}_j)$ &
\ding{51} & 1 &
\ding{51} & 1 &
N/T/R/A: \ding{51} & N/T/R/A: 1 \\

2B & $T_i<T_j$ & (1,0) &
$S(T_i|\mathbf{x}_i)>S(T_i|\mathbf{x}_j)$ &
\ding{51} & 0 &
\ding{51} & 0 &
N/T/R/A: \ding{51} & N/T/R/A: 0 \\

2C & $T_i<T_j$ & (1,0) &
$S(T_i|\mathbf{x}_i)=S(T_i|\mathbf{x}_j)$ &
\ding{51} & 0 &
\ding{51} & 0.5 &
N/T: \ding{55}; R/A: \ding{51} &
N/T: --; R/A: 0.5 \\

\midrule

3 & $T_i<T_j$ & (0,1) & Any &
\ding{55} & -- &
\ding{55} & -- &
N/T/R/A: \ding{55} & N/T/R/A: -- \\

\midrule

4 & $T_i<T_j$ & (0,0) & Any &
\ding{55} & -- &
\ding{55} & -- &
N/T/R/A: \ding{55} & N/T/R/A: -- \\

\midrule

5A & $T_i=T_j$ & (1,1) &
$S(T_i|\mathbf{x}_i)<S(T_i|\mathbf{x}_j)$ &
\ding{55} & -- &
\ding{51} & 0.5 &
N/R: \ding{55}; T/A: \ding{51} &
N/R: --; T/A: 0.5 \\

5B & $T_i=T_j$ & (1,1) &
$S(T_i|\mathbf{x}_i)>S(T_i|\mathbf{x}_j)$ &
\ding{55} & -- &
\ding{51} & 0.5 &
N/R: \ding{55}; T/A: \ding{51} &
N/R: --; T/A: 0.5 \\

5C & $T_i=T_j$ & (1,1) &
$S(T_i|\mathbf{x}_i)=S(T_i|\mathbf{x}_j)$ &
\ding{55} & -- &
\ding{51} & 1 &
N/R: \ding{55}; T/A: \ding{51} &
N/R: --; T/A: 0.5 \\

\midrule

6A & $T_i=T_j$ & (1,0) &
$S(T_i|\mathbf{x}_i)<S(T_i|\mathbf{x}_j)$ &
\ding{51} & 1 &
\ding{51} & 1 &
N/T/R/A: \ding{51} & N/T/R/A: 1 \\

6B & $T_i=T_j$ & (1,0) &
$S(T_i|\mathbf{x}_i)>S(T_i|\mathbf{x}_j)$ &
\ding{51} & 0 &
\ding{51} & 0 &
N/T/R/A: \ding{51} & N/T/R/A: 0 \\

6C & $T_i=T_j$ & (1,0) &
$S(T_i|\mathbf{x}_i)=S(T_i|\mathbf{x}_j)$ &
\ding{51} & 0 &
\ding{51} & 0.5 &
N/T: \ding{55}; R/A: \ding{51} &
N/T: --; R/A: 0.5 \\

\midrule

7A & $T_i=T_j$ & (0,1) &
$S(T_i|\mathbf{x}_i)<S(T_i|\mathbf{x}_j)$ &
\ding{55} & -- &
\ding{51} & 0 &
N/T/R/A: \ding{51} & N/T/R/A: 0 \\

7B & $T_i=T_j$ & (0,1) &
$S(T_i|\mathbf{x}_i)>S(T_i|\mathbf{x}_j)$ &
\ding{55} & -- &
\ding{51} & 1 &
N/T/R/A: \ding{51} & N/T/R/A: 1 \\

7C & $T_i=T_j$ & (0,1) &
$S(T_i|\mathbf{x}_i)=S(T_i|\mathbf{x}_j)$ &
\ding{55} & -- &
\ding{51} & 0.5 &
N/T: \ding{55}; R/A: \ding{51} &
N/T: --; R/A: 0.5 \\

\midrule

8 & $T_i=T_j$ & (0,0) & Any &
\ding{55} & -- &
\ding{55} & -- &
N/T/R/A: \ding{55} & N/T/R/A: -- \\

\bottomrule
\end{tabular}

\vspace{6pt}

\caption{{\bf \review{Definition of comparable and concordant pairs for \pkg{pycox} and \pkg{SurvivalEVAL} estimating the time-dependent concordance index $C_{td}$.}}
For each pair $(i,j)$, different cases are defined based on the observed survival times ($T_i$, $T_j$), the corresponding censoring indicators ($\Delta_i$, $\Delta_j$) and predicted survival probabilities ($S(T_i| \mathbf{x}_i)$, $S(T_i| \mathbf{x}_j)$). Note that survival probabilities are evaluated at $T_i$ for both subjects.  The table summarises the inclusion (\ding{51}) or exclusion (\ding{55}) of a pair $(i,j)$ as comparable (CP), and the respective contribution to concordant pairs (CN). \pkg{SurvivalEVAL} version 0.8.0 have user-defined tie-handling options; N = \texttt{None}, T = \texttt{Time}, R = \texttt{Risk}, and A = \texttt{All}, where the default is  { \tt ties = "Risk"}  with a established tolerance of 1e-8. Note that \pkg{SurvivalEVAL}  also supports optional truncation at $\tau$; this table presents the untruncated case.}

\label{supp:tableCtd}

\end{threeparttable}
\end{table}

\begin{table}[htbp]
\centering
\begin{threeparttable}
\centering
  \small
   {\tabcolsep=4.25pt
   \begin{tabular}{lcl}
    \toprule
    \multicolumn{1}{c}{\multirow{1}[1]{*}{\textbf{Package}/\textit{Function }}}  &  
   \multicolumn{1}{c}{\multirow{1}[1]{*}{IPCW}} &
   \multicolumn{1}{c}{\multirow{1}[1]{*}{Definition of $W_{ij}$}}\\
   
    \midrule
      \multicolumn{3}{l}{\bf Implementations that estimate ${C}$} \\
    \midrule
          \multicolumn{1}{c}{\multirow{1}[1]{*}{\pkg{Hmisc} / \textit{rcorr.cens}} } &
          \multicolumn{1}{c}{\multirow{1}[1]{*}{\ding{55}}} & 
          \multicolumn{1}{c}{\multirow{1}[1]{*}{-}}\\
       
          \multicolumn{1}{c}{\multirow{1}[1]{*}{\pkg{SurvMetrics} / \textit{Cindex}}} &
          \multicolumn{1}{c}{\multirow{1}[1]{*}{\ding{55}}} & 
          \multicolumn{1}{c}{\multirow{1}[1]{*}{-}}   \\
          
          \multicolumn{1}{c}{\multirow{1}[1]{*}{\pkg{lifelines} /\textit{concordance\_index} }}&
          \multicolumn{1}{c}{\multirow{1}[1]{*}{\ding{55}}} & 
          \multicolumn{1}{c}{\multirow{1}[1]{*}{-}}    \\

          \multicolumn{1}{c}{\multirow{1}[1]{*}{\pkg{pysurvival} /\textit{concordance\_index} }} &
          \multicolumn{1}{c}{\multirow{1}[1]{*}{\ding{51}}} & 
          \multicolumn{1}{c}{\multirow{1}[1]{*}{$\frac{1}{\widehat G(T_i-)} \frac{1}{\widehat G(T_i)}$}} \\

          \multicolumn{1}{c}{\multirow{1}[1]{*}{\pkg{scikit-survival (sksurv)}/ }} & 
          \multicolumn{1}{c}{\multirow{2}[1]{*}{\ding{55}}} & 
          \multicolumn{1}{c}{\multirow{2}[1]{*}{-}} \\
          \multicolumn{1}{c}{\multirow{1}[1]{*}{\textit{concordance\_index\_censored} }} &  \\ 
          
          \multicolumn{1}{c}{\multirow{1}[1]{*}{\pkg{SurvivalEVAL } / ({\tt method = "Harrell"/"Naive")}} } &
          \multicolumn{1}{c}{\multirow{1}[1]{*}{\ding{55}}} & 
          \multicolumn{1}{c}{\multirow{1}[1]{*}{-}}\\
          
     \midrule
       \multicolumn{3}{l}{\bf Implementations that estimate $ C_\tau$} \\
    \midrule  
          \multicolumn{1}{c}{\multirow{1}[1]{*}{\pkg{pec} /\textit{cindex} }} & 
          \multicolumn{1}{c}{\multirow{1}[1]{*}{\ding{51}}} & 
          \multicolumn{1}{c}{\multirow{1}[1]{*}{$\frac{1}{\widehat G(T_i-)} \frac{1}{\widehat G(T_i)}$}}     \\

          \multicolumn{1}{c}{\multirow{1}[1]{*}{\pkg{survival} /\textit{concordance}({\tt timewt = "n"}) }} &
          \multicolumn{1}{c}{\multirow{1}[1]{*}{\ding{55}}} & 
          \multicolumn{1}{c}{\multirow{1}[1]{*}{-}}\\

          \multicolumn{1}{c}{\multirow{1}[1]{*}{\pkg{survival} /\textit{concordance} ({\tt timewt = "n/G2"})} } &
          \multicolumn{1}{c}{\multirow{1}[1]{*}{\ding{51}$^*$}} & 
          \multicolumn{1}{c}{\multirow{1}[1]{*}{$1/(\widehat G(T_i)^2)$}}\\
          
          \multicolumn{1}{c}{\multirow{1}[1]{*}{\pkg{survC1} /\textit{Est.Val} }}&
          \multicolumn{1}{c}{\multirow{1}[1]{*}{\ding{51}$^*$}} & 
          \multicolumn{1}{c}{\multirow{1}[1]{*}{$1/(\widehat G(T_i)^2)$}}     \\

          \multicolumn{1}{c}{\multirow{1}[1]{*}{\pkg{Scikit-survival (sksurv) }/}} & 
          \multicolumn{1}{c}{\multirow{2}[1]{*}{\ding{51}$^1$}} & 
          \multicolumn{1}{c}{\multirow{2}[1]{*}{$1/(\widehat G(T_i)^2)$}} \\
          \multicolumn{1}{c}{\multirow{1}[1]{*}{\textit{concordance\_index\_ipcw} }} &  \\

          \multicolumn{1}{c}{\multirow{1}[1]{*}{\pkg{SurvivalEVAL} ({\tt method = "Uno"/"IPCW")}}} & 
          \multicolumn{1}{c}{\multirow{1}[1]{*}{\ding{51}$^1$}} & 
          \multicolumn{1}{c}{\multirow{1}[1]{*}{$1/(\widehat G(T_i)^2)$}} \\

            \multicolumn{1}{c}{\multirow{1}[1]{*}{\pkg{SurvivalEVAL} ({\tt method = "Margin")} }} &
          \multicolumn{1}{c}{\multirow{1}[1]{*}{\ding{55}}} & 
          \multicolumn{1}{c}{\multirow{1}[1]{*}{-}}\\
    \midrule
       \multicolumn{3}{l}{\bf Implementations that estimate $ C_{td}$} \\
    \midrule  
          \multicolumn{1}{c}{\multirow{1}[1]{*}{\pkg{pycox} /\textit{concordance\_td} }} &
          \multicolumn{1}{c}{\multirow{1}[1]{*}{\ding{55}}} &
          \multicolumn{1}{c}{\multirow{1}[1]{*}{-}} \\

            \multicolumn{1}{c}{\multirow{1}[1]{*}{\pkg{SurvivalEVAL} ({\tt method = "Antolini")} }} &
          \multicolumn{1}{c}{\multirow{1}[1]{*}{\ding{55}}} &
          \multicolumn{1}{c}{\multirow{1}[1]{*}{-}} \\
          
     \bottomrule
    \end{tabular}}
\end{threeparttable}
\vspace{6pt}
\caption{{\bf \review{IPCW estimation by different C-index software implementations.} }[\ding{51}] indicates adjustment for censoring via IPCW, and [\ding{55}] indicates the exclusion of censoring adjustment. The censoring distribution, $G(t)$, can be calculated non-parametrically with Kaplan-Meier, or parametrically. ($^*$) Methods that allow for parametrical estimation of $G(t)$. \addition{($1$) Methods that use the training set to estimate IPCW.} Note that \pkg{survival} has alternative weighting schemes apart from Uno's ("n/G2"). }
\end{table}

\begin{table}[ht]
\centering
\begin{tabular}{cccccccc}
\toprule
tiedOutcomeIn & tiedPredIn & tiedMatchIn & 5C (CP) & 5C (CN) & 6C (CP) & 6C (CN) \\
\midrule
1 & 1 & 1  & \ding{51} & 1 & \ding{51} & 0.5 \\
1 & 1 & 0  & \ding{51} & 0.5 & \ding{51} & 0.5 \\
1 & 0 & 1  & \ding{51} & 1 & \ding{55} & - \\
1 & 0 & 0  & \ding{55} & - & \ding{55} & - \\
0 & 1 & 1  & \ding{51} & 1 & \ding{51} & 0.5 \\
0 & 1 & 0  & \ding{51} & 0 & \ding{51} & 0.5 \\
0 & 0 & 1  & \ding{51} & 1 & \ding{55} & - \\
0 & 0 & 0  & \ding{55} & -& \ding{55} & - \\
\bottomrule
\end{tabular}
\vspace{6pt}
\caption{{\bf Definition of comparable and concordant pairs for \pkg{pec} for ambiguous cases dependent on user-defined parameters.} 
Case 5C refers to a pair of individuals with tied event times $(T_i = T_j)$, equal status $(\Delta_i = \Delta_j = 1)$, and tied predictions ($M(\mathbf{x}_i) = M(\mathbf{x}_j)$). Case 6C refers to a pair of individuals with tied event times and predictions, and different status $(\Delta_i = 1, \Delta_j = 0)$. The table summarises the inclusion (\ding{51}) or exclusion (\ding{55}) of a pair $(i,j)$ as comparable (CP), and their respective contributions to concordance (CN).}
\end{table}

\begin{table}[htbp]
\centering
\begin{tabular}{@{}lcccc@{}}
\toprule
\textbf{Hyper-parameter}     & \textbf{DeepSurv} & \textbf{Cox-Time} & \textbf{DeepHit}  &  \textbf{RSF} \\ \midrule
Optimizer                    & adam              & adam             & adam              & - \\
Activation                   & selu              & relu             & relu              & -\\
Dense Layers              & 1                 & 2                & 2                 & - \\
Nodes / Layer             & 41                & 32, 32           & 32, 32            & 100 \\
Learning Rate                & 0.0103            & 0.01             & 0.001             & - \\
Dropout                      & 0.1601            & 0.1              & 0.6               & - \\
LR Decay                     & 0.00417           & 0                & 0                 & - \\
Batch Norm                   & True             & True             & True              & - \\
Batch Size                   & 256               & 256              & 50                & - \\
Epochs                       & 500               & 512              & 100               & - \\
Early Stopping               & False             & True             & True              & - \\
mod\_alpha                   & -                & -               & 0.2               & - \\
Sigma                        & -                & -               & 0.1               & - \\
Cuts & -                & -               & 1400               & 1400 \\ \bottomrule
\end{tabular}
\vspace{6pt}
\caption{{ \bf METABRIC optimal model hyperparameters published in the original publications for DeepSurv, Cox-Time, DeepHit and RSF (\cite{deepsurv2018, kvamme2019time, deephit2008, Ishwaran2008})}. These hyperparameter configurations were selected by the respective authors through tuning procedures on the METABRIC dataset, commonly used to benchmark survival models.}
\label{tab:hyperparams}
\end{table}

\begin{table}[htbp]
\centering
\label{supp:tab:event_censor_by_fold}
\begin{tabular}{c cc cc}
\toprule
\textbf{Fold} & \multicolumn{2}{c}{\textbf{Train (1550 obs)}} & \multicolumn{2}{c}{\textbf{Test (387--388 obs)}} \\
             & \textbf{Event (\%)} & \textbf{Censor (\%)} & \textbf{Event (\%)} & \textbf{Censor (\%)} \\
\midrule
1 &  58.45 & 41.55 & 56.59 & 43.41 \\
2 &  58.49 & 41.51 & 56.44 & 43.56 \\
3 &  58.17 & 41.83 & 57.73 & 42.27 \\
4 &  58.13 & 41.87 & 57.88 & 42.12 \\
5 &  57.16 & 42.84 & 61.76 & 38.24 \\
\bottomrule
\end{tabular}
\vspace{6pt}
\caption{{ \bf Distribution of event and censoring rates by fold for both training and test sets across 5-fold cross-validation METABRIC splits.} Stratification was performed based on the event indicator to preserve the outcome distribution across folds. }
\end{table}

\begin{table}[htbp]
\centering
\begin{tabular}{lcccc}
\toprule
 &   \shortstack{No rounding}  & \shortstack{Rounded to the \\ nearest integer} \\
\midrule
Number of individuals          & 1937      & 1937    \\
Number of unique times      & 1711     & 303    \\
Number of tied time pairs   & 246       & 7150  \\
Tied time pairs (\%)         & 0.0131  & 0.3813  \\
\bottomrule
\end{tabular}
\label{tab:tied_times_summary}
\vspace{6pt}
\caption{ { \bf Summary statistics on the impact of rounding METABRIC observed times on the number of tied times ($T_i = T_j$).}}
\end{table}

\begin{table}[htbp]
\centering
\begin{tabular}{lccc}
\toprule
 & \shortstack{No rounding}  & \shortstack{Age rounded to the \\ nearest integer} & \shortstack{Age binned to \\ 5 year windows} \\
\midrule
Number of individuals  & 1937 & 1937 & 1937      \\
Number of unique predictions & 1858 & 296 & 68        \\
Number of tied predictions pairs  & 81  & 8139  & 39314 \\
Tied prediction pairs (\%)  & 0.0043   & 0.4341    & 2.0967    \\
\bottomrule
\end{tabular}
\vspace{6pt}
\caption{{ \bf Summary statistics on the impact of rounding age on the number of tied predictions ($M(\mathbf{x}_i) = M(\mathbf{x}_j)$).} Predictions were calculated using a CPH model with age as the only covariate, the resulting survival distribution is further reduced to a measure of risk as RMST. }
\label{tab:tied_predictions_summary}
\end{table}

\begin{table}[htbp]
\centering
\renewcommand{\arraystretch}{1.45}
\begin{threeparttable}
\small
\setlength{\tabcolsep}{3pt}

\begin{tabular}{lllllll}
\toprule
\textbf{Case} &
\textbf{Observed time} &
\textbf{$(\Delta_i,\Delta_j)$} &
\textbf{Effective time / Rank} &
\textbf{CP (Marginal)} &
\textbf{CN (Marginal)} &
\textbf{Weight} \\
\midrule

1A & $T_i<T_j$ & (1,1) &
$M_i>M_j$ &
N/T/R/A: \ding{51} &
N/T/R/A: 1 & $1$ \\

1B & $T_i<T_j$ & (1,1) &
$M_i<M_j$ &
N/T/R/A: \ding{51} &
N/T/R/A: 0 & $1$ \\

1C & $T_i<T_j$ & (1,1) &
$M_i=M_j$ &
N/T: \ding{55}; R/A: \ding{51} &
N/T: --; R/A: 0.5 & $1$ \\

\midrule

2A & $T_i<T_j$ & (1,0) &
$M_i>M_j$ &
N/T/R/A: \ding{51} &
N/T/R/A: 1 & $1$ \\

2B & $T_i<T_j$ & (1,0) &
$M_i<M_j$ &
N/T/R/A: \ding{51} &
N/T/R/A: 0 & $1$ \\

2C & $T_i<T_j$ & (1,0) &
$M_i=M_j$ &
N/T: \ding{55}; R/A: \ding{51} &
N/T: --; R/A: 0.5 & $1$ \\

\midrule

3A & $T_i<T_j$ & (0,1) &
$\widetilde T_i<T_j,\ M_i>M_j$ &
N/T/R/A: \ding{51} &
N/T/R/A: 1 & $p_i$ \\

3B & $T_i<T_j$ & (0,1) &
$\widetilde T_i<T_j,\ M_i<M_j$ &
N/T/R/A: \ding{51} &
N/T/R/A: 0 & $p_i$ \\

3C & $T_i<T_j$ & (0,1) &
$\widetilde T_i<T_j,\ M_i=M_j$ &
N/T: \ding{55}; R/A: \ding{51} &
N/T: --; R/A: 0.5 & $p_i$ \\

3D & $T_i<T_j$ & (0,1) &
$\widetilde T_i>T_j,\ M_i>M_j$ &
N/T/R/A: \ding{51} &
N/T/R/A: 0 & $p_i$ \\

3E & $T_i<T_j$ & (0,1) &
$\widetilde T_i>T_j,\ M_i<M_j$ &
N/T/R/A: \ding{51} &
N/T/R/A: 1 & $p_i$ \\

3F & $T_i<T_j$ & (0,1) &
$\widetilde T_i>T_j,\ M_i=M_j$ &
N/T: \ding{55}; R/A: \ding{51} &
N/T: --; R/A: 0.5 & $p_i$ \\

3G & $T_i<T_j$ & (0,1) &
$\widetilde T_i=T_j,\ \text{any rank}$ &
N/R: \ding{55}; T/A: \ding{51} &
N/R: --; T/A: 0.5 & $p_i$ \\

\midrule

4A & $T_i<T_j$ & (0,0) &
$\widetilde T_i<\widetilde T_j,\ M_i>M_j$ &
N/T/R/A: \ding{51} &
N/T/R/A: 1 & $p_ip_j$ \\

4B & $T_i<T_j$ & (0,0) &
$\widetilde T_i<\widetilde T_j,\ M_i<M_j$ &
N/T/R/A: \ding{51} &
N/T/R/A: 0 & $p_ip_j$ \\

4C & $T_i<T_j$ & (0,0) &
$\widetilde T_i<\widetilde T_j,\ M_i=M_j$ &
N/T: \ding{55}; R/A: \ding{51} &
N/T: --; R/A: 0.5 & $p_ip_j$ \\

4D & $T_i<T_j$ & (0,0) &
$\widetilde T_i>\widetilde T_j,\ M_i>M_j$ &
N/T/R/A: \ding{51} &
N/T/R/A: 0 & $p_ip_j$ \\

4E & $T_i<T_j$ & (0,0) &
$\widetilde T_i>\widetilde T_j,\ M_i<M_j$ &
N/T/R/A: \ding{51} &
N/T/R/A: 1 & $p_ip_j$ \\

4F & $T_i<T_j$ & (0,0) &
$\widetilde T_i>\widetilde T_j,\ M_i=M_j$ &
N/T: \ding{55}; R/A: \ding{51} &
N/T: --; R/A: 0.5 & $p_ip_j$ \\

4G & $T_i<T_j$ & (0,0) &
$\widetilde T_i=\widetilde T_j,\ \text{any rank}$ &
N/R: \ding{55}; T/A: \ding{51} &
N/R: --; T/A: 0.5 & $p_ip_j$ \\

\midrule

5A & $T_i=T_j$ & (1,1) &
$M_i>M_j$ &
N/R: \ding{55}; T/A: \ding{51} &
N/R: --; T/A: 0.5 & $1$ \\

5B & $T_i=T_j$ & (1,1) &
$M_i<M_j$ &
N/R: \ding{55}; T/A: \ding{51} &
N/R: --; T/A: 0.5 & $1$ \\

5C & $T_i=T_j$ & (1,1) &
$M_i=M_j$ &
N/R: \ding{55}; T/A: \ding{51} &
N/R: --; T/A: 0.5 & $1$ \\

\midrule

6A & $T_i=T_j$ & (1,0) &
$M_i>M_j$ &
N/T/R/A: \ding{51} &
N/T/R/A: 1 & $1$ \\

6B & $T_i=T_j$ & (1,0) &
$M_i<M_j$ &
N/T/R/A: \ding{51} &
N/T/R/A: 0 & $1$ \\

6C & $T_i=T_j$ & (1,0) &
$M_i=M_j$ &
N/T: \ding{55}; R/A: \ding{51} &
N/T: --; R/A: 0.5 & $1$ \\

\midrule

7 & $T_i=T_j$ & (0,0) &
$\widetilde T_i=\widetilde T_j,\ \text{any rank}$ &
N/R: \ding{55}; T/A: \ding{51} &
N/R: --; T/A: 0.5 & $p_ip_j$ \\

\bottomrule
\end{tabular}
\vspace{6pt}
\caption{{\bf \review{Definition of comparable and concordant pairs for
\pkg{SurvivalEVAL} using the marginal method to estimate the concordance index.}} For each pair $(i,j)$, different cases are defined based on the observed survival times ($T_i$, $T_j$), the corresponding event indicators ($\Delta_i$, $\Delta_j$), and the predicted risk scores ($M_i$, $M_j$), where $\Delta_k=1$ denotes an observed event and $\Delta_k=0$ denotes right censoring. For censored individuals, the marginal method estimates an effective event time $\widetilde{T}_k$ using the Kaplan--Meier best-guess estimator fitted on the training data; for individuals with observed events, $\widetilde{T}_k=T_k$. The table summarises the inclusion (\ding{51}) or exclusion (\ding{55}) of each pair as comparable (CP), its concordance contribution (CN), and its corresponding pair weight. Pairs with known observed event ordering retain unit weight,
whereas pairs requiring imputed event-time comparisons receive partial weights, with $p_k=1-\widehat{S}_{\mathrm{train}}(T_k)$ for censored individuals. CN denotes the concordance before multiplication by the pair weight: concordant pairs contribute $1$, discordant pairs
contribute $0$, and included tied pairs contribute $0.5$. The package has user-defined tie-handling options: N = \texttt{None}, T = \texttt{Time}, R = \texttt{Risk}, and A = \texttt{All}, where the default is  {\tt ties = "Risk"} with a established tolerance of 1e-8.
}
\end{threeparttable}
\end{table}

\newpage
\FloatBarrier

\section{Supplementary Figures}

\begin{figure}[htbp]
    \centering
    \includegraphics[width=\textwidth]{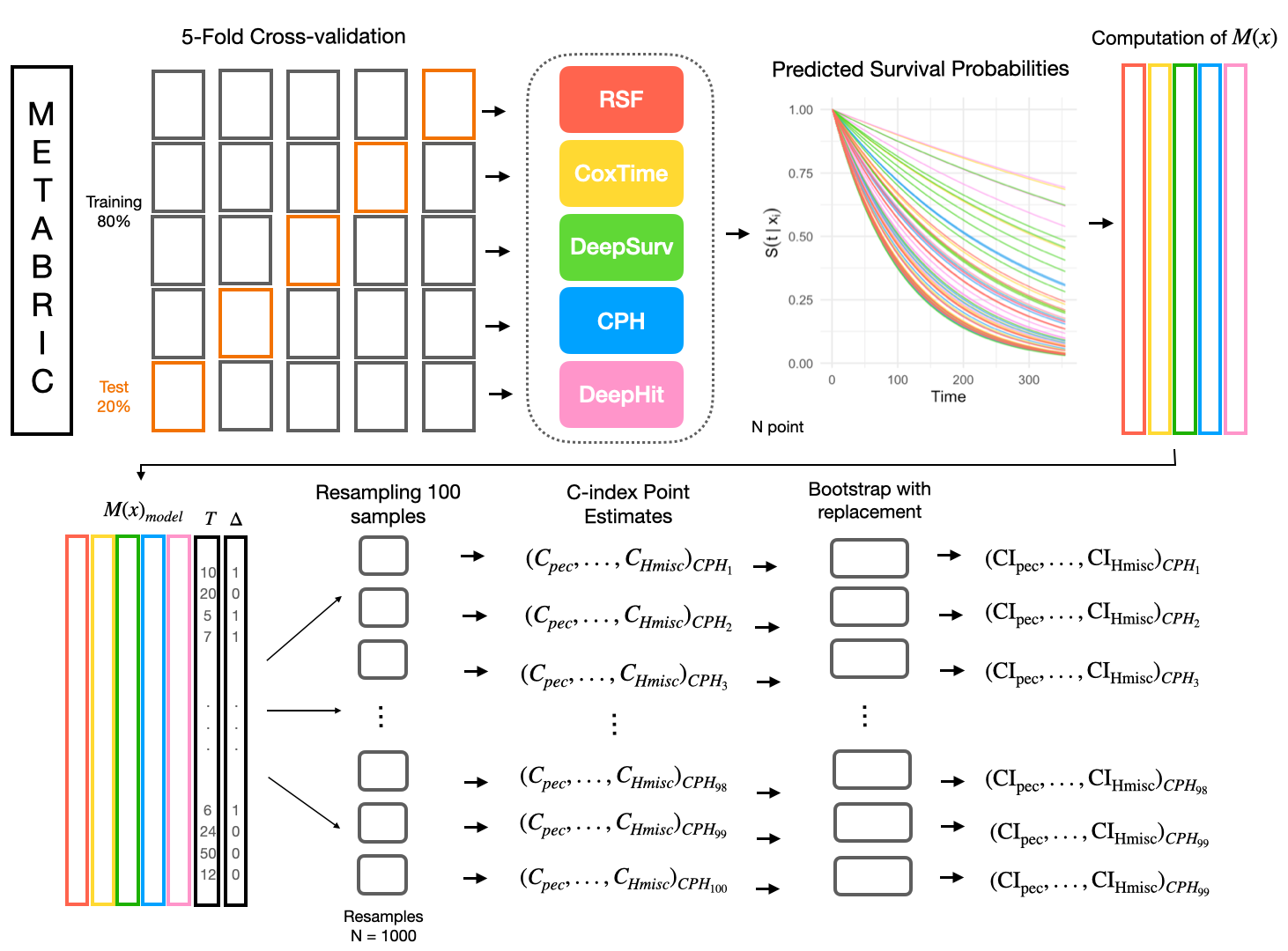}
    \caption{{\bf Overview of model evaluation using 5-fold cross-validation and bootstrap estimation of C-index implementations.} METABRIC dataset is split into 80\% training and 20\% test sets with stratified sampling to ensure similar event percentages between folds. Within each training set models are fit, and used for predicting on the respective test sets generating survival probabilities. Derived from the survival probabilities are the scalar risk predictions $M(x_i)$. These predictions, along with the observed times $T$ and event indicator $Delta$, are used to calculate C-index point estimates with multiple implementations (e.g $(C_{pec}, ..., C_{Hmisc})_{CPH_1}$, $(C_{pec}, ..., C_{Hmisc})_{DeepHit_1}$), resulting in a distribution of scores per implementation and model. To estimate uncertainty, 100 bootstrap samples (with replacement) of size 1000 are drawn. If hold-out validation is used, these 100 samples are drawn from the test set of its same size. For each bootstrap sample we compute the 95\% confidence intervals (e.g $(CI_{pec}, ..., CI_{Hmisc})_{CPH_1}$, $(CI_{pec}, ..., CI_{Hmisc})_{DeepHit_1}$). Models include CPH, DeepSurv, Cox-Time, DeepHit and RSF (\cite{Cox1972, deepsurv2018, kvamme2019time, deephit2008, Ishwaran2008}) }
    \label{supp:fig:5foldsetup}
\end{figure}

\begin{figure}[htbp]
    \centering
    \includegraphics[width=\textwidth]{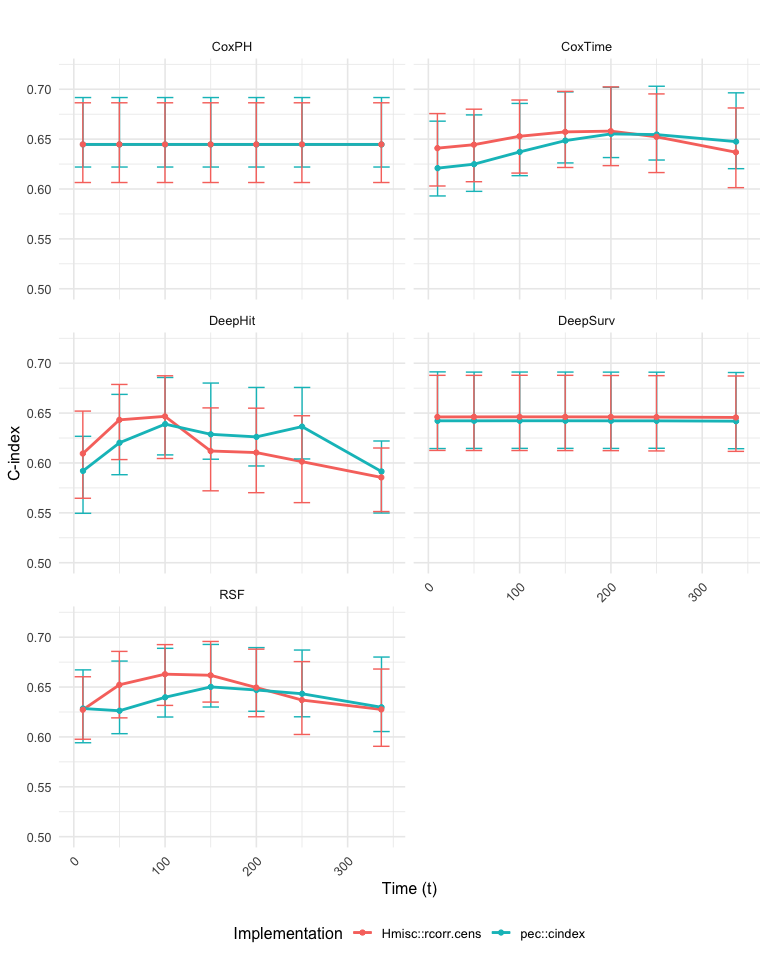}
    \caption{{ \bf Comparison of C-index estimates obtained using \pkg{Hmisc} and \pkg{pec} implementations across time points, evaluated on the first fold during hold-out cross-validation and bootstrap on METABRIC dataset.} The transformation from survival probability into risk is defined as $M(\mathbf{x}_i) = 1- S(t|\mathbf{x}_i)$. While \pkg{Hmisc} estimates $C$, \pkg{pec} estimates $C_\tau$ where $\tau = \max(T_i :\Delta_i=1)$ and adjust for censoring. }
    \label{fig:fold1_riskt}
\end{figure}

\begin{figure}[htbp]
    \centering
    \includegraphics[width=\textwidth]{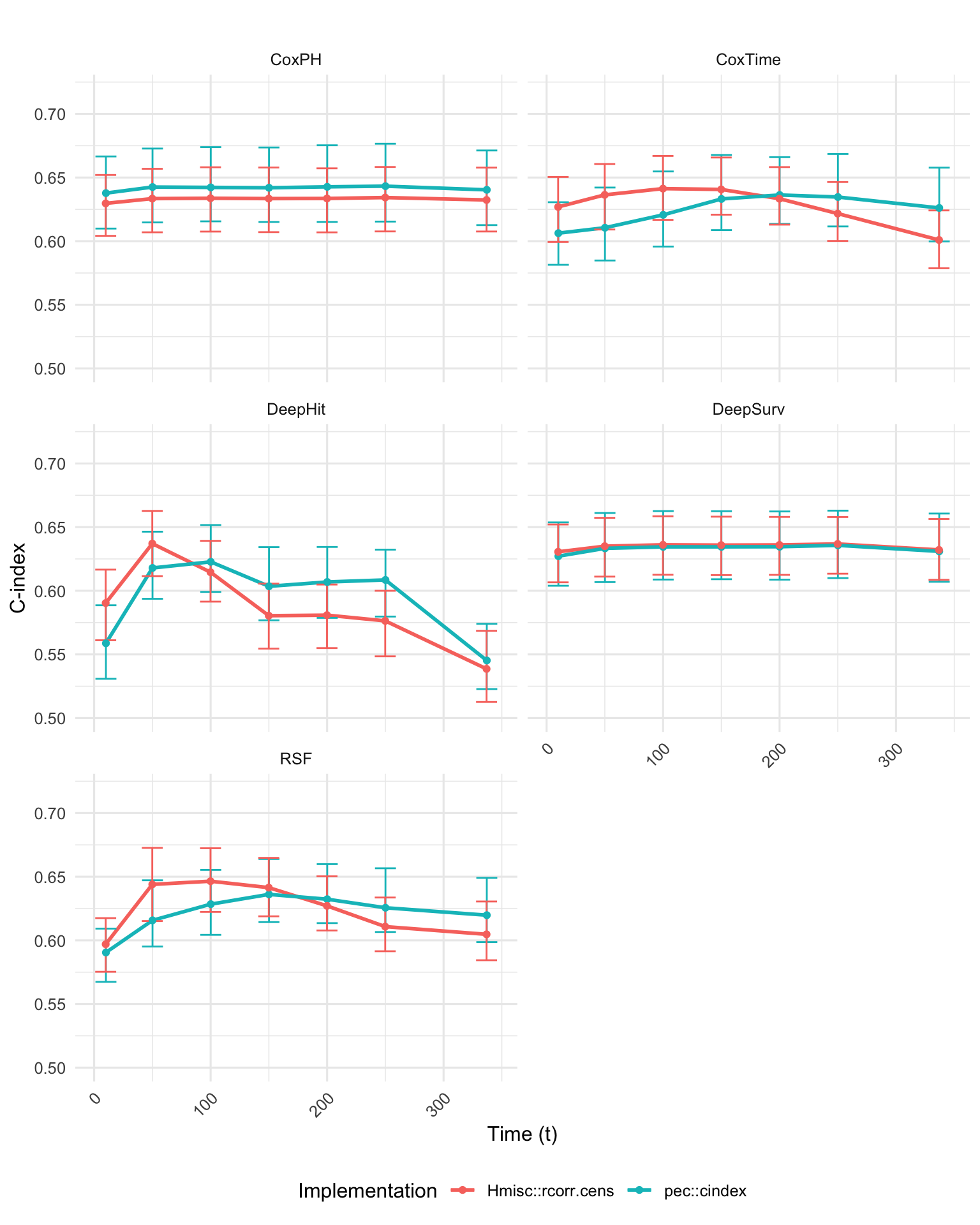}
    \caption{{ \bf Comparison of C-index estimates obtained using \pkg{Hmisc} and \pkg{pec} implementations across time points, evaluated on the first fold during 5-fold cross-validation and bootstrap on METABRIC dataset.} The transformation from survival probability into risk is defined as $M(\mathbf{x}_i) = 1- S(t|\mathbf{x}_i)$. While \pkg{Hmisc} estimates $C$, \pkg{pec} estimates $C_\tau$ where $\tau = \max(T_i :\Delta_i=1)$ and adjust for censoring. }
    \label{supp:fig:riskT5fold}
\end{figure}

\begin{figure}[htbp]
    \centering
    \includegraphics[width=\textwidth]{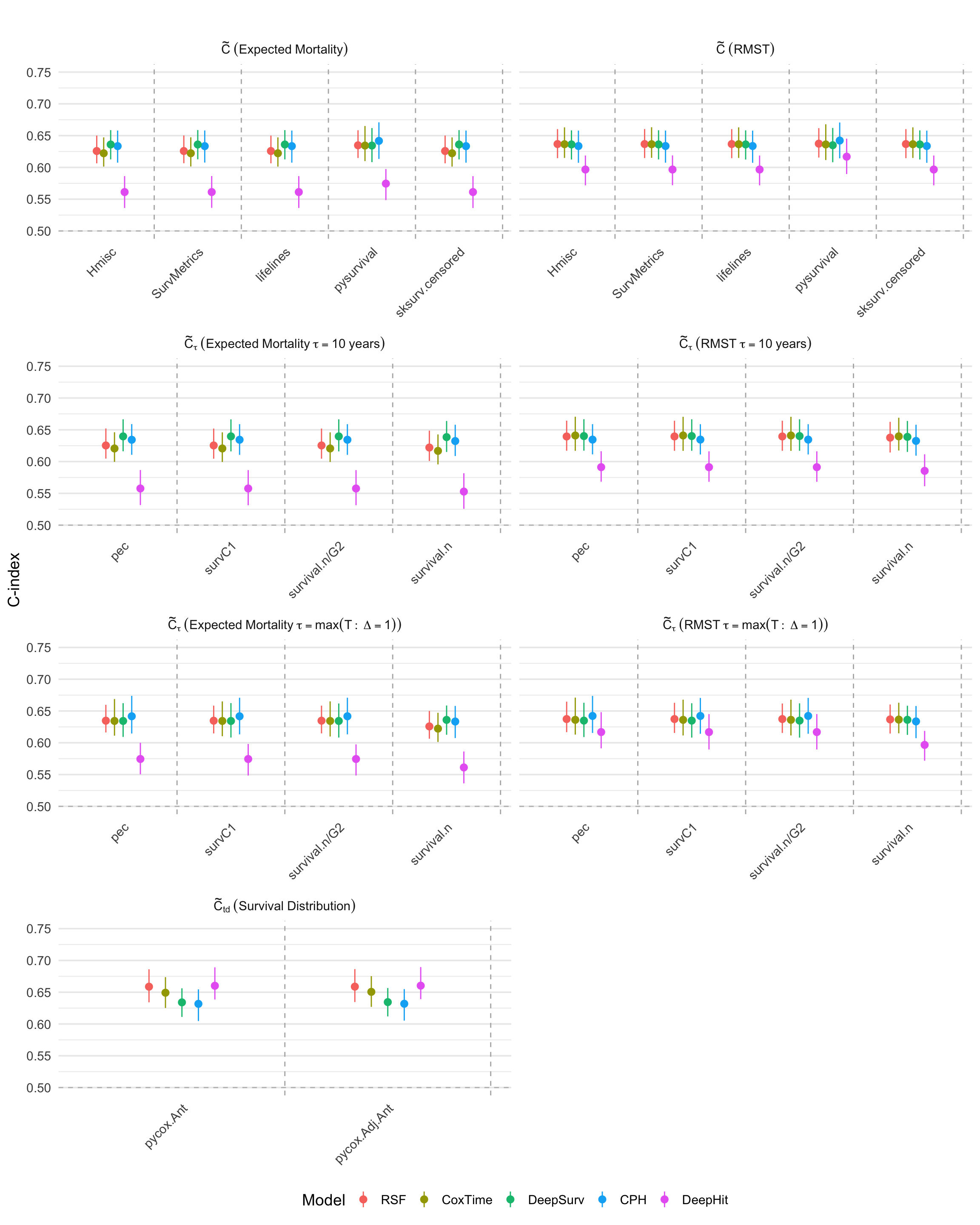}
    \caption{{ \bf Comparison of C-index estimates calculated by different implementations during 5-fold cross-validation within METABRIC.} Dots represent the estimates and vertical lines represent the 95\% confidence interval which was calculated using bootstrap. Estimates are grouped by the corresponding definition of the concordance probability ($C$, $C_\tau$ or $C_{td}$). For $C_\tau$, we consider $\tau $ equal to 10 years or to the maximum of the uncensored survival times. For $C$ and $C_{\tau}$ two transformations of the survival function were used: expected mortality and negative RMST. {\tt sksurv.censored} and {\tt sksurv.ipcw} indicate whether the {\tt concordance\_index\_censored} or {\tt concordance\_index\_ipcw} functions were used. {\tt survival.n/G2} and {\tt survival.n} indicate different weighting options (IPCW for {\tt survival.n/G2}, uniform for {\tt survival.n}). {\tt pycox.Ant} is the original Antolini's estimator, {\tt pycox.Adj.Ant} is an adjusted version which applies the modifications suggested by \cite{Ishwaran2008}. }
    \label{supp:fig:5fold_Cestimate}
\end{figure}

\begin{figure}[htbp]
    \centering
    \includegraphics[width=\textwidth]{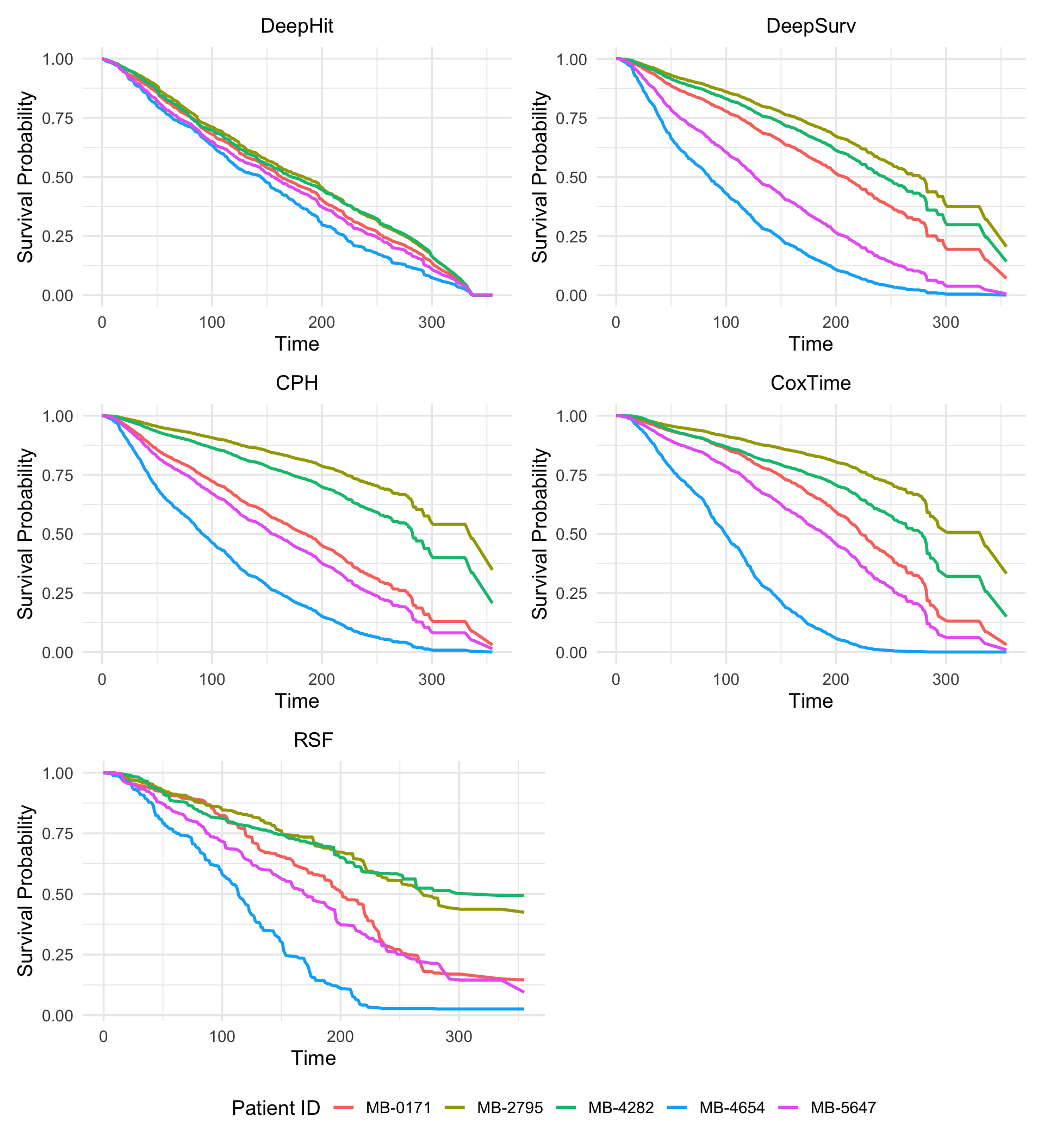}
    \caption{{ \bf Survival curves estimated for five random selected patients from the first fold during hold-out cross-validation on the METABRIC dataset.} Each curve represents the predicted survival probability $S(t|\mathbf{x}_i)$ over time for a given patient, based on the covariates $\mathbf{x}_i$. Note that lack of separation between subjects' survival curves, may indicate lack of model calibration (e.g DeepHit).}
    \label{supp:fig:curves}
\end{figure}

\begin{figure}[htbp]
    \centering
    \includegraphics[width=\textwidth]{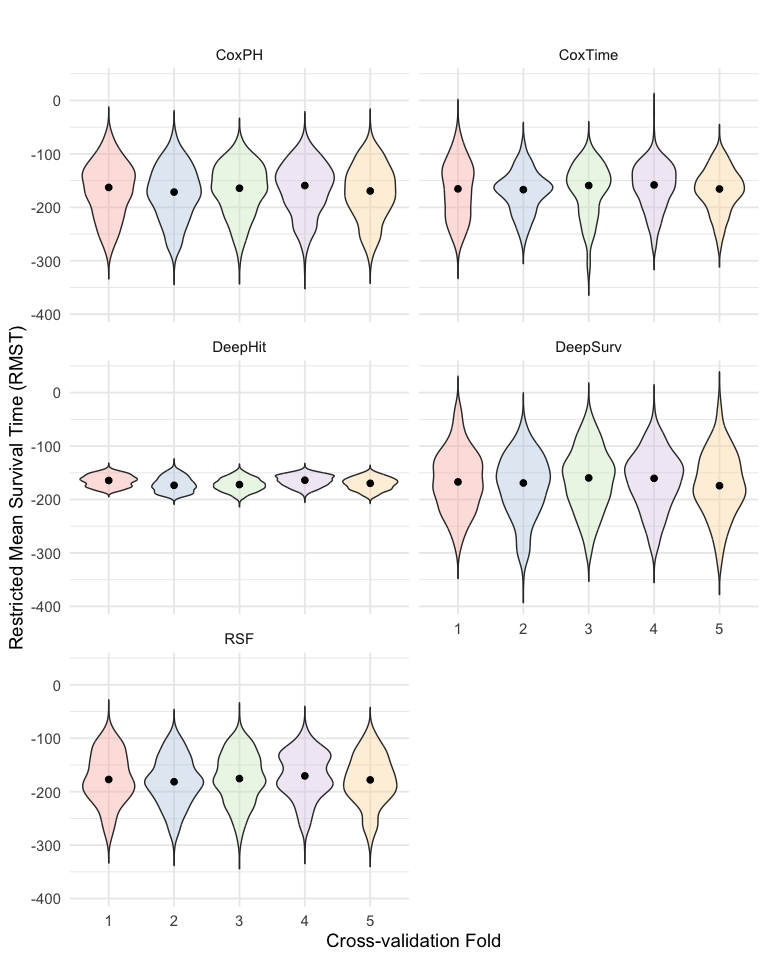}
    \caption{{ \bf Distribution of Restricted Mean Survival Time (RMST) estimates across different across different cross-validation folds and models.}  RMST values are computed as a reduction od the predicted survival probabilities to a scalar measure of risk: $- \sum_{t \in \mathcal{T},\, t \leq T^*} S(t \mid \mathbf{x}_i) \cdot \Delta t$. Further details in RMST calculation are described in Section 4.3}
    \label{supp:fig:RMSTacrossfolds}
\end{figure}

\begin{figure}[htbp]
    \centering
    \includegraphics[width=\textwidth]{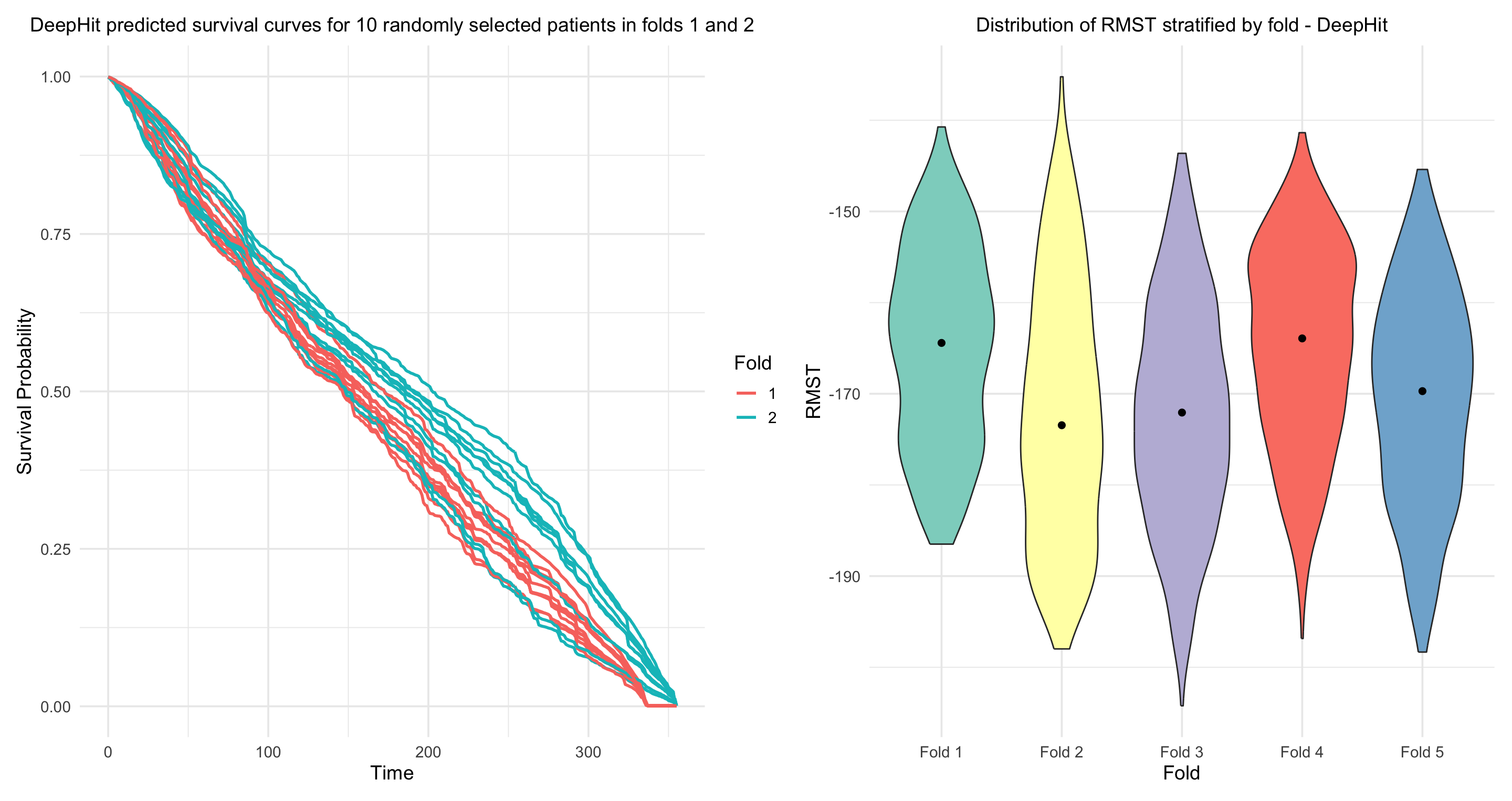}
    \caption{{ \bf DeepHit predictions stratified by fold.} The left panel shows survival curves predicted by the DeepHit model for 10 randomly selected patients in two different folds of the cross-validation, illustrating trajectories variability across patients and between folds. The right panel shows the distribution of Restricted Mean Survival Time (RMST) estimates for each fold. }
    \label{supp:fig:DeepHitFolds}
\end{figure}

\begin{figure}[htbp]
    \centering
    \includegraphics[width=\textwidth]{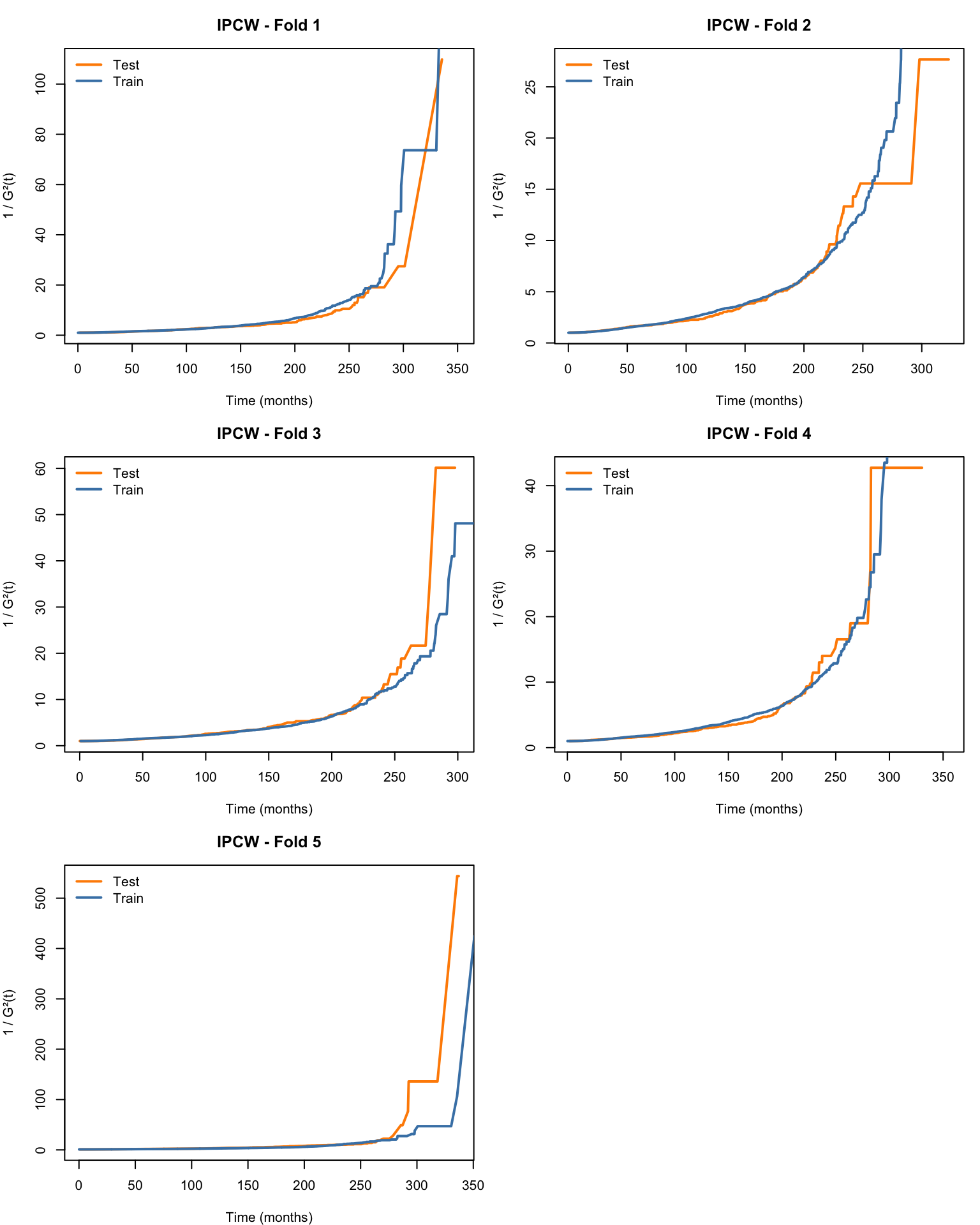}
    \caption{{ \bf Inverse probability of censoring weights (IPCW) for the training and test set estimated at unique times computed with \pkg{pec} on the whole METABRIC dataset and illustrated by fold.} The IPCW account for censoring and are utilized to adjust the C-index to censoring. Variations in the IPCW by training or test set indicate that C-index estimates may be sensitive to the train/test split censoring distribution, as well as by fold, affecting the stability of the C-index. }
    \label{supp:fig:ipcw}
\end{figure}

\begin{figure}[htbp]
    \centering
    \includegraphics[width=\textwidth]{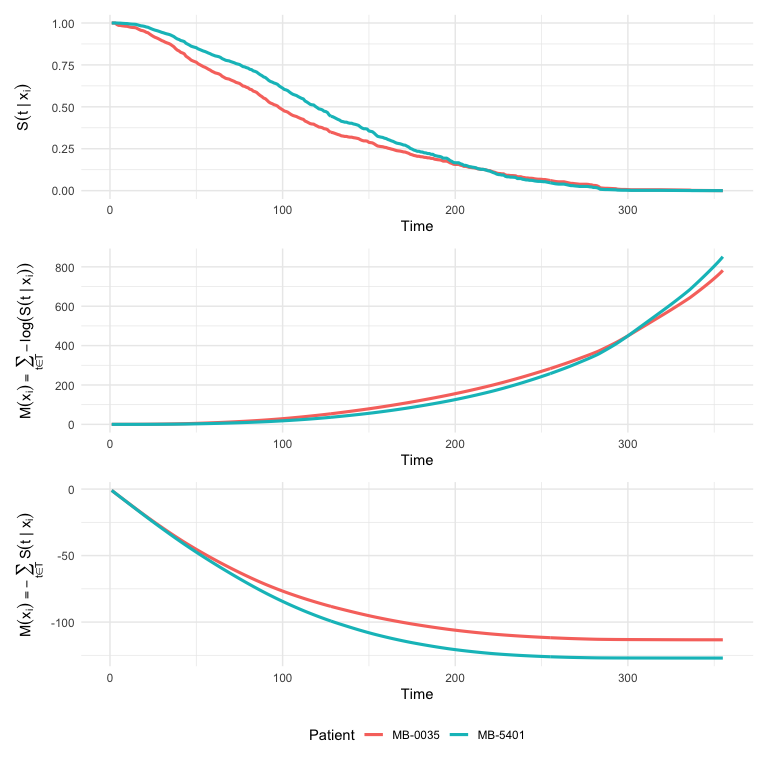}
    \caption{{ \bf Comparison of C-index inputs for two selected patients: survival probabilities, cumulative expected mortality (EM), and cumulative restricted mean survival time (RMST).} Survival probabilities are predicted by Cox-Time model in fold 1 during hold-out validation. Measures of risk $M(\mathbf{x}_i)$, such as EM and -RMST (i.e used as a risk measure) are derived from the predicted survival probabilities. Higher estimates of EM and -RMST indicate a higher risk, which relates to lower survival probabilities. At later time points, EM and -RMST differ in their ranking for these two subjects, indicating that EM weights later time points in comparison with RMST, potentially due to numerical instabilities in the EM logarithmic transformation of the survival probabilities. While -RMST tends to preserve the overall ranking across time. This figure illustrates that a reduction from survival probabilities to a measure of risk is non-trivial. Further details of the transformations are detailed in Section 4.3.
    }
    \label{supp:fig:inestability}
\end{figure}

\begin{figure}[htbp]
    \centering
    \includegraphics[width=\textwidth]{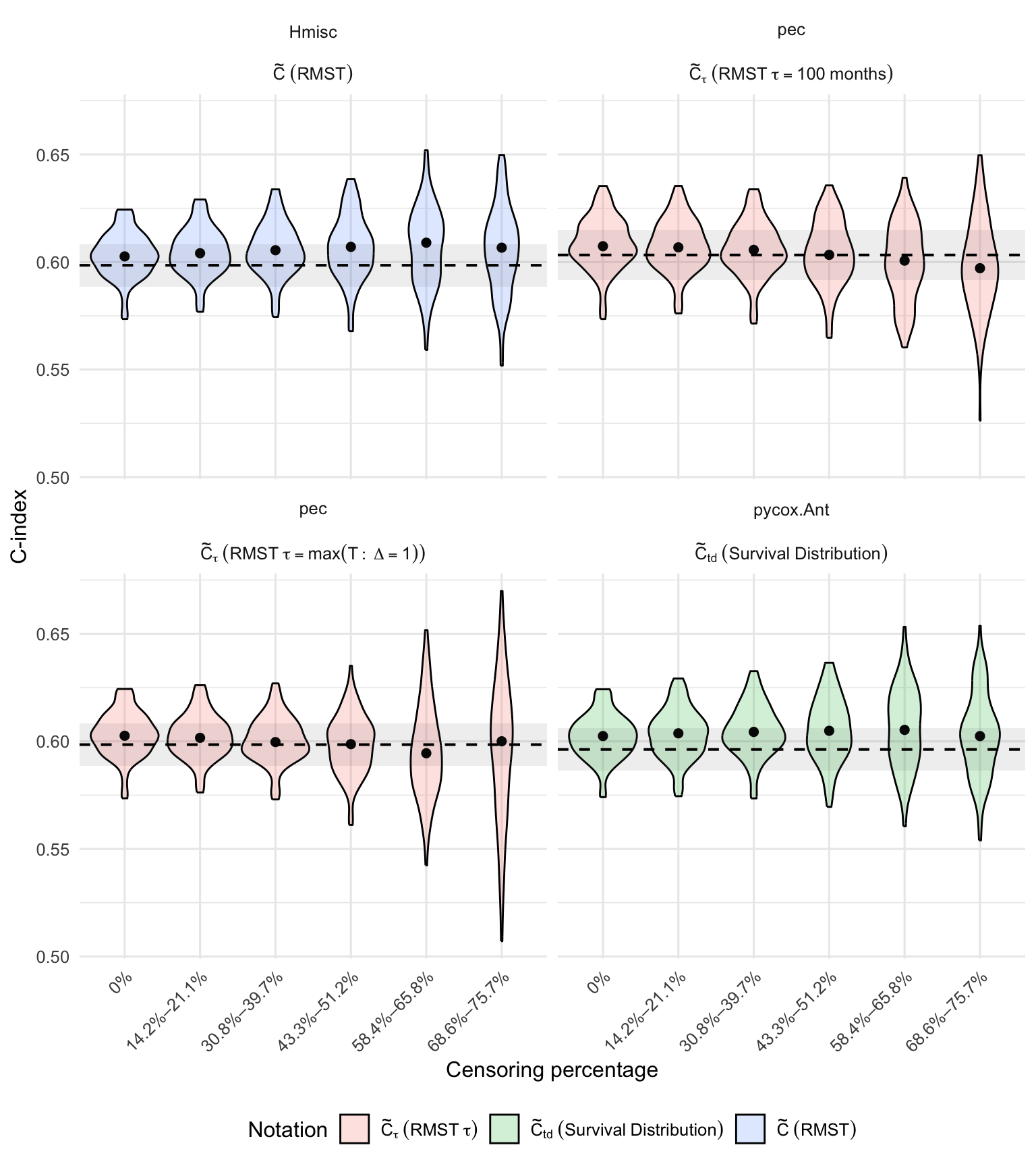}
    \caption{{ \bf Variation in C-index estimates across increasing levels of censoring in semi-synthetic data.} C-index values are estimated from CPH model predictions via 5-fold cross-validation. C-index estimates are estimated per each censoring level from 100 semi-synthetic datasets, where the dot represents the median across datasets. Data generation mechanism for censoring and event times is Weibull. C-index estimates are computed based on the survival distribution ($C_{td}$), and RMST as the transformation of choice ($C$ and $C_{\tau}$) with different time truncations ($\tau = 100$ or $\tau = max(T : \Delta =1)$). Dotted line represent the median C-index oracles, and grey band the standard deviation. Further details on semi-synthetic data generation and oracle calculation are included in Supplementary Note S2 and S3.}
    \label{fig:Weibull4panels}
\end{figure}

\begin{figure}[htbp]
    \centering
    \includegraphics[width=\textwidth]{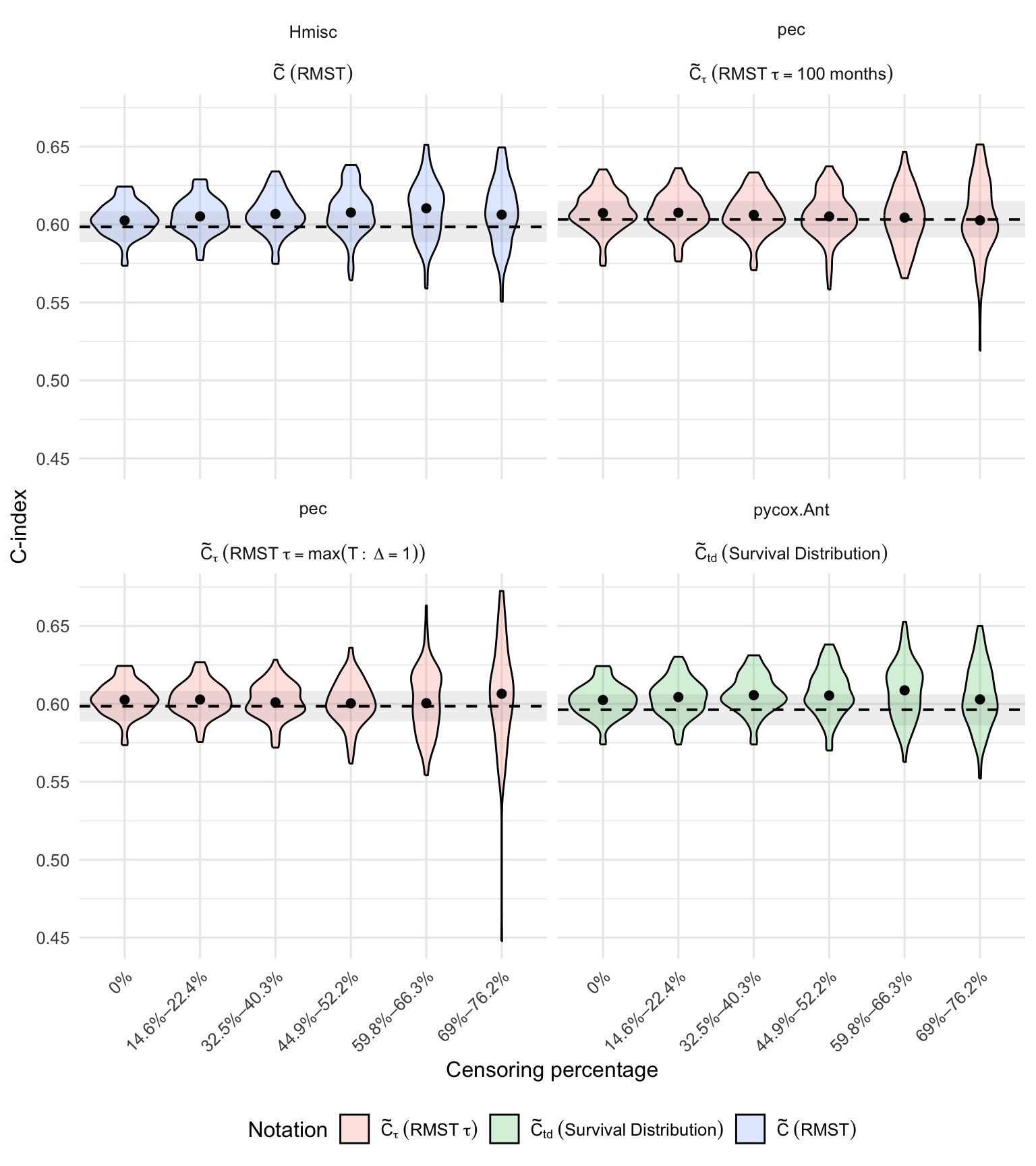}
    \caption{{ \bf Variation in C-index estimates across increasing levels of age-informed censoring in semi-synthetic data.} C-index values are estimated from CPH model predictions via 5-fold cross-validation. Each censoring level contains 100 semi-synthetic datasets, with C-index estimates computed per dataset together with the median within each censoring level. Event times are generated from a Weibull distribution and censoring from an age-inform Weibull distribution. C-index estimates are computed based on the survival distribution ($C_{td}$), and RMST as the transformation of choice ($C$ and $C_{\tau}$) with different time truncations ($\tau = 100$ or $\tau = max(T : \Delta =1)$). Dotted line represent the median C-index oracles, and grey band the standard deviation. Further details on semi-synthetic data generation and oracle calculation are included in Supplementary Note S2 and S3.}
    \label{fig:Inform4panels}
\end{figure}

\begin{figure}[htbp]
    \centering
    \includegraphics[width=\textwidth]{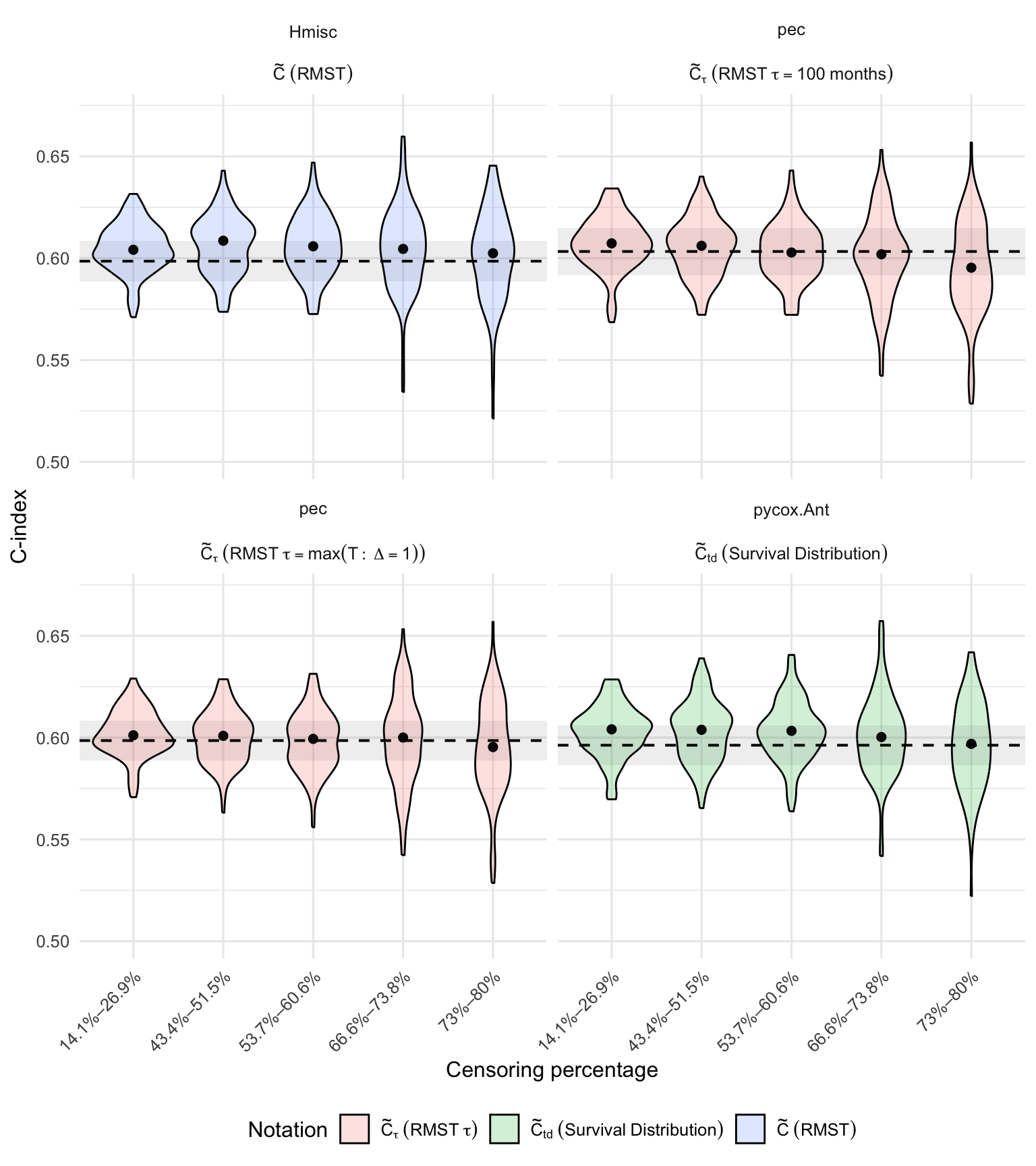}
    \caption{{ \bf Variation in C-index estimates across increasing levels of uniform censoring in semi-synthetic data.} C-index values are estimated from CPH model predictions via 5-fold cross-validation. Each censoring level contains 100 semi-synthetic datasets, with C-index estimates computed per dataset together with the median within each censoring level. Event times are generated from a Weibull distribution and censoring from a Uniform distribution. C-index estimates are computed based on the survival distribution ($C_{td}$), and RMST as the transformation of choice ($C$ and $C_{\tau}$) with different time truncations ($\tau = 100$ or $\tau = max(T : \Delta =1)$). Dotted line represent the median C-index oracles, and grey band the standard deviation. Further details on semi-synthetic data generation and oracle calculation are included in Supplementary Note S2 and S3.}
    \label{fig:Uniform4panels}
\end{figure}

\newpage
\FloatBarrier

\begin{figure}[H]
\centering
\includegraphics[width=\textwidth]{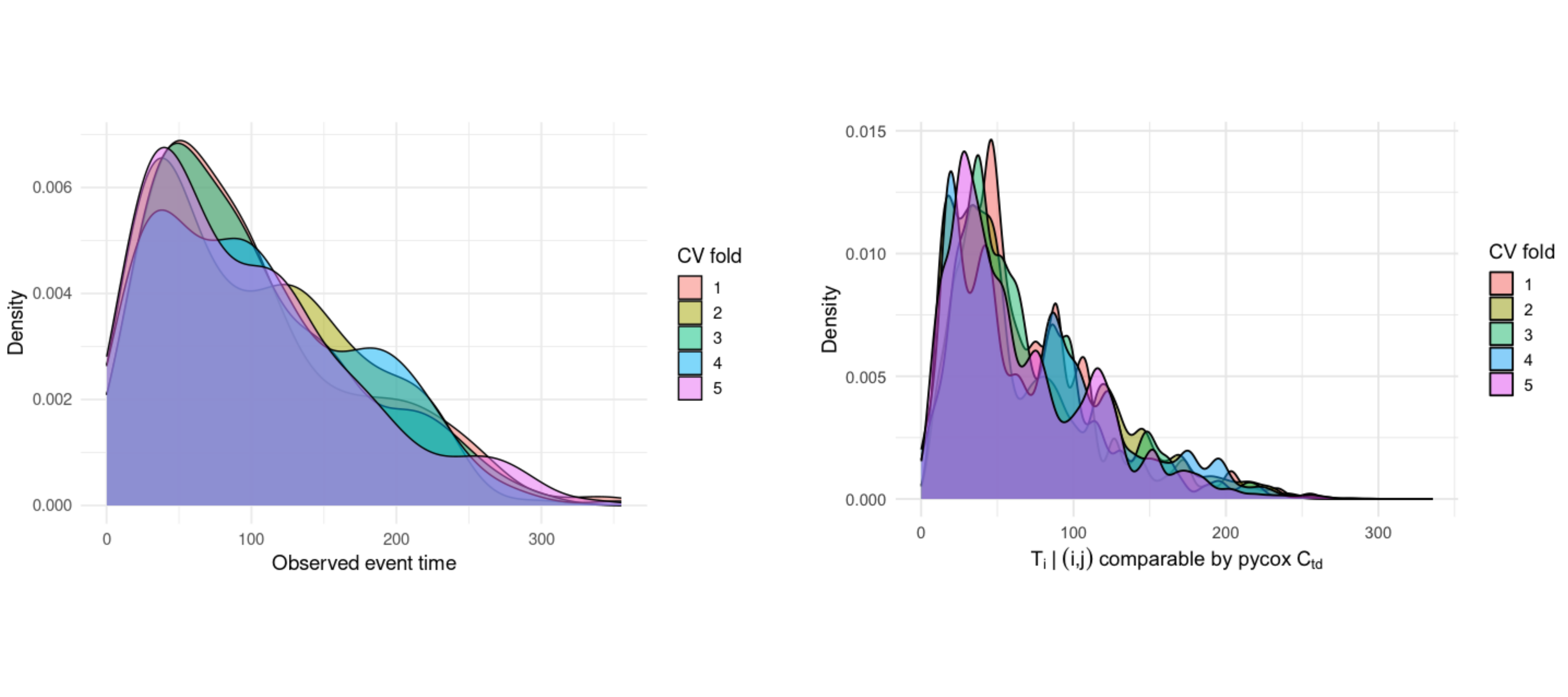}
\vspace{-10mm}
\caption{Distribution of observed event times in METABRIC and the distribution of $T_i$ among the comparable pairs used by $C_{td}$. The distributions are stratified by cross-validation fold. Left: density of observed event times. Right: density of $T_i$ in comparable pairs $(i,j)$ according to $C_{td}$. A subject's observed event time $T_i$ may be used across multiple observation pairs, particularly for early event times.}
\label{fig:supp_ct_distrib}
\end{figure}

\begin{figure}[H]
\centering
\includegraphics[width=\textwidth]{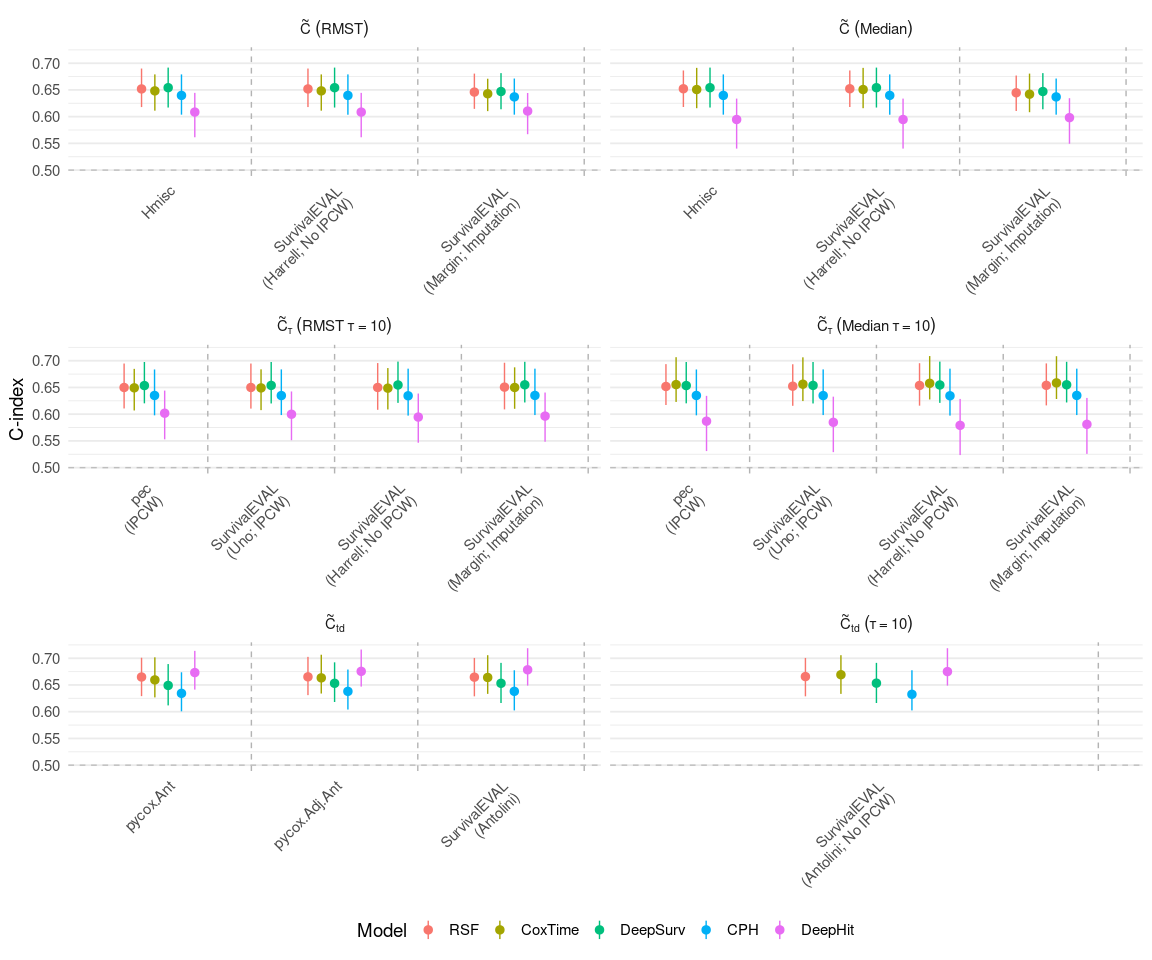}
\caption{{\bf Comparison of C-index estimates calculated by different implementations during 5-fold cross-validation within METABRIC.} Dots represent the estimates and vertical lines the 95\% bootstrap confidence intervals. Estimates are grouped according to the corresponding definition of the concordance probability ($C$, $C_\tau$, or $C_{td}$) and the transformation used for ranking. For $C$ and $C_\tau$, predictions are summarized using either negative restricted mean survival time (RMST) or negative predicted median survival time, the latter being the default transformation in \pkg{SurvivalEVAL} version 0.8.0. For $C_\tau$, $\tau$ is set to 10 years. Labels additionally indicate whether censoring is not adjusted for (No IPCW), handled using inverse probability of censoring weighting (IPCW) or through ``Imputation'' in the marginal C-index. For $C_{td}$, {\tt pycox.Ant} denotes the original Antolini estimator and {\tt pycox.Adj.Ant} the adjusted version incorporating the modifications suggested by Ishawaran and others (2008). The final panel shows Antolini's $C_{td}$ with $\tau=10$ years as implemented in \pkg{SurvivalEVAL}; this combination does not use IPCW. Throughout, prediction ties are handled using the \pkg{SurvivalEVAL} ``Risk'' option.}
\label{fig:event_time_distribution}
\end{figure}


\end{document}